\documentclass[11pt,twoside]{article}

\usepackage[dvipsnames]{xcolor}

\usepackage{fullpage}
\usepackage{mathtools}

\usepackage{epsf}
\usepackage{fancyhdr}
\usepackage{graphics}
\usepackage{graphicx}
\usepackage{psfrag}

\usepackage{booktabs}
\usepackage[linesnumbered,ruled]{algorithm2e}
\DontPrintSemicolon
\SetAlFnt{\small\sffamily} 

\usepackage{caption,subcaption}

\usepackage{float}
\usepackage{color}

\usepackage{etoc}

\usepackage{enumitem}

\usepackage{amsfonts,amsmath,amssymb,bbm}

\usepackage{nicefrac}
\usepackage{chngpage}

\usepackage{jmlr2e}

\newcommand{\myvec}{u} 
\newcommand{\myvectwo}{v} 
\newcommand{\mymat}{B}
\newcommand{\set}{{A}}
\newcommand{\setiso}{{S}}
\newcommand{\scalar}{\alpha}
\newcommand{\tvscalar}{\epsilon}
\newcommand{\tailboundscalar}{t}

\newcommand{\term}{M}
\newcommand{\bigterm}{T}
\newcommand{\tagmala}{\text{\tiny MALA}}
\newcommand{\tagmrw}{\text{\tiny MRW}}

\newcommand{\scparam}{m}
\newcommand{\smoothness}{L}
\newcommand{\target}{\pi}
\newcommand{\trunctarget}{\target_\convex}
\newcommand{\truncTarget}{\Target_\convex}

\newcommand{\myfun}{f}
\newcommand{\gradf}{\nabla \myfun}
\newcommand{\hessf}{\nabla^2 \myfun}

\newcommand{\indicator}{\mathbbm{1}}

\newcommand{\step}{h}
\newcommand{\stepfun}{w}
\newcommand{\stepext}{\tilde\stepfun}
\newcommand{\truncball}{\mathcal{R}}
\newcommand{\radius}{r}
\newcommand{\maxgrad}{\mathcal{D}}
\newcommand{\mean}{\mu}

\newcommand{\diracdelta}{\mathbf{\delta}}
\newcommand{\proposal}{\mathcal{P}}
\newcommand{\density}{p}
\newcommand{\transdensity}{q}
\newcommand{\Target}{\ensuremath{\Pi}}
\newcommand{\stationary}{\Target}

\newcommand{\initial}{\mu_0}
\newcommand{\initialstar}{\mu_\star}
\newcommand{\rvg}{\xi} 

\newcommand{\approxmyfun}{\tilde{\myfun}}
\newcommand{\approxTarget}{\tilde{\Target}}

\newcommand{\threshold}{\delta}

\newcommand{\noise}{\sqrt{2\step}\rvg}

\newcommand{\conductance}{\Phi} 
\newcommand{\res}{s}
\newcommand{\sconductance}{\conductance_\res}
\newcommand{\truncballres}{\truncball_\res}

\newcommand{\warmparam}{\beta}

\newcommand{\MWARM}{\ensuremath{\mathcal{P}_\warmparam(\Target)}}

\newcommand{\UNICON}{\ensuremath{c}}
\newcommand{\unicontwo}{\ensuremath{c'}}

\newcommand{\lovtv}{\ensuremath{\rho}}

\newcommand{\lovdis}{\ensuremath{\Delta}}

\newcommand{\tvnorm}[1]{\ensuremath{\| #1\|_{\mbox{\tiny{TV}}}}}

\newcommand{\kldiv}[2]{\ensuremath{\operatorname{KL}(#1 \Vert #2)}}

\newcommand{\myinitial}{\ensuremath{\pi^0}}

\newcommand{\mytrans}{\ensuremath{p}}

\newcommand{\transition}{\mathcal{T}}

\newcommand{\defn}{:=}
\newcommand{\rdefn}{=:}

\newcommand{\tmix}{t_\text{mix}}

\newcommand{\convex}{\ensuremath{\mathcal{K}}}

\newcommand{\condition}{\ensuremath{\kappa}}

\newcommand{\myset}{\ensuremath{\mathcal{X}}}

\newlength{\widebarargwidth}
\newlength{\widebarargheight}
\newlength{\widebarargdepth}

\makeatletter
\long\def\@makecaption#1#2{
        \vskip 0.8ex
        \setbox\@tempboxa\hbox{\small {\bf #1:} #2}
        \parindent 1.5em  
        \dimen0=\hsize
        \advance\dimen0 by -3em
        \ifdim \wd\@tempboxa >\dimen0
                \hbox to \hsize{
                        \parindent 0em
                        \hfil
                        \parbox{\dimen0}{\def\baselinestretch{0.96}\small
                                {\bf #1.} #2
                                }
                        \hfil}
        \else \hbox to \hsize{\hfil \box\@tempboxa \hfil}
        \fi
        }
\makeatother

\long\def\comment#1{}

\newcommand{\matsnorm}[2]{|\!|\!| #1 | \! | \!|_{{#2}}}
\newcommand{\vecnorm}[2]{\left\| #1\right\|_{#2}}

\newcommand{\enorm}[1]{\vecnorm{#1}{2}}

\newcommand{\opnorm}[1]{\ensuremath{\matsnorm{#1}{\tiny{\mbox{op}}}}}

\newcommand{\inprod}[2]{\ensuremath{\langle #1 , \, #2 \rangle}}

\newcommand{\Exs}{\ensuremath{{\mathbb{E}}}}
\newcommand{\Prob}{\ensuremath{{\mathbb{P}}}}

\newcommand{\numobs}{\ensuremath{n}}
\newcommand{\usedim}{\ensuremath{d}}
\newcommand{\dims}{\usedim}
\newcommand{\obs}{\numobs}
\newcommand{\sampleobs}{\ensuremath{N}}

\newcommand{\NORMAL}{\ensuremath{\mathcal{N}}}

\newcommand{\xstar}{\ensuremath{x^\star}}
\newcommand{\xbar}{\ensuremath{\bar{x}}}

\newcommand{\xtilde}{\ensuremath{\tilde{x}}}
\newcommand{\initialtilde}{\tilde{\mu}}
\newcommand{\modetolerance}{\epsilon}

\newcommand{\yvec}{\ensuremath{Y}}
\newcommand{\Xmat}{\ensuremath{X}}

\newcommand{\widgraph}[2]{\includegraphics[keepaspectratio,width=#1]{#2}}

\newcommand{\Ind}{\ensuremath{\mathbb{I}}}

\newcommand{\real}{\ensuremath{\mathbb{R}}}
\newcommand{\realdim}{\ensuremath{\real^\usedim}}

\newcommand{\brackets}[1]{\left[ #1 \right]}
\newcommand{\parenth}[1]{\left( #1 \right)}

\newcommand{\biggparenth}[1]{\bigg( #1 \bigg)}
\newcommand{\braces}[1]{\left\{ #1 \right \}}
\newcommand{\abss}[1]{\left| #1 \right |}

\newcommand{\tp}{^\top}

\newcommand{\gaussiandistribution}{\mathcal{G}}
\newcommand{\order}[1]{\ensuremath{\mathcal{O}\parenth{#1}}}

\newcommand{\calo}{\mathcal{O}}

\usepackage{lastpage}

\jmlrheading{20}{2019}{1-\pageref{LastPage}}{4/19; Revised
11/19}{12/19}{19-306}{Dwivedi,
Chen, Wainwright and Yu}
\ShortHeadings{Metropolis-Hastings algorithms are
        fast}{Dwivedi, Chen, Wainwright and Yu}

\begin{document}


\title{Log-concave sampling:\\ Metropolis-Hastings algorithms are
        fast}

\author{\name Raaz Dwivedi$^{\ast, \dagger}$ \email raaz.rsk@berkeley.edu\\
    \name Yuansi Chen$^{\ast, \diamondsuit}$ \email yuansi.chen@berkeley.edu \\
      \name Martin J. Wainwright$^{\diamondsuit, \dagger, \ddagger}$ \email wainwrig@berkeley.edu\\
      \name Bin Yu$^{\diamondsuit,\dagger}$ \email binyu@berkeley.edu\\
        \addr Department of Statistics$^{\diamondsuit}$ \\
        Department of Electrical Engineering and Computer Sciences$^{\dagger}$ \\
        University of California, Berkeley \\
        Voleon Group$^{\ddagger}$, Berkeley
       }

\editor{Suvrit Sra}

\maketitle

\begin{abstract}
We study the problem of sampling from a strongly log-concave density
supported on $\mathbb{R}^d$, and prove a non-asymptotic upper bound on
the mixing time of the Metropolis-adjusted Langevin algorithm (MALA).
The method draws samples by simulating a Markov chain obtained from
the discretization of an appropriate Langevin diffusion, combined with
an accept-reject step.  Relative to known guarantees for the
unadjusted Langevin algorithm (ULA), our bounds show that the use of
an accept-reject step in MALA leads to an exponentially improved
dependence on the error-tolerance.  Concretely, in order to obtain
samples with TV error at most $\delta$ for a density with condition
number $\kappa$, we show that MALA requires $\mathcal{O} \big(\kappa d
\log(1/\delta) \big)$ steps from a warm start, as compared to the
$\mathcal{O} \big(\kappa^2 d/\delta^2 \big)$ steps established in past
work on ULA.  We also demonstrate the gains of a modified version of
MALA over ULA for weakly log-concave densities.  Furthermore, we
derive mixing time bounds for the Metropolized random walk (MRW) and
obtain $\mathcal{O}(\kappa)$ mixing time slower than MALA.  We provide
numerical examples that support our theoretical findings, and
demonstrate the benefits of Metropolis-Hastings adjustment for
Langevin-type sampling algorithms.
\end{abstract}


{\let\thefootnote\relax\footnotetext{*Raaz Dwivedi and Yuansi Chen contributed
equally to this work.}}



\section{Introduction}
\label{sec:introduction}

Drawing samples from a known distribution is a core computational
challenge common to many disciplines, with applications in statistics,
probability, operations research, and other areas involving stochastic
models. In statistics, these methods are useful for both estimation
and inference. Under the frequentist inference framework, samples drawn from
a suitable distribution can form confidence intervals for a point
estimate, such as those obtained by maximum likelihood.  Sampling
procedures are also standard in the Bayesian setting, used for
exploring posterior distributions, obtaining credible intervals, and
solving inverse problems.  Estimating the mean, posterior mean in a
Bayesian setting, expectations of desired quantities, probabilities of
rare events and volumes of particular sets are settings in which Monte
Carlo estimates are commonly used.

Recent decades have witnessed great success of Markov Chain Monte
Carlo (MCMC) algorithms in generating random samples; for
instance, see the handbook by~\cite{brooks2011handbook} and references
therein. In a broad sense, these methods are based on two steps.  The
first step is to construct a Markov chain whose stationary
distribution is either equal to the target distribution or close to it
in a suitable metric.  Given this chain, the second step is to draw
samples by simulating the chain for a certain number of steps.

Many algorithms have been proposed and studied for sampling from probability distributions with a density on a continuous state space.
Two broad categories of these methods are \emph{zeroth-order methods} and
\emph{first-order methods}. On one hand, a zeroth-order method is
based on querying the density of the distribution (up to a
proportionality constant) at a point in each iteration.  By contrast,
a first-order method makes use of additional gradient information about the
density.  A few popular examples of zeroth-order algorithms include
Metropolized random walk
(MRW)~\citep{mengersen1996rates,roberts1996geometric}, Ball
Walk~\citep{lovasz1990ballwalk,dyer1991random,lovasz1993random} and the
Hit-and-run algorithm~\citep{belisle1993hit, kannan1995isoperimetric,
  lovasz1999hit,lovasz2006hit, lovasz2007geometry}.  A number of
first-order methods are based on the Langevin diffusion.  Algorithms
related to the Langevin diffusion include the Metropolis adjusted
Langevin Algorithm (MALA)~\citep{roberts1996exponential,
  roberts2002langevin,bou2012nonasymptotic}, the unadjusted Langevin
algorithm (ULA)~\citep{parisi1981correlation,
  grenander1994representations,
  roberts1996exponential,dalalyan2016theoretical}, underdamped (kinetic)
Langevin MCMC~\citep{cheng2017underdamped,eberle2019couplings}, Riemannian
MALA~\citep{xifara2014langevin},
Proximal-MALA~\citep{pereyra2016proximal, durmus2016efficient},
Metropolis adjusted Langevin truncated
algorithm~\citep{roberts1996exponential}, Hamiltonian Monte
Carlo~\citep{neal2011mcmc} and Projected ULA~\citep{bubeck2015sampling}.
There is now a rich body of work on these methods, and we do not
attempt to provide a comprehensive summary in this paper.  More
details can be found in the survey by~\cite{roberts2004general}, which
covers MCMC algorithms for general distributions, and the
survey by~\cite{vempala2005geometric}, which focuses on random walks for
compactly supported distributions.

In this paper, we study sampling algorithms for sampling from a
log-concave distribution $\Target$ equipped with a density.  The
density of log-concave distribution can be written in the form
\begin{align}
\label{eq:target}
  \target\parenth{x} =
  \frac{e^{-\myfun(x)}}{\displaystyle\int_{\realdim} e^{-\myfun(y)}
    dy} \quad \mbox{for all $x \in \realdim$,}
\end{align}
where $\myfun$ is a convex function on $\realdim$.  Up to an additive
constant, the function $-\myfun$ corresponds to the log-likelihood
defined by the density.  Standard examples of log-concave
distributions include the normal distribution, exponential
distribution and Laplace distribution.

Some recent work has provided non-asymptotic bounds on the mixing
times of Langevin type algorithms for sampling from a log-concave
density.  The mixing time corresponds to the number of steps, as
A function of both the problem dimension $d$ and the error tolerance
$\threshold$, to obtain a sample from a distribution that is
$\threshold$-close to the target distribution in total variation
distance or other distribution distances. It is known that both the ULA
updates~\cite[see,
  e.g.,][]{dalalyan2016theoretical,
  durmus2016high,cheng2017convergence} as well as underdamped Langevin
MCMC~\citep{cheng2017underdamped} have mixing times that scale
polynomially in the dimension $\dims$, as well the inverse of the
error tolerance $1/\threshold$.

Both the ULA and underdamped-Langevin MCMC methods are based on
evaluations of the gradient~$\gradf$, along with the addition of
Gaussian noise.  \cite{durmus2016high} shows that for an appropriate
decaying step size schedule, the ULA algorithm converges to the
correct stationary distribution. However, their results, albeit
non-asymptotic, are hard to quantify. In the sequel, we limit our
discussion to Langevin algorithms based on constant step sizes, for
which there are a number of explicit quantitative bounds on the mixing
time.  When one uses a fixed step size for these algorithms, an
important issue is that the resulting random walks are asymptotically
biased: due to the lack of Metropolis-Hastings correction step, the
algorithms \emph{will not} converge to the stationary distribution if
run for a large number of steps.  Furthermore, if the step size is not
chosen carefully the chains may become
transient~\citep{roberts1996exponential}.  Thus, the typical theory is
based on running such a chain for a pre-specified number of steps,
depending on the tolerance, dimension and other problem parameters.

In contrast, the Metropolis-Hastings step that underlies the MALA
algorithm ensures that the resulting random walk has the correct
stationary distribution.  \cite{roberts1996exponential} derived
sufficient conditions for exponential convergence of the Langevin
diffusion and its discretizations, with and without
Metropolis-adjustment.  However, they considered the distributions
with $\myfun(x) = \enorm{x}^\alpha$ and proved geometric convergence
of ULA and MALA under some specific conditions.  In a more general
setting, \cite{bou2012nonasymptotic} derived non-asymptotic mixing
time bounds for MALA.  However, all these bounds are non-explicit, and
so makes it difficult to extract an explicit dependency in terms of
the dimension $\usedim$ and error tolerance~$\threshold$.  A precise
characterization of this dependence is needed if one wants to make
quantitative comparisons with other algorithms, including ULA and
other Langevin-type schemes.  \cite{eberle2014error} derived mixing
time bounds for MALA albeit in a more restricted setting compared to
the one considered in this paper.  In particular, Eberle's convergence
guarantees are in terms of a modified Wasserstein distance, truncated
so as to be upper bounded by a constant, for a subset of strongly
concave measures which are four-times continuously differentiable and
satisfy certain bounds on the derivatives up to order four.  With this
context, one of the main contributions of our paper is to provide an
explicit upper bound on the mixing time bounds in total variation
distance of the MALA algorithm for general log-concave distributions.


\paragraph{Our contributions:} This paper contains two main results,
both having to do with the upper bounds on mixing times of MCMC
methods for sampling. As described above, our first and primary
contribution is an explicit analysis of the mixing time of Metropolis
adjusted Langevin Algorithm (MALA).  A second contribution is to use
similar techniques to analyze a zeroth-order method called
Metropolized random walk (MRW) and derive an explicit non-asymptotic
mixing time bound for it.  Unlike the ULA, these methods make use of
the Metropolis-Hastings accept-reject step and consequently converge
to the target distributions in the limit of infinite steps.  Here we
provide explicit non-asymptotic mixing time bounds for MALA and MRW,
thereby showing that MALA converges significantly faster than ULA, at
least in terms of the best known upper bounds on their respective
mixing times.\footnote{Throughout the paper, we make comparisons
  between sampling algorithms based on known upper bounds on
  respective mixing times; obtaining matching lower bounds is also of
  interest.}  In particular, we show that if the density is strongly
log-concave and smooth, the $\threshold$-mixing time for MALA scales
as $\condition \dims\log(1/\threshold)$ which is significantly faster
than ULA's convergence rate of order
$\condition^2\dims/\threshold^2$. On the other hand, Moreover, we also
show that MRW mixes $\order{\condition}$ slower when compared to
MALA. Furthermore, if the density is weakly log-concave, we show that
(a modified version of) MALA converges in $\calo\parenth{
  \dims^2/\threshold^{1.5} }$ time in comparison to the
$\calo\parenth{\dims^3/\threshold^{4}}$ mixing time for ULA. As
alluded to earlier, such a speed-up for MALA is possible since we can
choose a large step size for it which in turn is possible due to its
unbiasedness in the limit of infinite steps. In contrast, for ULA the
step-size has to be small enough to control the bias of the
distribution of the ULA iterates in the limit of infinite steps,
leading to a relative slow down when compared to MALA.


The remainder of the paper is organized as follows.  In
Section~\ref{sec:background_on_the_langevin_monte_carlo_algorithm}, we
provide background on a suite of MCMC sampling algorithms based on the
Langevin diffusion.  Section~\ref{sec:results} is devoted to the statement
of our mixing time bounds for MALA and MRW, along with a discussion of
some consequences of these results.  Section~\ref{sec:simulations} is
devoted to some numerical experiments to illustrate our guarantees.
We provide the proofs of our main results in Section~\ref{sec:proofs},
with certain more technical arguments deferred to the appendices.  We
conclude with a discussion in Section~\ref{sec:discussion}.


\paragraph{Notation:}
\label{par:notation}
For two sequences $a_\tvscalar$ and $b_\tvscalar$ indexed by a scalar
$\tvscalar \in I \subseteq \real$, we say that $a_\tvscalar =
\order{b_\tvscalar}$ if there exists a universal constant $c > 0 $
such that $a_\tvscalar \leq c b_\tvscalar$ for all $\tvscalar \in I$.
The Euclidean norm of a vector $x \in \realdim$ is denoted by
$\vecnorm{x}{2}$.  The Euclidean ball with center $x$ and radius $r$
is denoted by $\mathbb{B}(x, r)$.  For two distributions $\proposal_1$
and $\proposal_2$ defined on the space $(\realdim, \mathcal{B}(\realdim))$ where
$\mathcal{B}(\realdim)$ denotes the Borel-sigma algebra on $\realdim$,
we use $\tvnorm{\proposal_1 - \proposal_2}$ to denote their total variation
distance given by
\begin{align*}
  \tvnorm{\proposal_1 - \proposal_2} = \sup_{A \in \mathcal{B}(\realdim)} \abss{\proposal_1(A) - \proposal_2(A)}.
\end{align*}
Furthermore, $\kldiv{\proposal_1}{\proposal_2}$ denotes their
Kullback-Leibler (KL) divergence.  We use $\Target$ to denote the
target distribution with density $\target$.


\section{Background and problem set-up}
\label{sec:background_on_the_langevin_monte_carlo_algorithm}

In this section, we briefly describe the general framework for MCMC algorithms and review the
rates of convergence of existing random walks for log-concave
distributions.

\subsection{Markov chains and mixing}

\label{sub:markov_chains_and_mixing}

Here we consider the task of drawing samples from a \emph{target
  distribution} $\Target$ with density $\target$.  A popular class of
methods are based on setting up of an irreducible and aperiodic
discrete-time Markov chain whose stationary distribution is equal to
or close to the target distribution $\Target$ in certain metric, e.g.,
total variation (TV) norm.  In order to obtain a $\threshold$-accurate
sample, one simulates the Markov chain for a certain number of steps
$k$, as determined by a mixing time analysis.

Going forward, we assume familiarity of the reader with a basic
background in Markov chains. See, e.g., Chapters 9-12 in the standard reference textbook \citep{meyn2012markov} for a rigorous and detailed introduction. 
For a more rapid introduction to the basics of continuous state-space Markov chains, we refer the reader to the expository paper~\citep{diaconis1997markov} or Section 1 and 2 of the
papers~\citep{lovasz1993random,vempala2005geometric}.  We now briefly
describe a certain class of Markov chains that are of
\emph{Metropolis-Hastings
  type}~\citep{metropolis1953equation,hastings1970monte}.  See the
books~\citep{robert2004monte,brooks2011handbook} and references
therein for further background on these chains.

Starting at a given initial density $\myinitial$ over $\real^\usedim$,
any such Markov chain is simulated in two steps: (1) a proposal step,
and (2) an accept-reject step.
For the proposal step, we make use of a \emph{proposal
function} \mbox{$\mytrans: \real^\usedim \times \real^\usedim \in
\real_+$,} where $\mytrans(x, \cdot)$ is a density function for each
$x \in \real^\usedim$. At each iteration, given a current state $x \in
\real^\usedim$ of the chain, the algorithm proposes a new vector
$z \in \real^\usedim$ by sampling from the proposal density
$\mytrans(x, \cdot)$.
In the second step, the algorithm accepts $z \in \real^\usedim$ as the
new state of the Markov chain with probability
\begin{align}
  \label{EqnMHCorrection}
  \alpha(x, z) \defn \displaystyle\min\braces{1, \;
    \frac{\target(z)\density(z,
      x)}{\target(x)\density(x, z)}}.
\end{align}
Otherwise, with probability equal to $1-\alpha(x, z)$, the chain stays
at $x$.  Consequently, the overall transition kernel $\transdensity$ for the Markov
chain is defined by the function
\begin{align*}
\transdensity(x, z) & \defn \density(x, z) \alpha(x, z) \qquad \mbox{for
$z \neq x$,}
\end{align*}
and a probability mass at $x$ with weight $1- \int_\myset
\transdensity(x,z) dz$.  The purpose of the Metropolis-Hastings
correction~\eqref{EqnMHCorrection} is to ensure that
the target density
$\target$ is stationary for the Markov chain.

Overall, this set-up defines an operator $\transition_\density$ on the
space of probability distributions: given the distribution $\mu_k$ of
the chain at time $k$, the distribution at time $k+1$ is given by
$\transition_\density(\mu_k)$.  In fact, with the starting
distribution $\initial$, the distribution of the chain at $k$th step
is given by $\transition_\density^k(\initial)$.  Note that in this
notation, the transition distribution at any state $x$ is given by
$\transition_\density(\diracdelta_x)$ where $\diracdelta_x$ denotes
the dirac-delta distribution \mbox{at $x$}.  Our assumptions and
set-up ensure that the chain converges to target distribution in the
limit of infinite steps, i.e., $\lim_{k \rightarrow \infty}
\transition_\density^k(\initial) = \Target$. However, a more practical
notion of convergence is how many steps of the chain suffice to ensure
that the distribution of the chain is ``close'' to the target
$\Target$.  In order to quantify the closeness, for a given tolerance
parameter $\threshold \in (0,1)$ and initial distribution $\initial$,
we define the \emph{$\threshold$-mixing time} as
\begin{align}
\label{EqnDefnMixingTime}
\tmix(\threshold; \initial) & \defn \min \Big \{ k \; \vert \;
\tvnorm{\transition_\density^k(\initial) - \Target} \leq \threshold
\Big \},
  \end{align}
corresponding to the minimum number of steps that the chain takes to
reach within $\threshold$ in TV-norm of the target distribution, given
that it starts with distribution $\initial$.


\subsection{Sampling from log-concave distributions}
\label{sub:sampling_from_log_concave_distributions}

Given the set-up in the previous subsection, we now describe several
algorithms for sampling from log-concave distributions.  Let
$\proposal_x$ denote the proposal distribution at $x$ corresponding to
the proposal density $\density(x, \cdot)$.  Possible choices of this
proposal function include:
\begin{itemize}
  \item Independence sampler: the proposal distribution does not
    depend on the current state of the chain, e.g., rejection sampling
    or when $ \proposal_x = \NORMAL(0, \Sigma)$, where $\Sigma$ is a
    hyper-parameter.
  \item Random walk: the proposal function satisfies $\density(x, y) =
    q(y-x)$ for some probability density $q$, e.g., when
    \mbox{$\proposal_x = \NORMAL(x, 2\step
      \Ind_\dims)$} where $\step$ is a hyper-parameter.
  \item Langevin algorithm: the proposal distribution is shaped
  according to the target distribution and is given by
  $\proposal_x = \NORMAL(x-\step \gradf(x), 2\step \Ind_\dims)$,
  where $\step$ is chosen suitably.
      \item Symmetric Metropolis algorithms: the proposal function
        $\density$ satisfies $\density(x, y) = \density(y, x)$.  Some
        examples are Ball Walk~\citep{frieze1994sampling}, and
        Hit-and-run~\citep{lovasz1999hit}.
\end{itemize}
Naturally, the convergence rate of these algorithms depends on the
properties of the target density $\target$, and the degree to which
the proposal function $\density$ is suited for the task at hand.  A
key difference between Langevin algorithm and other algorithms is that
the former makes use of first-order (gradient) information about the
target distribution $\Target$.  We now briefly discuss the existing
theoretical results about the convergence rate of different MCMC
algorithms.  Several results on MCMC algorithms have focused on on
establishing behavior and convergence of these sampling algorithms in
an asymptotic or a non-explicit sense, e.g., geometric and uniform
ergodicity, asymptotic variance, and central limit theorems.  For more
details, we refer the readers to the papers
by~\cite{talay1990expansion,meyn1994computable,
  roberts1996geometric,roberts1996exponential,jarner2000geometric,
  roberts2001optimal,roberts2002langevin,pillai2012optimal,
  roberts2014complexity}, the survey by~\cite{roberts2004general} and
the references therein.  Such results, albeit helpful for gaining
insight, do not provide user-friendly rates of convergence.  In other
words, from these results, it is not easy to determine the
computational complexity of various MCMC algorithms as a function of
the problem dimension $\dims$ and desired accuracy $\threshold$.
Explicit non-asymptotic convergence bounds, which provide useful
information for practice, are the focus of this work.  We discuss
the results of such type and the Langevin algorithm in more detail in
\mbox{Section~\ref{ssub:langevin_diffusion_and_algorithms}}.  We begin
with the Metropolized random walk.

\subsubsection{Metropolized random walk}
\label{ssub:metropolized_random_walk}

\cite{roberts1996geometric} established sufficient conditions on the
proposal function $\density$ and the target distribution $\Target$ for
the geometric convergence of several random walk Metropolis-Hastings
algorithms.  In Section~\ref{sec:results}, we establish non-asymptotic
convergence rate for the Metropolized random walk, which is based on
Gaussian proposals.  That is when the chain is at state $x_k$, a
proposal is drawn as follows
\begin{align}
\label{eq:MRW}
  z_{k+1} = x_k + \sqrt{2 \step} \; \rvg_{k+1},
\end{align}
where the noise term $\rvg_{k+1} \sim \NORMAL(0, \Ind_\dims)$ is
independent of all past iterates.  The chain then makes the transition
according to an accept-reject step with respect to $\Target$.  Since
the proposal distribution is symmetric, this step can be described as
\begin{align*}
  x_{k+1} = \begin{cases} z_{k+1} \quad &\text{with probability}
    \min\braces{1,
      \displaystyle\frac{\target(z_{k+1})}{\target(x_{k})}}\\ x_k
    \quad &\text{otherwise}.
  \end{cases}
\end{align*}
This sampling algorithm is an instance of a zeroth-order method since
it makes use of only the function values of the density $\target$.  We
refer to this algorithm as MRW in the sequel.  It is easy to see that
the chain has a positive density of jumping from any state $x$ to $y$ in
$\realdim$ and hence is strongly $\Target$-irreducible and
aperiodic. Consequently, Theorem~1 by~\cite{diaconis1997markov}
implies that the chain has a unique stationary distribution $\Target$
and converges to as the number of steps increases to infinity.  Note
that this algorithm has also been referred to as Random walk
Metropolized (RWM) and Random walk Metropolis-Hastings (RWMH) in the
literature.


\subsubsection{Langevin diffusion and related sampling algorithms}
\label{ssub:langevin_diffusion_and_algorithms}

Langevin-type algorithms are based on Langevin diffusion, a stochastic
process whose evolution is characterized by the stochastic
differential equation (SDE):
\begin{align}
\label{eq:Langevin_sde}
  d X_t = -\gradf(X_t)dt + \sqrt{2} \;  d W_t,
\end{align}
where $\braces{W_t \mid t \geq 0}$ is the standard Brownian motion on
$\realdim$.  Under fairly mild conditions on $\myfun$, it is known
that the diffusion~\eqref{eq:Langevin_sde} has a unique strong
solution $\braces{X_t, t\geq 0}$ that is a Markov
process~\citep{roberts1996exponential,meyn2012markov}. Furthermore, it
can be shown that the distribution of $X_t$ converges as $t
\rightarrow +\infty$ to the invariant distribution~$\Target$
characterized by the density $\target(x) \propto \exp(-\myfun(x))$.


\paragraph{Unadjusted Langevin algorithm}
\label{par:unadjusted_langevin_algorithm}

A natural way to simulate the Langevin
diffusion~\eqref{eq:Langevin_sde} is to consider its forward Euler
discretization, given by
\begin{align}
\label{eq:ULA}
  x_{k+1} = x_k - \step \gradf(x_k) + \sqrt{2\step} \rvg_{k+1},
\end{align}
where the driving noise \mbox{$\rvg_{k+1} \sim \NORMAL(0,
  \Ind_\dims)$} is drawn independently at each time step.  The use of
iterates defined by equation~\eqref{eq:ULA} can be traced back at
least to~\cite{parisi1981correlation} for computing correlations; this
use was noted by Besag in his commentary on the paper
by~\cite{grenander1994representations}.

However, even when the SDE is well behaved, the iterates defined by
this discretization have mixed behavior.  For sufficiently large step
sizes $\step$, the distribution of the iterates defined by
equation~\eqref{eq:ULA} converges to a stationary distribution that is
no longer equal to $\Target$.  In fact, \cite{roberts1996exponential} showed that if the step size
$\step$ is not chosen carefully, then the Markov chain defined by
equation~\eqref{eq:ULA} can become transient and have no~stationary
distribution.  However, in a series of recent works
\citep{dalalyan2016theoretical,durmus2016high,cheng2017convergence}, it
has been established that with a careful choice of step-size $\step$
and iteration count $K$, running the chain~\eqref{eq:ULA} for exactly
$K$ steps yields an iterate $x_K$ whose distribution is close to
$\Target$.  This more recent body of work provides non-asymptotic
bounds that explicitly quantify the rate of convergence for this
chain.  Note that the algorithm~\eqref{eq:ULA} does not belong to the
class of Metropolis-Hastings algorithms since it does not involve an
accept-reject step and does not have the target distribution~$\Target$
as its stationary distribution. Consequently, in the literature,
this algorithm is referred to as the \emph{unadjusted Langevin
  Algorithm}, or ULA for short.


\paragraph{Metropolis adjusted Langevin algorithm}
\label{par:metropolis_adjusted_langevin_algorithm}

An alternative approach to handling the discretization error is
to adopt
$\NORMAL(x_k-\step\gradf(x_k), 2\step\Ind_\dims)$ as the proposal
distribution, and perform the Metropolis-Hastings accept-reject step.
Doing so leads to the \emph{Metropolis-adjusted Langevin Algorithm},
or MALA for short. We describe the different steps of MALA
in Algorithm~\ref{algo:mala}.
As mentioned earlier, the Metropolis-Hastings
correction ensures that the distribution of the MALA iterates
$\braces{x_k}$ converges to the correct distribution $\Target$ as $k
\rightarrow \infty$. Indeed, since at each step the chain can reach any
state $x \in \realdim$, it is strongly $\Target$-irreducible and thereby
ergodic~\citep{meyn2012markov,diaconis1997markov}.

Both MALA and ULA are instances of first-order
sampling methods since they make use of both the function and the
gradient values of $\myfun$ at different points.  A natural question
is if employing the accept-reject step for the
discretization~\eqref{eq:ULA} provides any gain in the convergence
rate. Our analysis to follow answers this question in the affirmative.

\begin{algorithm}[h]
  \KwIn{Step size $\step$ and a
  sample $x_0$ from a starting distribution $\initial$}
  \KwOut{Sequence $x_1, x_2,\ldots$}
  \For{$i=0, 1, \ldots $}
  {%
      $z_{i+1} \sim \NORMAL(\displaystyle x_{i} - \step \gradf(x_{i}), 2\step
      \Ind_\usedim)$  \quad \% {\footnotesize{propose a new state}}\\
      $\alpha_{i+1} = \displaystyle\min \braces{1,
          \frac{\exp\parenth{-f(z_{i+1})-\enorm{x_{i}-z_{i+1}+\step\gradf(z_{i+1})}^2/4\step}}{\exp\parenth{-f(x_{i})-\enorm{z_{i+1}-x_i+\step\gradf(x_i)}^2/4\step}}
        }$\\ $U_{i+1} \sim U[0, 1]$\\ \lIf( \% {\footnotesize{accept the
    proposal}}){$U_{i+1} \leq \alpha_{i+1}$}{$ x_
    {i+1} \gets z_{i+1} $
        }
        \lElse(  \% {\footnotesize{reject the
    proposal}} ){$ x_{i+1} \gets
          x_{i} $} }
  \caption{Metropolis adjusted Langevin algorithm (MALA)}
  \label{algo:mala}
\end{algorithm}


\subsection{Problem set-up}
\label{sub:set_up}
We study MALA and MRW and contrast their performance with existing
algorithms for the case when the negative log density $f(x) \defn
-\log \target(x) $ is smooth and strongly convex.  A function $\myfun$
is said to be $L$-smooth if
\begin{subequations}
\begin{align}
\label{eq:smoothness}
      \myfun(y) - \myfun(x) - \gradf(x)\tp(y-x) & \leq
      \frac{\smoothness}{2} \enorm{x-y}^2  \quad \mbox{for all $x, y
        \in \realdim$.}
\end{align}
In the other direction, a convex function $\myfun$ is said to be
$\scparam$-strongly convex if
\footnote{ See Appendix~\ref{sec:some_basic_properties} for a
  statement of some well-known properties of smooth and strongly
  convex functions.}
\begin{align}
\label{eq:strongly_convex}
      \myfun(y) - \myfun(x) - \gradf(x)\tp(y-x) & \geq
      \frac{\scparam}{2} \enorm{x-y}^2 \quad \quad \mbox{for all $x, y
        \in \realdim$.}
\end{align}
\end{subequations}
The rates derived in this paper apply to log-concave\footnote{ While
  our techniques can yield sharper guarantees under more idealized
  assumptions, e.g., like when $f$ is Lipschitz (bounded gradients)
  and the target distribution satisfies an isoperimetry inequality,
  here we focus on deriving explicit guarantees with log-concave
  distributions.}  distributions~\eqref{eq:target} such that $\myfun$
is continuously differentiable on $\realdim$, and is both
$\smoothness$-smooth and $\scparam$-strongly convex.  For such a
function~$\myfun$, its condition number $\condition$ is defined as
$\condition \defn \smoothness/\scparam$.  We also refer to
$\condition$ as the condition number of the target distribution
$\Target$.  We summarize the mixing time bounds of several sampling
algorithms in Tables~\ref{tab:mixing_times} and
\ref{tab:mixing_times_feasible_start}, as a function of the
dimension~$\dims$, the error-tolerance~$\threshold$, and the condition
number $\condition$.  In Table~\ref{tab:mixing_times}, we state the
results when the chain has a warm-start defined below (refer to the
definition~\eqref{EqnDefnMstart}).
Table~\ref{tab:mixing_times_feasible_start} summarizes mixing time
bounds from a particular distribution~$\initialstar$.  Furthermore, in
Section~\ref{sub:nonstrongly_log_concave_densities} we discuss the
case when the $\myfun$ is smooth but not strongly convex and show that
a suitable adaptation of MALA has a faster mixing rate compared to ULA
for this case.

\begin{table}
    \centering
    \resizebox{0.89\textwidth}{!}{
    {
    \renewcommand{\arraystretch}{2.1}
    \begin{tabular}{ccc}
        \toprule
        \small {\bf Random walk} & {\bf Strongly log-concave}
        & {\bf Weakly log-concave}
        \\ \midrule
        ULA~\citep{cheng2017convergence}
        &$\mathcal{O}\parenth{\displaystyle\frac{\dims\condition^2\log((\log \warmparam)/\threshold)}{\threshold^2}}$
        & $\displaystyle\tilde{\mathcal{O}}\parenth{\frac{\dims\smoothness^{2}}{\threshold^{6}}}$
        \\[2mm]
        ULA~\citep{dalalyan2016theoretical}
        &$\mathcal{O}\parenth{\displaystyle\frac{\dims\condition^2 \log^2(\warmparam/\threshold)}{\threshold^2}}$
        & $\displaystyle\tilde{\mathcal{O}}\parenth{\frac{\dims^3\smoothness^{2}}{\threshold^{4}}}$
        \\[2mm]
        MRW (this work)
        &$\mathcal{O}\parenth{\displaystyle\dims \condition^2 \log\parenth{\frac\warmparam\threshold}}$
        & $\displaystyle\tilde{\mathcal{O}}\parenth{\frac{\dims^3\,\smoothness^{2}}{\threshold^{2}}}$
        \\[2mm]
        MALA (this work)
        &$\mathcal{O}\parenth{\displaystyle \max\braces{\dims\condition, \dims^{0.5}\condition^{1.5}}
                \log\parenth{\frac\warmparam\threshold}}$
        & $\displaystyle\tilde{\mathcal{O}}\parenth{\frac{\dims^2\, \smoothness^{1.5}}{\threshold^{1.5}}}$
        \\[2ex]
        \bottomrule
        \hline
    \end{tabular}
    }
    }
    \caption{Scalings of upper bounds on $\threshold$-mixing time for different random walks in
      $\realdim$ with target $\target \propto e^{-\myfun}$.
      In the second column, we consider smooth and strongly log-concave densities, and report the bounds from a $\warmparam$-warm start for densities such that
      $\scparam \Ind_\dims \preceq \hessf(x) \preceq
      \smoothness\Ind_\dims$ for any $x \in \realdim$ and use $\condition
      \defn \smoothness/\scparam$ to denote the condition number of the density.
      The big-O notation hides universal constants.
      We remark that the presented bounds for ULA in this column are not stated
      in the corresponding papers, and are derived by us, using their framework.
      In the last column, we summarize the scaling for weakly log-concave smooth densities: $0 \preceq \hessf(x) \preceq
      \smoothness\Ind_\dims$ for all $x \in \realdim$. For this case, the $\tilde\calo$ notation is used to track scaling only with respect to $\dims, \threshold$ and $\smoothness$ and ignore dependence on the starting distribution and a few other parameters.
      }
    \label{tab:mixing_times}
\end{table}

\begin{table}
    \centering
  \resizebox{0.85\textwidth}{!}{
    {\renewcommand{\arraystretch}{2.1}
    \begin{tabular}{ccc}
        \toprule
        \small {\bf Random walk} & {\bf Distribution $\initialstar$} & {\bf $\tmix
        (\threshold; \initialstar)$ } \\
          \midrule
        ULA~\citep{cheng2017convergence} &$\NORMAL(\xstar,
        \scparam^{-1}\Ind_\dims)$
        &$\displaystyle\mathcal{O}\parenth{\frac{\dims\condition^2\log(\dims\condition/\threshold)}{\threshold^2}}$
        \\[2mm]
        ULA~\citep{dalalyan2016theoretical} &$\NORMAL(\xstar,
        \smoothness^{-1}\Ind_\dims)$
        &$\displaystyle\mathcal{O}\parenth{\frac{(\dims^3+\dims\log^2(1/\threshold))\condition^2}{\threshold^2}}$
        \\[2mm]
        MRW (this work)
        &$\NORMAL(\xstar, \smoothness^{-1}\Ind_\dims)$
        &$\displaystyle\mathcal{O}\parenth{\dims^2 \condition^2 \log^{1.5}\parenth{
                \frac{\condition}{\threshold}}}$
        \\[2mm]
        MALA (this work) &$\NORMAL(\xstar, \smoothness^{-1}\Ind_\dims)$
        &$\displaystyle \mathcal{O}\parenth{\dims^{2} \condition
                \log\parenth{\frac{\condition}{\threshold}}}$
        \\[2ex] \bottomrule \hline
    \end{tabular}
    }
    }
    \caption{Scalings of upper bounds on $\threshold$-mixing time,
      from the starting distribution~$\initialstar$ given in column two,
      for different random walks in $\realdim$ with target $\target
      \propto e^{-\myfun}$ such that $\scparam \Ind_\dims \preceq
      \hessf(x) \preceq \smoothness\Ind_\dims$ for any $x \in
      \realdim$ and $\condition \defn \smoothness/\scparam$.
      Here $\xstar$ denotes
      the unique mode of the target density $\target$.
      }
    \label{tab:mixing_times_feasible_start}
\end{table}


\section{Main results}
\label{sec:results}

We now state our main results for mixing time bounds for MALA and MRW.
In our results, we use $\UNICON,\unicontwo$ to denote universal
positive constants.  Their values can change depending on the context,
but do not depend on the problem parameters in all cases.  In this
section, we begin by discussing the case of strongly log-concave
densities. We state results for MALA and MRW from a warm start in
Section~\ref{sub:mala}, and from certain feasible starting
distributions in Section~\ref{sub:non_warm_start}.
Section~\ref{sub:nonstrongly_log_concave_densities} is devoted to the
case of weakly log-concave densities.


\subsection{Mixing time bounds for warm start}
\label{sub:mala}

In the analysis of Markov chains, it is convenient to have a rough
measure of the distance between the initial distribution $\initial$
and the stationary distribution.  As in past work on the problem, we
adopt the following notion of \emph{warmness}:
For a finite scalar $\warmparam > 0$, the initial distribution $\initial$
is said to be $\warmparam$-warm with respect to the stationary
distribution $\Target$ if
\begin{align}
  \label{EqnDefnMstart}
  \sup_{\set} \parenth{\frac{\initial(\set)}{\Target(\set)}} \leq
  \warmparam,
  \end{align}
where the supremum is taken over all measurable sets $\set$.
In parts of our work, we provide bounds on the quantity
\begin{align*}
  \tmix(\threshold; \warmparam)
  = \sup\limits_{\initial \in \MWARM} \tmix(\threshold; \initial)
 \end{align*}
where $\MWARM$ denotes the set of all distributions that are
$\warmparam$-warm with respect to $\Target$.  Naturally, as the value
of $\warmparam$ decreases, the task of generating samples from the
target distribution becomes easier.\footnote{For instance, $\warmparam=1$
implies that the chain starts at the stationary distribution and has
already mixed.}
However, access to a good ``warm'' distribution (small $\warmparam$)
may not be feasible for many applications, and
thus deriving bounds on mixing time of the Markov chain from non-warm
starts is also desirable.  Consequently, in the sequel, we also
provide practical initialization methods and polynomial-time mixing
time guarantees from such starts.

Our mixing time bounds involve the functions $\radius$ and $\stepfun$
given by
\begin{subequations}
\begin{align}
    \radius(\res) &= 2 + 2 \cdot \max\braces{ \frac{1}{\dims^{0.25}}
      \log^{0.25}\parenth{\frac{1}{\res}}, \frac{1}{\dims^{0.5}}
      \log^{0.5}\parenth{\frac{1}{\res}}}
    \label{eq:defn_r},\quad \text{and}\\
    \label{eq:defn_h}
    \stepfun\parenth{\res} &=
    \min\braces{\frac{\sqrt{\scparam}}{\radius(\res) \cdot
        \smoothness\sqrt{\dims\smoothness}},
      \ \frac{1}{\smoothness\dims}} \qquad \mbox{for $s \in \big (0,
      \frac{1}{2} \big)$.}
\end{align}
\end{subequations}
We use $\transition_{\tagmala(\step)}$ to denote the transition
operator on probability distributions induced by one step of MALA.
With this notation, we have the following mixing time bound for the
MALA algorithm for a strongly-log concave measure from a warm start.
\begin{theorem}
    \label{thm:mala_mixing}
    For any $\warmparam$-warm initial distribution $\initial$ and any
    error tolerance $\threshold \in (0, 1]$, the Metropolis adjusted
      Langevin algorithm with step size $\step =
      \UNICON\,\stepfun(\threshold/(2\warmparam))$ satisfies the bound
      $\tvnorm{\transition_{\tagmala(\step)}^k(\initial)-\Target} \leq
      \threshold$ for all iteration numbers
 \begin{align}
   k \geq \unicontwo
   \log\parenth{\frac{2\warmparam}{\threshold}} \max \Biggr \{
   \dims \condition,
   \ \dims^{0.5}\condition^{1.5} \radius
   \biggr(\frac{\threshold}{2\warmparam} \biggr) \Biggr \},
    \end{align}
    where $\UNICON, \unicontwo$ denote universal constants.
\end{theorem}
\noindent See Section~\ref{sub:proof_of_theorem_thm:mala_mixing} for the
proof.\\

Note that $\radius(\res) \leq 4$ for $\res \geq e^{-\dims}$ and thus
we can treat $\radius(\threshold/2\warmparam)$ as small constant for
most interesting values of $\threshold$ if the warmness parameter
$\warmparam$ is not too large.  Consequently, we can run MALA with a
fixed step size~$\step$ for a large range of
error-tolerance~$\threshold$.  Treating the function $\radius$ as a
constant, we obtain that if $\condition = o(\dims)$, the mixing time
of MALA scales as $\order{\dims \condition\log(1/\threshold)}$.  Note
that the dependence on the tolerance $\threshold$ is exponentially
better than the $\order{\dims \condition^2
  \log^2(1/\threshold)/\threshold^2}$ mixing time of ULA, and has
better dependence on $\condition$ while still maintaining linear
dependence on $\dims$.  In fact, for any setting of $\condition,
\dims$ and $\threshold$, MALA always has a better mixing time bound
compared to ULA.  A limitation of our analysis is that the constant
$\unicontwo$ is not small.  However, we demonstrate in
Section~\ref{sec:simulations} that in practice small constants provide
performance that matches the scalings suggested by our theoretical
bounds.

Let $\transition_{\tagmrw(\step)}$ denote the transition operator on the
space of probability distributions induced by one step
of MRW. We now state the convergence rate for Metropolized random walk
for strongly-log concave density.
\begin{theorem}
    \label{thm:mrw_mixing}
For any $\warmparam$-warm initial distribution $\initial$ and any $\threshold
\in (0, 1]$, the Metropolized random walk with step size $\step =
  \frac{\UNICON\scparam}{\dims\smoothness^2
    \radius(\threshold/2\warmparam)} $ satisfies
    \begin{align}
        \tvnorm{\transition_{\tagmrw(\step)}^k(\initial)-\Target} \leq
        \threshold \quad \text{for all } \quad k \geq \unicontwo\,
        \dims\condition^2 \radius \bigg(
        \frac{\threshold}{2\warmparam} \bigg) \;
        \log\parenth{\frac{2{\warmparam}}{\threshold}},
    \end{align}
    where $\UNICON, \unicontwo$ denote universal constants.
\end{theorem}
\noindent See Section~\ref{sub:proof_of_theorem_thm:mrw_mixing} for
the proof.\\

Again treating $\radius(\threshold/2\warmparam)$ as a small constant,
we find that the mixing time of MRW scales as
$\order{\dims\condition^2\log(1/\threshold)}$ which has an exponential
factor in $\threshold$ better than ULA.
Compared to the mixing time bound for MALA, the bound in
Theorem~\ref{thm:mrw_mixing} has an extra factor of $\calo(
\condition)$.  While such a factor is conceivable given that MALA's
proposal distribution uses first-order information about the target
distribution and MRW uses only the function values, it would be
interesting to determine if this gap can be improved.
See Section~\ref{sec:discussion} for a discussion on possible future work
in this direction.


\subsection{Mixing time bounds for a feasible start}
\label{sub:non_warm_start}

In many cases, a good warm start may not be available.  Consequently,
mixing time bounds from a feasible starting distribution can be useful
in practice.  Letting $\xstar$ denote the unique mode of the
target distribution $\Target$, we claim that the distribution
\mbox{$\initialstar = \NORMAL(\xstar, \smoothness^{-1}\Ind_\dims)$} is
one such choice.  Recalling the condition number $\condition =
\smoothness/\scparam$, we claim that the warmness parameter for
$\initialstar$ can be bounded as follows:
    \begin{align}
    \label{eq:warm_star}
        \sup_{\set}\frac{\initialstar(\set)}{\Target(\set)}
        \leq \condition^{\dims/2}
        = \warmparam_\star,
    \end{align}
where the supremum is taken over all measurable sets $\set$.  When the
gradient $\gradf$ is available, finding $\xstar$ comes at nominal
additional cost: in particular, standard optimization algorithms such
as gradient descent be used to compute a $\delta$-approximation of
$\xstar$ in $\order{\condition \log(1/\delta)}$ steps (e.g., see the
monograph by~\citealt{bubeck2015convex}). Also refer to
Section~\ref{ssub:starting_with_inexact_parameters} for more details
when we have inexact parameters.

Assuming claim~\eqref{eq:warm_star} for the moment, we now provide
mixing time bounds for MALA and MRW with $\initialstar$ as the
starting distribution.  For any threshold $\threshold \in (0, 1]$, we
  define the step sizes $\step_1 =
  \unicontwo\stepfun(\threshold/2\warmparam_\star)$ and $\step_2 =
  \frac{\unicontwo\scparam}{\dims \smoothness^2 \cdot
    \radius(\threshold/2\warmparam_\star)}$, where the
  function~$\stepfun$ was previously defined in
  equation~\eqref{eq:defn_h}.
\begin{corollary}
 \label{corr:non_warm_start}
With $\initialstar$ as the starting distribution, we have
\begin{subequations}
  \begin{align}
    \tvnorm{\transition_{\tagmrw(\step_2)}^k(\initialstar)\!-\!\Target} & \leq
\threshold \quad \text{for all } k \geq \UNICON\ \dims^2\condition^2
\log^{1.5}\parenth{\frac{\condition}{\threshold^{1/\dims}}}, \quad \mbox{and} \\
\tvnorm{\transition_{\tagmala(\step_1)}^k(\initialstar)\!-\!\Target} &\leq
\threshold \quad \text{for all } k \geq \UNICON\ \dims^2 \condition
\log\parenth{\frac{\condition}{\threshold^{1/\dims}}}
\max\braces{1, \:
  \sqrt{\frac{\condition}{\dims} \,
    \log\parenth{\frac{\condition}{\threshold^{1/\dims}}}} }.
    \end{align}
    \end{subequations}
\end{corollary}
\noindent The proof follows by plugging the bound~\eqref{eq:warm_star}
in Theorems~\ref{thm:mala_mixing} and \ref{thm:mrw_mixing} and is
thereby omitted.

We now prove the claim~\eqref{eq:warm_star}.  Without loss of
generality, we can assume that $\myfun(\xstar) = 0.$ Such an
assumption is possible because substituting $\myfun(\cdot)$ by
$\myfun(\cdot) + \scalar$ for any scalar $\scalar$ leaves the
distribution $\Target$ unchanged.  Since $\myfun$ is
$\scparam$-strongly convex and $\smoothness$-smooth, applying
Lemma~\ref{lemma:strong_convexity}\ref{item:sc_quadratic_bound} and
Lemma~\ref{lemma:smoothness}\ref{item:smooth_quadratic_bound}, we
obtain that
\begin{align*}
    \frac{\smoothness}{2} \enorm{x-\xstar}^2 \geq \myfun(x)
    \geq \frac{\scparam}{2} \enorm{x-\xstar}^2, \quad \forall x \in \real^\dims.
\end{align*}
Consequently, we find that $\int_{\realdim} e^{-f(x)}dx \leq
(2\pi/\scparam)^{\dims/2}$.
Making note of the lower bound
\begin{align}
\label{eq:target_lower_bound}
    \target(x)
    \geq \frac{e^{-\frac{\smoothness}{2}\enorm{x-\xstar}^2}}{(2\pi\scparam^{-1})^{\dims/2}},
\end{align}
and plugging in the expression for the density of $\initialstar$
yields the claim~\eqref{eq:warm_star}.

We now derive results for the case when we do not have access to exact parameters, e.g.,
if the mode $\xstar$ is known approximately, and/or we only have
an upper bound for the smoothness parameter $\smoothness$---a situation quite prevalent in practice.

\subsubsection{Starting distribution with inexact parameters} 
\label{ssub:starting_with_inexact_parameters}

Note that $\xstar$ is also the unique global minimum of the negative
log-density $\myfun$.  For the strongly convex function~$\myfun$,
using a first-order method, like gradient descent, we can obtain an
$\modetolerance$-approximate mode $\xtilde$ using
$\condition\log(1/\modetolerance)$ evaluations of the
gradient~$\gradf$.  Suppose we have access to a point $\xtilde$ such
that $\enorm{\xtilde-\xstar} \leq \modetolerance$ and have an upper
bound estimate $\tilde{\smoothness} \geq \smoothness$ for the
smoothness.

We now consider the case of starting distribution $\initialtilde =
\NORMAL(\xtilde, (2\tilde{\smoothness})^{-1}\Ind_\dims)$, as a proxy
for the feasible start $\initialstar = \NORMAL(\xstar,
\smoothness^{-1}\Ind_\dims)$ discussed above. Note the difference in
mean and the covariance between the distributions $\initialtilde$ and
$\initialstar$.  Given the handy result in
Theorem~\ref{thm:mala_mixing}, it suffices to bound the warmness
parameter for the distribution~$\initialtilde$.  Applying the triangle
inequality, we find that
\begin{align}
\label{eq:xtilde_triangle_inequality}
  \enorm{x-\xtilde}^2 \geq \frac{1}{2}\enorm{x-\xstar}^2 - \enorm{\xstar-\xtilde}^2
\end{align}
and consequently that
\begin{align*}
  \initialtilde(x)
  &= (\pi\tilde{\smoothness}^{-1})^{-\dims/2}
  \exp\parenth{-{\tilde{\smoothness}\enorm{x-\xtilde}^2}}\\
  &\leq (\pi\tilde{\smoothness}^{-1})^{-\dims/2}
  \exp\parenth{-\frac{\tilde{\smoothness}\enorm{x-\xstar}^2}{2}
  + \tilde{\smoothness}\enorm{\xtilde-\xstar}^2}
\end{align*}
Using the lower bound~\eqref{eq:target_lower_bound} on the target density, we find that
\begin{align*}
  \frac{\initialtilde(x)}{\target(x)} &\leq
  \parenth{\frac{\tilde{\smoothness}}{\smoothness} \cdot 2\condition}^{\dims/2}
  \exp\parenth{\tilde{\smoothness}\enorm{\xtilde-\xstar}^2
  - \frac{(\tilde{\smoothness}-\smoothness)\enorm{x-\xstar}^2}{2}}\\
  &\leq \exp\parenth{\frac{\dims}{2} \log (2\condition
  \tilde{\smoothness}/\smoothness) + \tilde{\smoothness}\modetolerance^2},
\end{align*}
where the last inequality follows from the fact that $\tilde{\smoothness} \geq \smoothness$.
In other words, the distribution $\initialtilde$ is $\tilde\warmparam$-warm with respect to the target distribution $\target$, where \mbox{$\tilde\warmparam =
\exp\parenth{\frac{\dims}{2} \log (2\condition\tilde{\smoothness}/\smoothness) + \tilde{\smoothness}\modetolerance^2}$.}

Using Theorem~\ref{thm:mala_mixing}, we now derive a mixing time bound for MALA with starting distribution $\initialtilde$.
For any threshold $\threshold \in (0, 1]$, we use the step size $\step_3 = \unicontwo \stepfun(\threshold/(2\tilde\warmparam))$.
Invoking Theorem~\ref{thm:mala_mixing} and plugging in the definition~\eqref{eq:defn_r} of $\stepfun$, we find that
$\tvnorm{\transition_{\tagmala(\step_3)}^k(\initialtilde)-\Target}\leq \threshold$, for all
\begin{align}
\label{eq:tv_bound_xtilde}
  k \geq \UNICON\dims^2\condition\parenth{\log\frac{2\condition\tilde{\smoothness}/\smoothness}{\threshold^{1/\dims}} +
  \frac{\tilde{\smoothness}\modetolerance^2}{\dims}}
  \max\braces{1, \sqrt{\frac\condition\dims}\parenth{\sqrt{\log\frac{2\condition\tilde{\smoothness}/\smoothness}{\threshold^{1/\dims}}} + \frac{\sqrt{\tilde{\smoothness}}\modetolerance}{\sqrt{\dims}}} },
\end{align}
which also recovers the bound from corollary~\ref{corr:non_warm_start} for MALA as $\modetolerance \rightarrow 0$
and $\tilde{\smoothness} \to \smoothness$.
Note that the mixing time increases (additively) by
$\calo\parenth{\condition\dims\modetolerance^2\tilde{\smoothness}/\smoothness}$
when we only have an $\modetolerance$-approximate mode,
which is an $(\tilde{\smoothness}/\smoothness \cdot \modetolerance/\dims)$-fraction
increase in the mixing time bound with starting distribution $\initialstar$.
A mixing time bound for MRW with starting distribution~$\initialtilde$ can be obtained in a similar fashion and is thereby omitted.


\subsection{Weakly log-concave densities} 
\label{sub:nonstrongly_log_concave_densities}

In this section, we show that MALA can also be used for approximate
sampling from a density which is $\smoothness$-smooth
but not necessarily strongly log-concave (also referred to as weakly 
log-concave, see, e.g.,~\citealt{dalalyan2016theoretical}).
In simple words, the negative log-density $\myfun$ satisfies the
condition~\eqref{eq:smoothness} with parameter $\smoothness$
and satisfies the condition~\eqref{eq:strongly_convex} with parameter
$\scparam=0$,
(equivalently we have $ \smoothness \Ind_\dims\succeq\nabla^2 \myfun(x) \succeq
0 $; see Appendix~\ref{sec:some_basic_properties} for further details.)
Note that we still assume that
$\int_{\mathbb{R}^d} e^{-f(x)} dx <\infty$ so that the distribution $\Pi$
is well-defined.

In order to make use of our previous machinery for such a case, we approximate
the given log-concave density~$\Target$ with a strongly log-concave density~$\approxTarget$
such that $\tvnorm{\approxTarget - \Target}$ is small.  Next, we use
MALA to sample from $\approxTarget$ and consequently obtain an
approximate sample from $\Target$.  In order to construct
$\approxTarget$, we use a scheme previously suggested by
\cite{dalalyan2016theoretical}.  With $\lambda$ as a tuning
parameter, consider the distribution $\approxTarget$ given by the
density
\begin{align}
    \label{eq:approxmyfun}
    \tilde{\target}(x) = \frac{1}{\displaystyle \int_{\realdim}
      e^{-\approxmyfun(y)}dy} e^{-\approxmyfun(x)} \quad \text{where}
    \quad \approxmyfun(x) = \myfun(x) + \frac{\lambda}{2} \enorm{x -
      \xstar}^2.
\end{align}
\cite{dalalyan2016theoretical} (Lemma 3) showed
that that the total variation distance between $\Target$ and
$\approxTarget$ is bounded as follows:
\begin{align*}
    \tvnorm{\approxTarget - \Target} \leq
    \frac{1}{2}\vecnorm{\approxmyfun - \myfun}{L^2(\target)} \leq
    \frac{\lambda}{4} \parenth{\int_{\realdim} \enorm{x-\xstar}^4
      \target(x) dx}^{1/2}.
\end{align*}
Suppose that the original distribution~$\Target$ has its fourth moment
bounded as
\begin{align}
\label{eq:log_concave_assumption}
  \int_{\realdim} \enorm{x-\xstar}^4 \target(x) dx \leq \dims^2 \nu^2.
\end{align}
We now set $\lambda \defn 2\threshold/(\dims\nu)$ to obtain
$\tvnorm{\approxTarget - \Target}  \leq \threshold/2$.
Since $\approxmyfun$ is $\lambda/2$-strongly convex and
$\smoothness + \lambda/2$-smooth,
the condition number of $\approxTarget$ is given by
$\tilde\condition = 1 + \smoothness\dims\nu/\threshold$.
We substitute $\tilde\condition = \smoothness\dims\nu/\threshold$
to obtain simplified expressions for mixing time bounds
in the results that follow.
Since now the target distribution is $\approxTarget$,
we suitably modify the step size for MALA as follows:
\begin{align*}
    \stepfun_{\tiny\text{lc}}(\res) = \frac{1}{\smoothness\dims}
    \min\braces{\frac{\sqrt{\res}}{\radius(\res) \,
        \sqrt{\nu\smoothness}}, \; 1}
\end{align*}
where the function $\radius$ was previously defined in
equation~\eqref{eq:defn_r}.  We refer to this new set-up with a
modified target distribution~$\approxTarget$ as the \emph{modified
  MALA method}.  To keep our results simple to state, we assume that
we have a warm start with respect to $\approxTarget$.
\begin{corollary}
    \label{cor:mala_non_strongly_mixing}
Assume that $\Target$ satisfies the 
condition~\eqref{eq:log_concave_assumption}.
Then for any given error-tolerance $\threshold \in (0, 1)$,
and, any $\warmparam$-warm start $\initial$, the modified MALA
 method with step size $\step =
 \UNICON\stepfun_{\tiny\text{lc}}(\threshold/(2\warmparam))$ satisfies
 $ \tvnorm{\transition_{\tagmala(\step)}^k(\initial)-\Target} \leq
    \threshold $ for all
\begin{align*}
   k \geq
    \unicontwo \log\parenth{\frac{4{\warmparam}}{\threshold}}
 \max\braces{ {\frac{\dims^2\smoothness \nu}{\threshold}},
   \ \dims^{2}\parenth{\frac{\smoothness \nu}{\threshold}}^{1.5}
   \radius\parenth{\frac{\threshold}{4\warmparam}} },
\end{align*}
 where $\UNICON, \unicontwo$ denote universal positive constants.
\end{corollary}
The proof follows by combining the triangle inequality, as applied to
the TV norm, along with the bound from Theorem~\ref{thm:mala_mixing}.
Thus, for weakly log-concave densities, modified MALA mixes in
$\calo\parenth{\dims^2/\threshold^{1.5}}$, which improves upon the
$\calo\parenth{\dims^3/\threshold^4}$ mixing time bound for a ULA
scheme on $\approxTarget$, as established by \cite{dalalyan2016theoretical}.
A mixing time bound of $\calo\parenth{\dims^3/\threshold^{2}}$ for MRW can be derived similarly for this case,
simply by noting that the new condition number
$\tilde\condition = \smoothness\dims\nu/\threshold$ for the
modified density and the fact that the mixing time of MRW is
$\calo\parenth{\dims\tilde\condition^2}$ in the strongly log-concave
setting.


\section{Numerical experiments}
\label{sec:simulations}

In this section, we compare MALA with ULA and MRW in various
simulation settings. The step-size choice of ULA follows
from the paper by~\cite{dalalyan2016theoretical} in the case of a warm start.
The step-size choice of MALA and MRW used in our experiments in our results
are summarized in Table~\ref{tab:step_size}.

\paragraph{Summary of experiment set-ups and diagnostic tools:}
We consider four different experiments: (i) sampling a multivariate
Gaussian (Section~\ref{sub:multivariate_gaussian_dimension_dependency}),
(ii) sampling a Gaussian mixture (Section~\ref{sub:close_gaussian_mixture}),
(iii) estimating the MAP with credible intervals in a Bayesian logistic
regression set-up (Section~\ref{sub:bayesian_logistic_regression})
and (iv) studying the effect of step-size on the accept reject step (Section~\ref{sub:step_size_vs_accept_reject_rate}).
Since TV distance for continuous measures is hard to estimate, we use several
proxy measures for convergence diagnostics: (a) errors in quantiles,
(b) $\ell_1$-distance in histograms (which we refer to as discrete tv-error),
(c) error in sample
MAP estimate, (d) trace-plot along different coordinates and (e) autocorrelation plot.
While the first three measures (a-c) are useful
for diagnosing the convergence of random walks over several independent runs,
the last two measures (d-e) are useful for diagnosing the rate
of convergence of the Markov chain in a single long run.

\subsection{Dimension dependence for multivariate Gaussian}
\label{sub:multivariate_gaussian_dimension_dependency}

The goal of this simulation is to demonstrate the dimension dependence
in experiments, for mixing time of ULA, MALA, and MRW when the target
is a non-isotropic multivariate Gaussian.
Note that Theorems~\ref{thm:mala_mixing} and \ref{thm:mrw_mixing}
imply that the dimension dependency for both MALA and MRW is $\dims$.
We consider sampling from multivariate Gaussian with density~$\target$
defined by
\begin{align}
\label{eq:gaussian_density}
  x \mapsto \target(x) \propto e^{-\frac{1}{2} x\tp \Sigma^{-1} x},
\end{align}
where $\Sigma \in \real^{\dims \times \dims}$ the covariance matrix to
be specified.  For this target distribution, the function $\myfun$,
its derivatives are given by
\begin{align*}
  \myfun(x) = \frac{1}{2} x\tp \Sigma^{-1} x,\quad \gradf(x) =
  \Sigma^{-1} x, \quad\text{and}\quad \hessf(x) = \Sigma^{-1}.
\end{align*}
Consequently, the function $\myfun$ is strongly convex with parameter
$\scparam = 1/\lambda_\text{max}(\Sigma)$ and smooth with parameter
$\smoothness = 1/\lambda_\text{min}(\Sigma)$.  For convergence
diagnostics, we use the error in quantiles along different directions.
Using the exact quantile information for each direction for Gaussians,
we measure the error in the $75\%$ quantile of the sample distribution
and the true distribution in the \emph{least favorable direction},
i.e., along the eigenvector of $\Sigma$ corresponding to the
eigenvalue $\lambda_\text{max}(\Sigma)$.  The \emph{approximate mixing time}
$\hat{k}_{\text{mix}}(\delta)$
is defined as the smallest iteration when this error falls below
$\delta$.  We use $\initialstar$ as the
initial distribution where $\initialstar = \NORMAL\parenth{0,
  \smoothness^{-1}\Ind_d}$.

\subsubsection{Strongly log-concave density} 
\label{ssub:strongly_log_concave_density}

\begin{figure}
  \begin{center}
    \begin{tabular}{cc}
      \widgraph{0.48\textwidth}{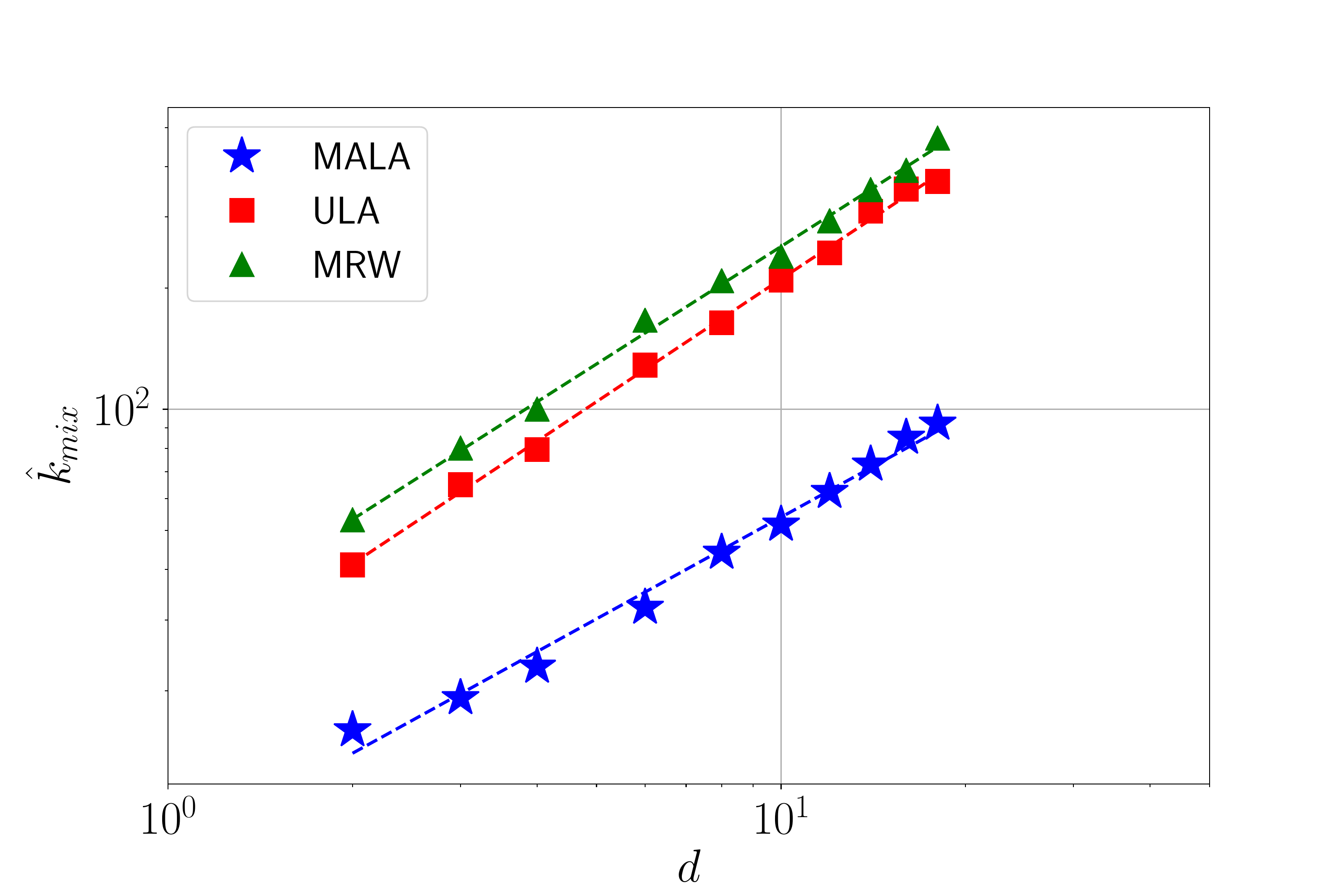}
      & \widgraph{0.48\textwidth}{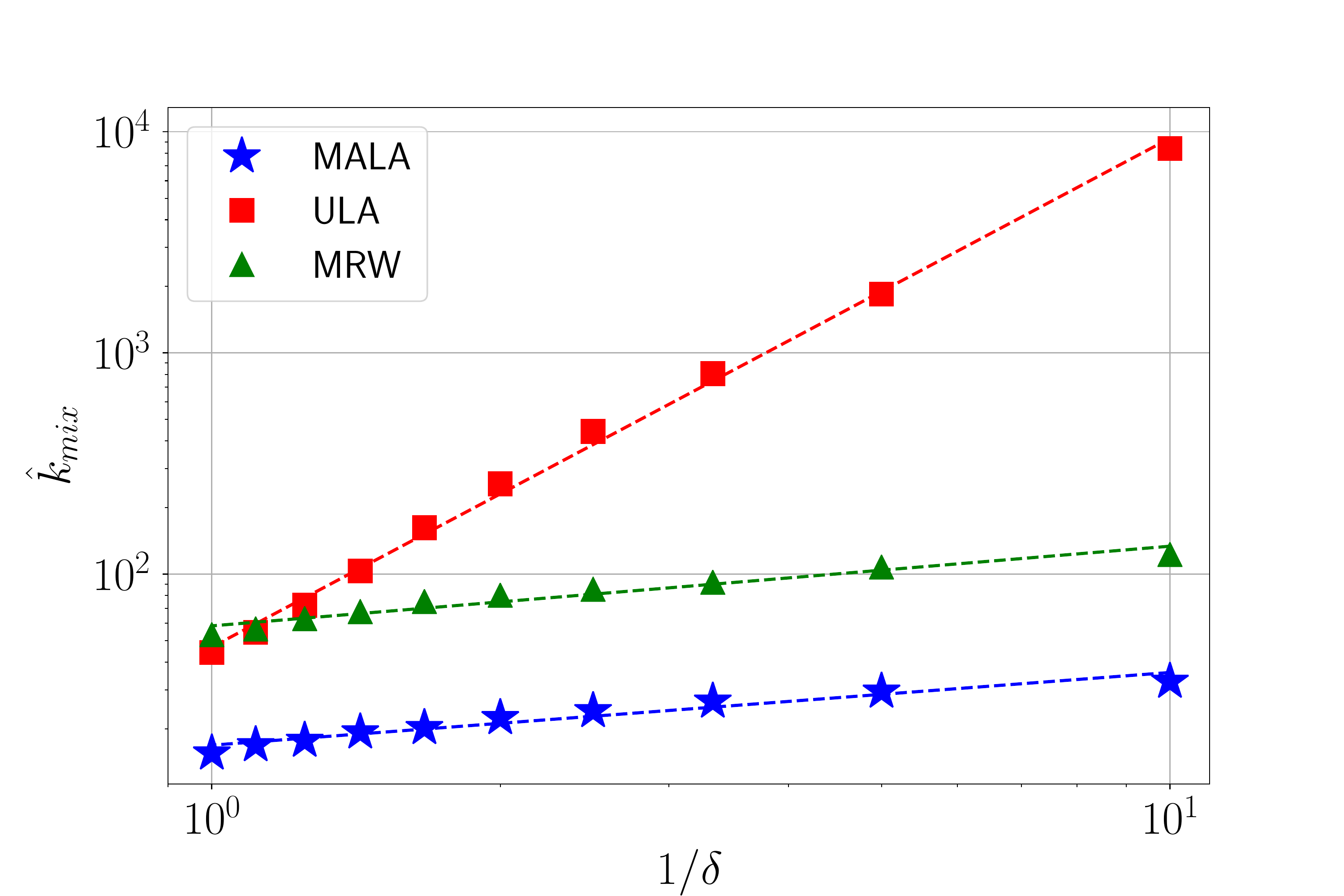}\\
      (a) & (b)
    \end{tabular}
    \caption{Scaling of the approximate mixing time $\hat{k}_{\text{mix}}$
    (refer to the discussion after equation~\eqref{eq:gaussian_density}
    for the definition) on multivariate Gaussian density~\eqref{eq:gaussian_density}
    where the covariance has condition number $\condition=4$.  (a) Dimension
    dependency.  \mbox{(b) Error-tolerance} dependency.}
    \label{fig:nonisotropic_gaussian_dependency}
  \end{center}
\end{figure}

The step-sizes are chosen according to
Table~\ref{tab:step_size}.  For ULA, the error-tolerance~$\threshold$ is
chosen to be $0.2$.  We set $\Sigma$ as a diagonal matrix with the
largest eigenvalue $4.0$ and the smallest eigenvalue $1.0$ so that the
$\condition = 4$ is fixed across different settings. For a fixed
dimension~$\dims$, we simulate $10$ independent runs of the three
chains each with $\sampleobs = 10,000$ samples to determine the
approximate mixing time. The final approximate mixing time for each
walk is the average of that over these $10$ independent runs.
Figure~\ref{fig:nonisotropic_gaussian_dependency}(a) shows the
dependency of the approximate mixing time as a function of dimension
$\dims$ for the three random walks in log-log scale.
To examine the dimension dependency, we perform
linear regression for approximate mixing time with respect to
dimensions in the log-log scale.
The computations reveal that the dimension dependency of MALA, ULA and MRW
are all close to order $\dims$ (slope $0.84$, $1.01$ and $0.97$).
Figure~\ref{fig:nonisotropic_gaussian_dependency}(b) shows the
dependency of the approximate mixing time on the inverse
error $1/\threshold$ for the three random walks in log-log scale. For
ULA, the step-size is error-dependent, precisely chosen to be $10$
times of $\threshold$.  A linear regression of the approximate mixing
time on the inverse error $1/\threshold$ yields a slope of $2.23$
suggesting the error dependency of order $1/{\threshold^2}$ for ULA.
A similar computation for MALA and MRW yields a slope of $0.33$ for both
the cases which not only suggests a significantly better error dependency
for these two chains but also partly verifies their theoretical mixing
time bounds of order $\log(1/\threshold)$.

\begin{table}[h]
 \centering
  \begin{tabular}{cccc}
    \hline
    \toprule
    {\bf Random walk}   & {ULA}   & {MALA}     & {MRW} \\
    \midrule
    {\bf Step size}      & $\displaystyle\frac{\threshold^2}{\dims\condition\smoothness}$  & $\displaystyle\frac{1}{\smoothness}
          \min\braces{\frac{1}{\sqrt{\dims\condition}}, \frac{1}{\dims}}$ & $\displaystyle\frac{1}{\dims\condition\smoothness}$ \\[3ex]
    \bottomrule
    \hline
  \end{tabular}
  \caption{Step size used in simulations to obtain $\threshold$-accuracy
  for different random walks in
      $\realdim$ with target $\target \propto e^{-\myfun}$ such that
      $\scparam \Ind_\dims \preceq \hessf(x) \preceq
      \smoothness\Ind_\dims$ for any $x \in \realdim$ and $\condition
      \defn \smoothness/\scparam$.}
  \label{tab:step_size}
\end{table}

\subsubsection{Weakly log-concave density} 
\label{ssub:weakly_log_concave_density}
\begin{figure}[t]
  \begin{center}
    \begin{tabular}{cc}
      \widgraph{0.48\textwidth}{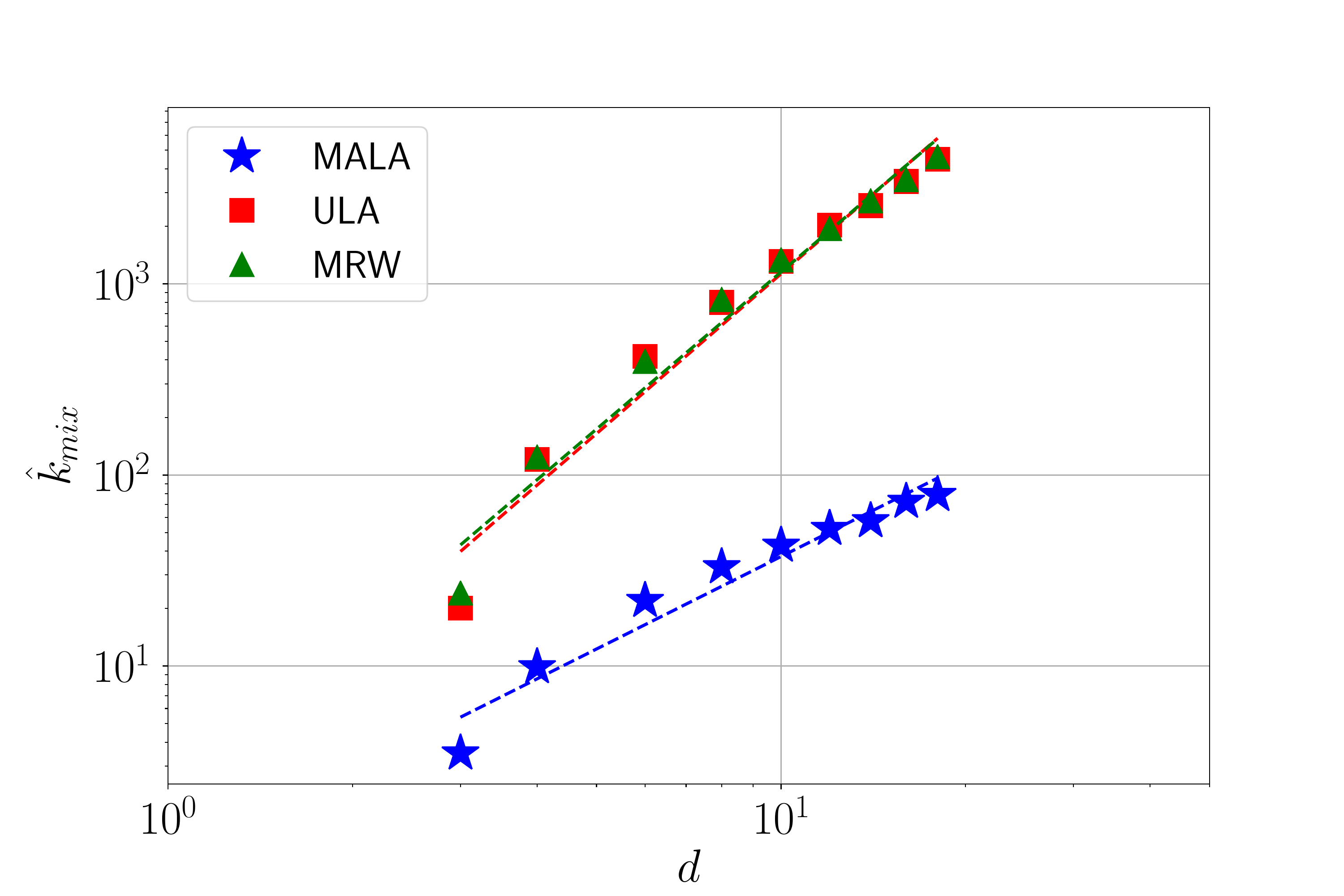}
      & \widgraph{0.48\textwidth}{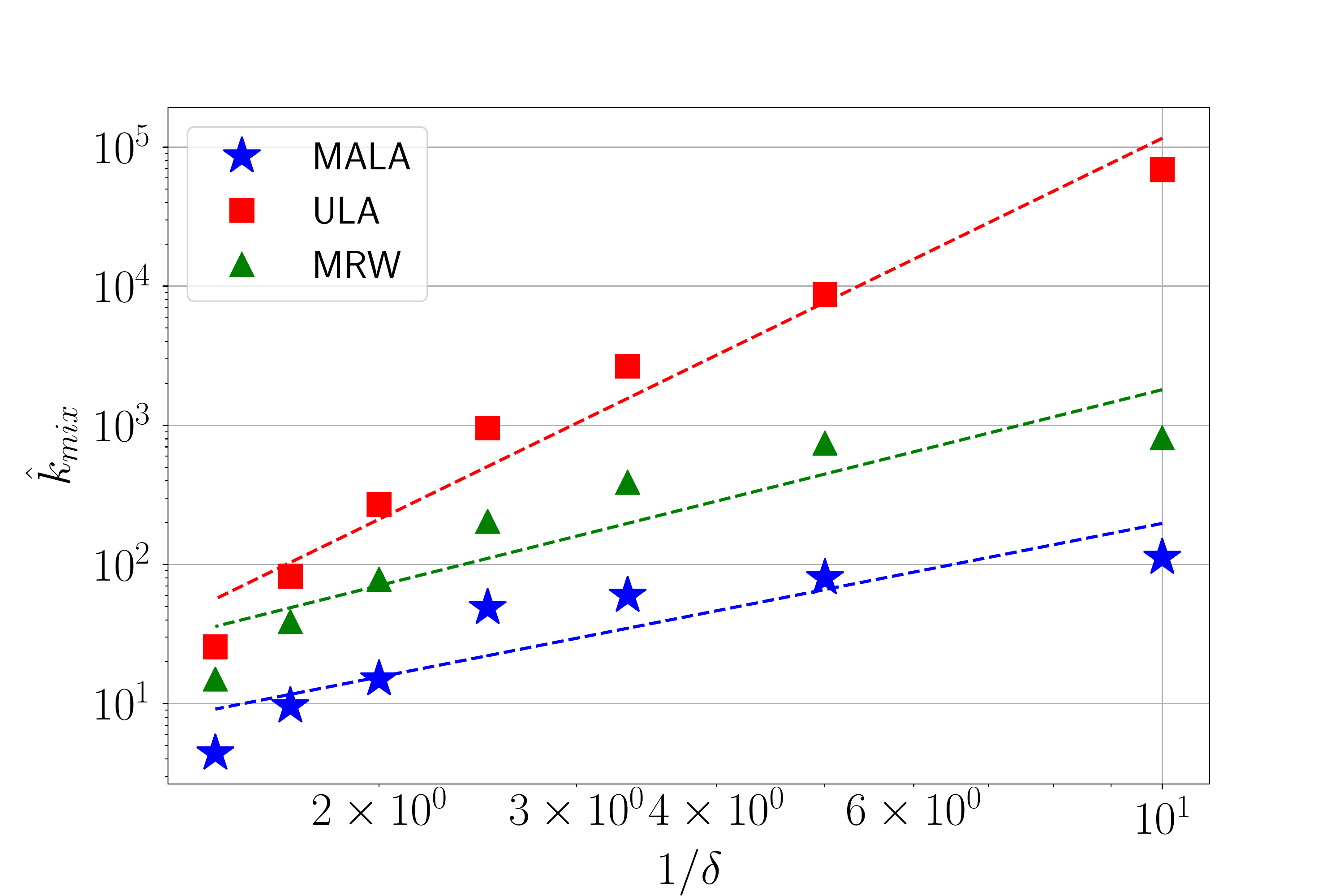}\\
      (a) & (b)
    \end{tabular}
    \caption{Scaling of the approximate mixing time $\hat{k}_{\text{mix}}$
    (refer to the discussion after equation~\eqref{eq:gaussian_density}
    for the definition) for a close to weakly log-concave Gaussian density.
    (a) Dimension dependency.
    \mbox{(b) Error-tolerance} dependency for fixed dimension .}
  \label{fig:nonisotropic_weakly_gaussian_dependency}
  \end{center}
\end{figure}

We now discuss the convergence of the random walks when the Gaussian is flat along a direction.
In particular, we consider the Gaussian distribution such that
$\lambda_\text{max}(\Sigma)=1000$ and $\lambda_\text{min}(\Sigma) = 1$.
Such a setting implies that the strong convexity parameter $\scparam = 0.001
\approx 0$ and hence our target density mimics a weakly log-concave density.
For convergence diagnostics, we use the error in quantiles along one direction other than the ones which correspond to $\lambda_\text{max}(\Sigma)$ and $\lambda_\text{min}(\Sigma)$.
Using the exact quantile information for each direction for Gaussians,
we measure the error between the $75\%$ quantile of the sample distribution
and the true distribution in that direction. The approximate mixing time
is defined as the smallest iteration when this error falls below
$\delta$.  We use $\initialstar$ as the
initial distribution where $\initialstar = \NORMAL\parenth{0,
  \smoothness^{-1}\Ind_d}$.  The step-sizes are chosen according to
Table~\ref{tab:step_size} where $\scparam$ is chosen to be $\threshold/(\dims\smoothness)$.
For dimension dependence experiments, we fix the error-tolerance~$\threshold$ as $0.2$.
For a fixed
dimension~$\dims$, we simulate $10$ independent runs of the three
chains each with $\sampleobs = 10,000$ samples to determine the
approximate mixing time. The final approximate mixing time for each
walk is the average of that over these $10$ independent runs.
Figure~\ref{fig:nonisotropic_weakly_gaussian_dependency}(a) and \ref{fig:nonisotropic_weakly_gaussian_dependency}(b) show the
dependency of the approximate mixing time as a function of dimension
$\dims$ and the inverse
error $1/\threshold$ respectively, for the three random walks on this weakly log-concave density (log-log scale).
Linear fits on the log-log scale reveal that the dimension dependence of mixing time for MALA is close to $\dims^2$ (slope $1.61$), and that for ULA is close to $\dims^3$ (slope $2.78$) and for MRW it is approximately of order $\dims^3$ (slope $2.73$).
Linear fits of the approximate mixing
time on the inverse error $1/\threshold$ yield a slope of $3.92$ for ULA thereby
suggesting an error dependence of order $1/{\threshold^4}$, while for MALA and MRW this dependence is of order $1/{\threshold^{1.5}}$ (slope $1.56$) and of order $1/{\threshold^2}$ (slope $2.01$), respectively.
These scalings partly verify the rates derived in Corollary~\ref{cor:mala_non_strongly_mixing} and demonstrate the gains of MALA over ULA for the weakly log-concave densities.

\subsubsection{Warmness in simulations} 
\label{ssub:warmness_in_simulations}
Strictly speaking, for both the cases considered above, the starting
distribution was not warm since we used $\initialstar$ as the
starting distribution and the corresponding warmness parameter
$\warmparam = \calo({e^\dims})$ scales exponentially with dimension
$\dims$.  However, the mixing time observed in the simulations, albeit
with a heuristic measure, are $\dims$ times faster than those stated
with $\initialstar$ as the starting distribution in
Corollary~\ref{corr:non_warm_start}, and are in fact consistent with
the results for the warm-start which are stated in
Theorems~\ref{thm:mala_mixing} and \ref{thm:mrw_mixing}.  We believe
that the results stated in Corollary~\ref{corr:non_warm_start}, with
$\initialstar$ as the starting distribution can be improved by a
factor of $\dims$.  See Section~\ref{sec:discussion} for a discussion on
recent work on this question.

\subsection{Behavior for Gaussian mixture distribution}
\label{sub:close_gaussian_mixture}

We now consider the task of sampling from a two component Gaussian
mixture distribution, as previously considered by
\cite{dalalyan2016theoretical} for illustrating the behavior
of ULA.  Here compare the behavior of MALA to ULA for this case.  The
target density is given by
\begin{align*}
x \mapsto \target(x) = \frac{1}{2 \parenth{2 \pi}^{\dims/2}}
\parenth{e^{-\vecnorm{x - a}{2}^2/2} + e^{-\vecnorm{x + a}{2}^2/2}},
\end{align*}
where $a \in \real^\dims$ is a fixed vector.  This density corresponds
to the two-mixture of equal weighted Gaussians $\NORMAL(a,
\Ind_\dims)$ and $\NORMAL(-a, \Ind_\dims)$.  In our notation, the
function $\myfun$ and its derivatives are given by: $\myfun(x) =
\frac{1}{2} \|x - a\|_2^2 - \log (1 + e^{-2x\tp a})$,
\begin{align*}
 \nabla \myfun(x) &= x - a + 2 a (1 + e^{2x \tp a})^{-1},
 \text{ and }, \nabla^2 \myfun(x) =  \Ind_d - 4 a a\tp \frac{e^{2x
   \tp a}}{\parenth{1 + e^{2x\tp a}}^{2}}.
\end{align*}
From examination of the Hessian, we see that the function $\myfun$ is
smooth with parameter $\smoothness = 1$, and whenever $\vecnorm{a}{2}
< 1$, it is also strongly convex with parameter $\scparam = 1 -
\vecnorm{a}{2}^2$.

\begin{figure}[h]
    \begin{center}
    \widgraph{0.45\linewidth}{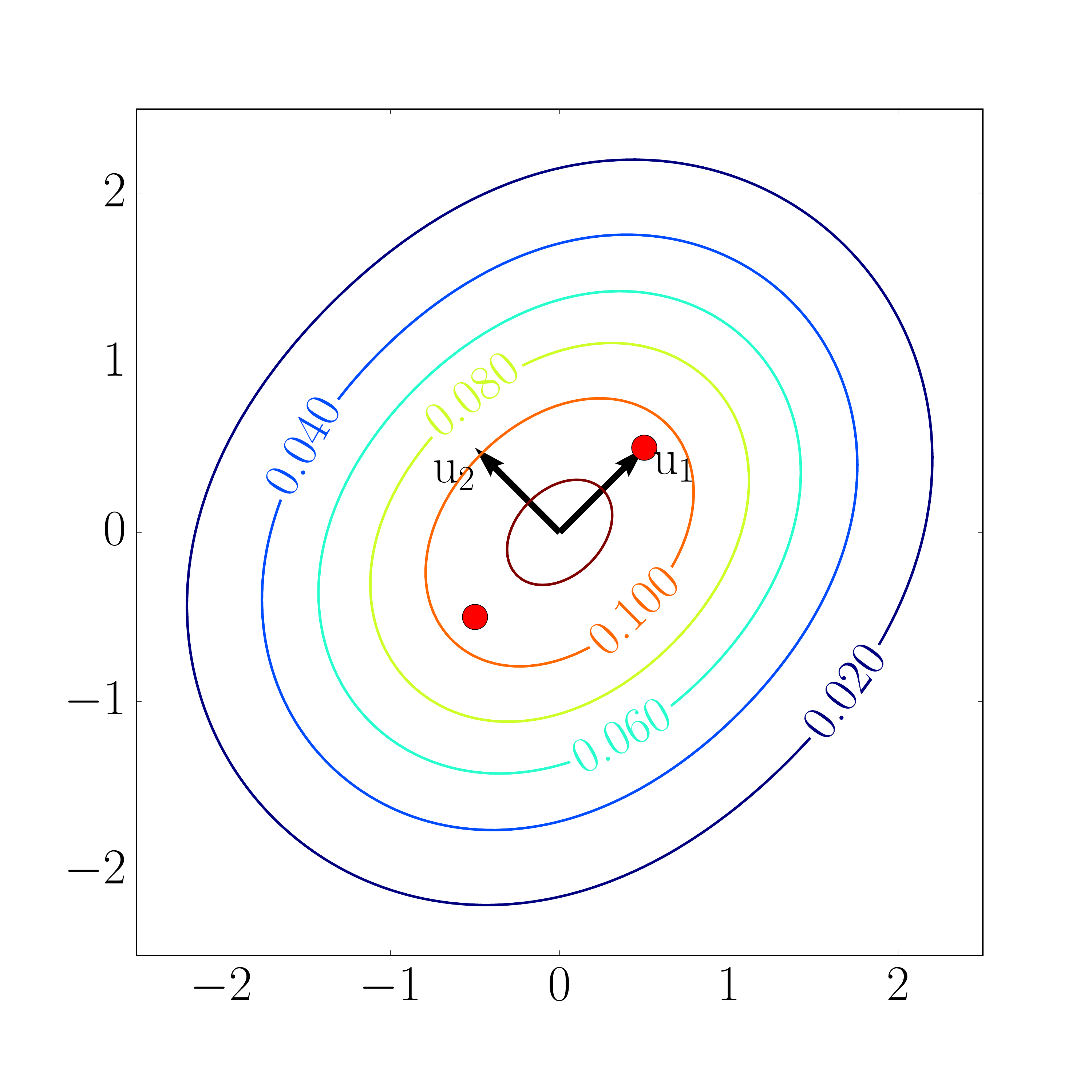}
    \caption{Level set of the density of the 2D Gaussian mixture.  The
      red dots are the location of the means $a$ and $-a$, where $a$
      is chosen such that $\vecnorm{a}{2}^2 = \frac{1}{2}$.  The
      arrows indicate the two principal directions $u_1$ and $u_2$
      along which the TV error is measured.}
    \label{fig:gaussian_mixture_level_set}
    \end{center}
\end{figure}

For dimension $\dims = 2$, setting $a = \parenth{\frac{1}{2},
  \frac{1}{2}}$ yields the parameters $\scparam = \frac{1}{2}$
and \mbox{$\smoothness = 1$}.  Figure~\ref{fig:gaussian_mixture_level_set}
shows the level sets of the density of this 2D-Gaussian mixture.  The
initial distribution is chosen as $\initialstar = \NORMAL\parenth{0,
  \smoothness^{-1}\Ind_d}$ and the step-sizes are chosen according to
Table~\ref{tab:mixing_times}, where for ULA, we set three different
choices of $\threshold = 0.2$ (ULA), $\threshold = 0.1$ (small-step
ULA) and $\threshold= 1.0$ (large-step ULA). Note that choosing a
smaller threshold~$\threshold$ implies that the ULA has a smaller
step size and consequently the chain takes larger to converge.
However, the asymptotic TV error with respect to the target
distribution $\Target$ for ULA also decreases with a decrease in step size.
These different choices of step sizes are made to
demonstrate the fundamental
trade-off between the rate of convergence and asymptotic error
for ULA and its inability to mix faster than MALA for different settings.

\begin{figure}[h]
  \begin{center}
    \begin{tabular}{cc}
    \widgraph{0.48\textwidth}{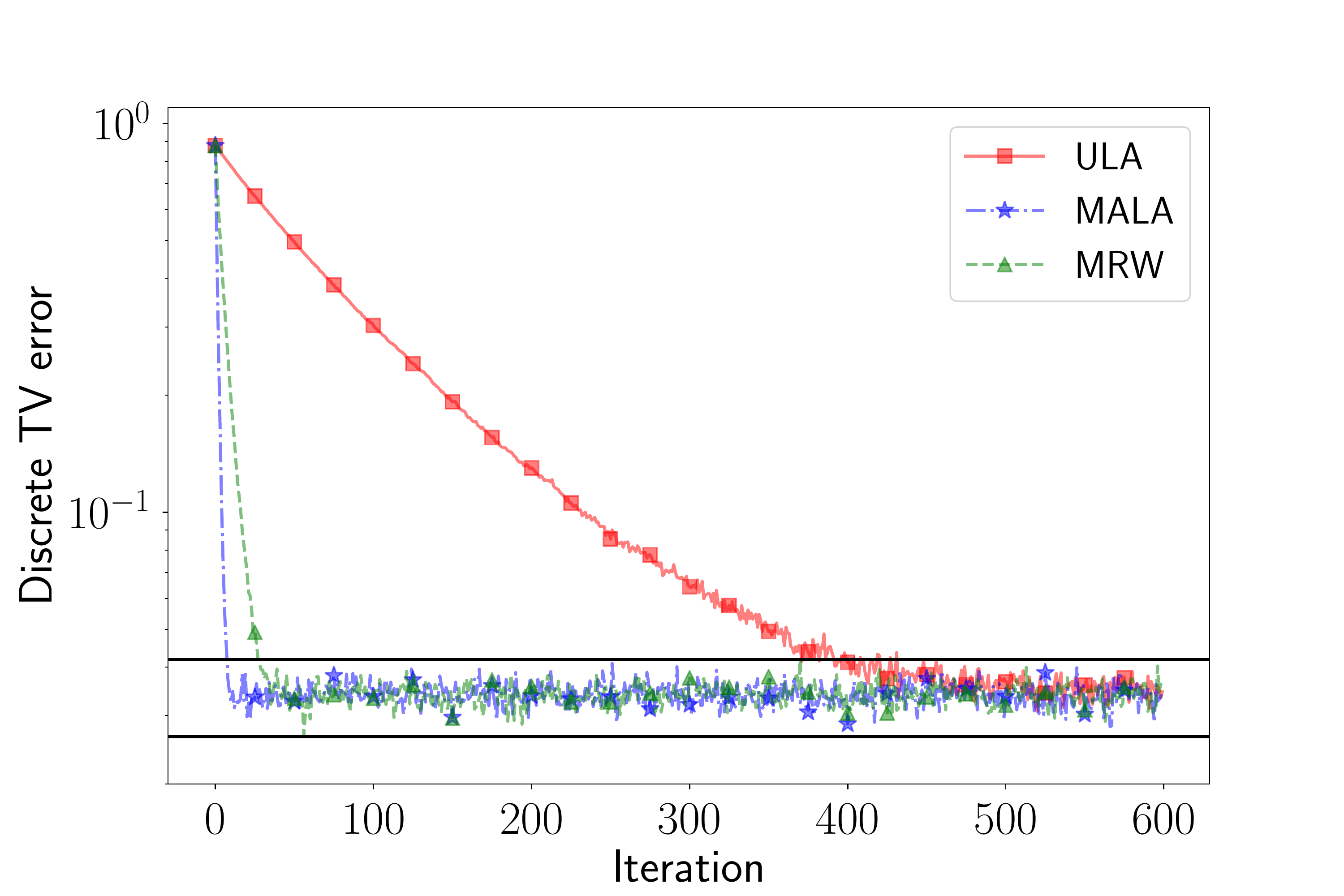}
    &
    \widgraph{0.48\textwidth}{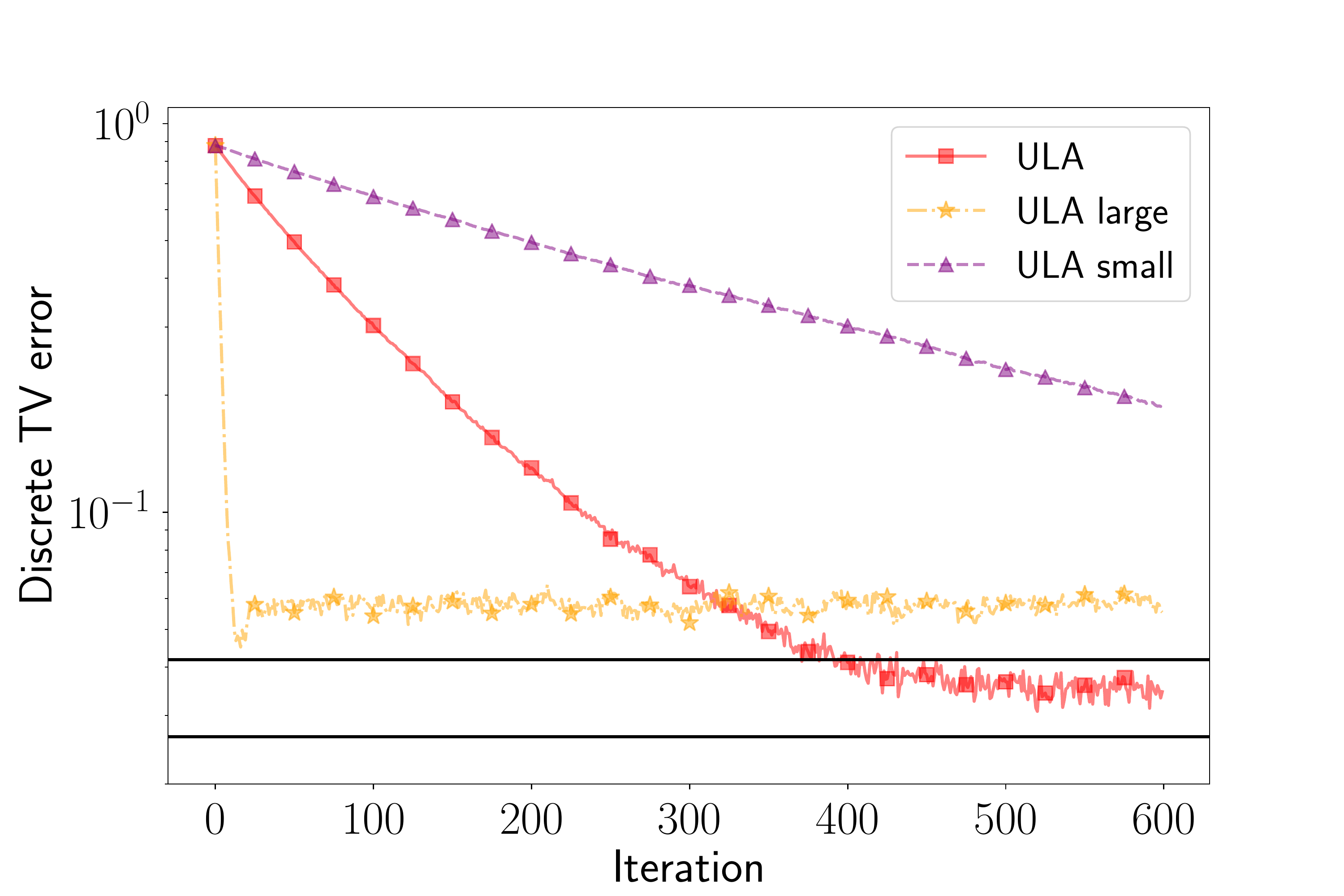} \\ (a) &
    (b)
    \end{tabular}
    \caption{Discrete TV error on a two component Gaussian mixture.
      (a) Behavior of three different random walks.  (b) Behavior of
      ULA with different choices of step sizes.}
    \label{fig:gaussian_mixture_tv_error}
  \end{center}
\end{figure}

Note that one can sample directly from the mixture of Gaussian under
consideration by drawing independently a Bernoulli$(1/2)$ random
variable $y$ and a standard normal variable $ z \sim \NORMAL(0,
\Ind_\dims)$, and then outputting the random variable
\begin{align*}
    x = y \cdot (z - a) + (1-y) \cdot (z + a)
\end{align*}
This observation makes it easy to diagnose the convergence of our
Markov chains with target $\target$.  In order to estimate the total
variation distance, we discretize the distribution of $\sampleobs=
250,000$ samples from $\target$ over a set of bins, and consider the
total variation of this discrete distribution from the empirical
distribution of the Markov chain over these bins.  We refer to this
measure as the discretized TV error.  We measure the sum of two
discrete TV errors of $250,000$ samples from $\target$ with the
empirical distribution obtained by simulating the chains ULA, MALA or
MRW, projected on two principal directions ($u_1$ and $u_2$), over a
discrete grid of size $B = 100$.
Figure~\ref{fig:gaussian_mixture_tv_error} shows the sum of the
discretized TV errors along $u_1$ and $u_2$, as a function of
iterations.  The true total variation distance between the
distribution of the iterate and the target distribution is upper
bounded by the sum of (a) the discretized TV error and (b) the error
caused by discretization.  In order to obtain a sense of the magnitude
of the error of type (b), we simulate $100$ runs of the discrete TV
error between two independent drawings from the true distribution
$\target$.  The two black lines in
Figure~\ref{fig:gaussian_mixture_tv_error} are the maximum and minimum
of these $100$ values.  The sample distribution at convergence is
expected to lie between the two black lines.

Figure~\ref{fig:gaussian_mixture_tv_error}(a) shows that ULA converges
significantly slower than MALA to the right distribution.
Figure~\ref{fig:gaussian_mixture_tv_error}(b) illustrates this point
further and shows that when compared to the ULA, the small-step ULA
($\threshold = 0.1$) converges at a much slower rate and large-step
ULA ($\threshold = 1.0$) has a larger approximation error (asymptotic
bias).

\begin{figure}[h]
  \begin{center}
    \begin{tabular}{ccc}
    \widgraph{0.33\textwidth}{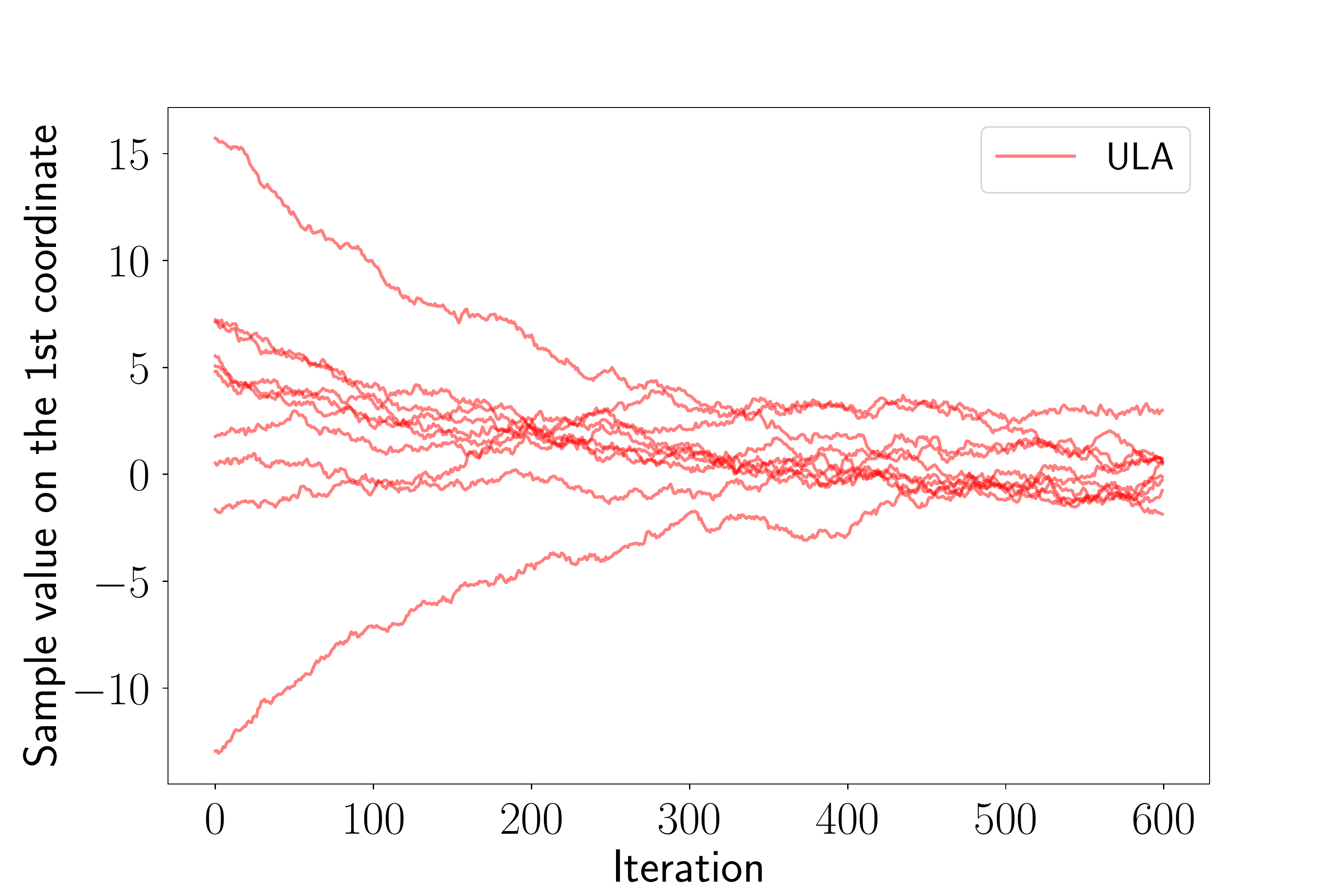} &
    \widgraph{0.33\textwidth}{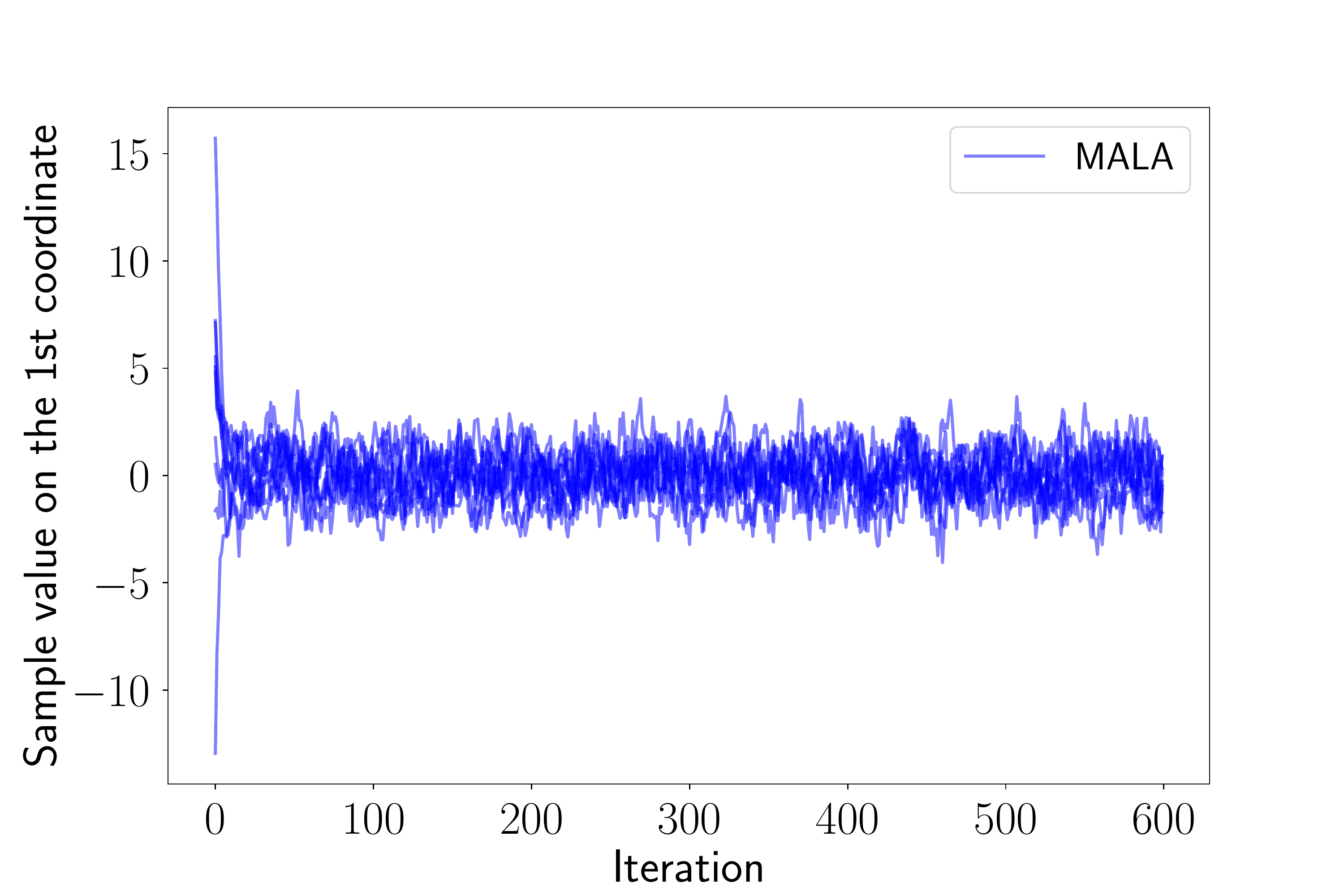} &
    \widgraph{0.33\textwidth}{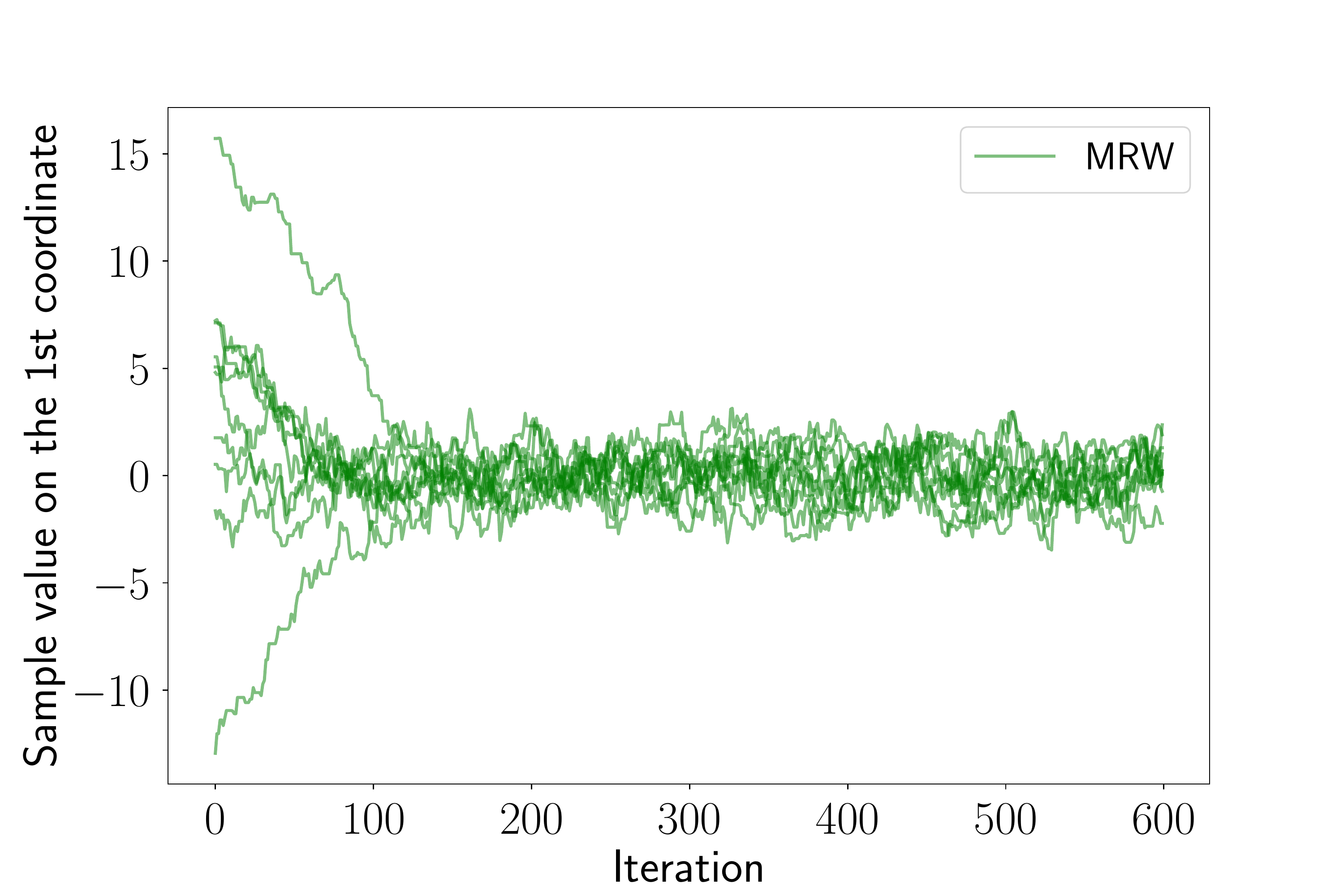} \\ (a) &
    (b) & (c)
    \end{tabular}
    \caption{Trace-plot of the first coordinate on a two component Gaussian mixture.
      (a)~Trace-plot of ULA.  (b) Trace-plot of MALA. (c) Trace-plot of MRW.}
    \label{fig:gaussian_mixture_traceplot}
  \end{center}
\end{figure}

\begin{figure}[h]
    \begin{center}
    \widgraph{0.5\linewidth}{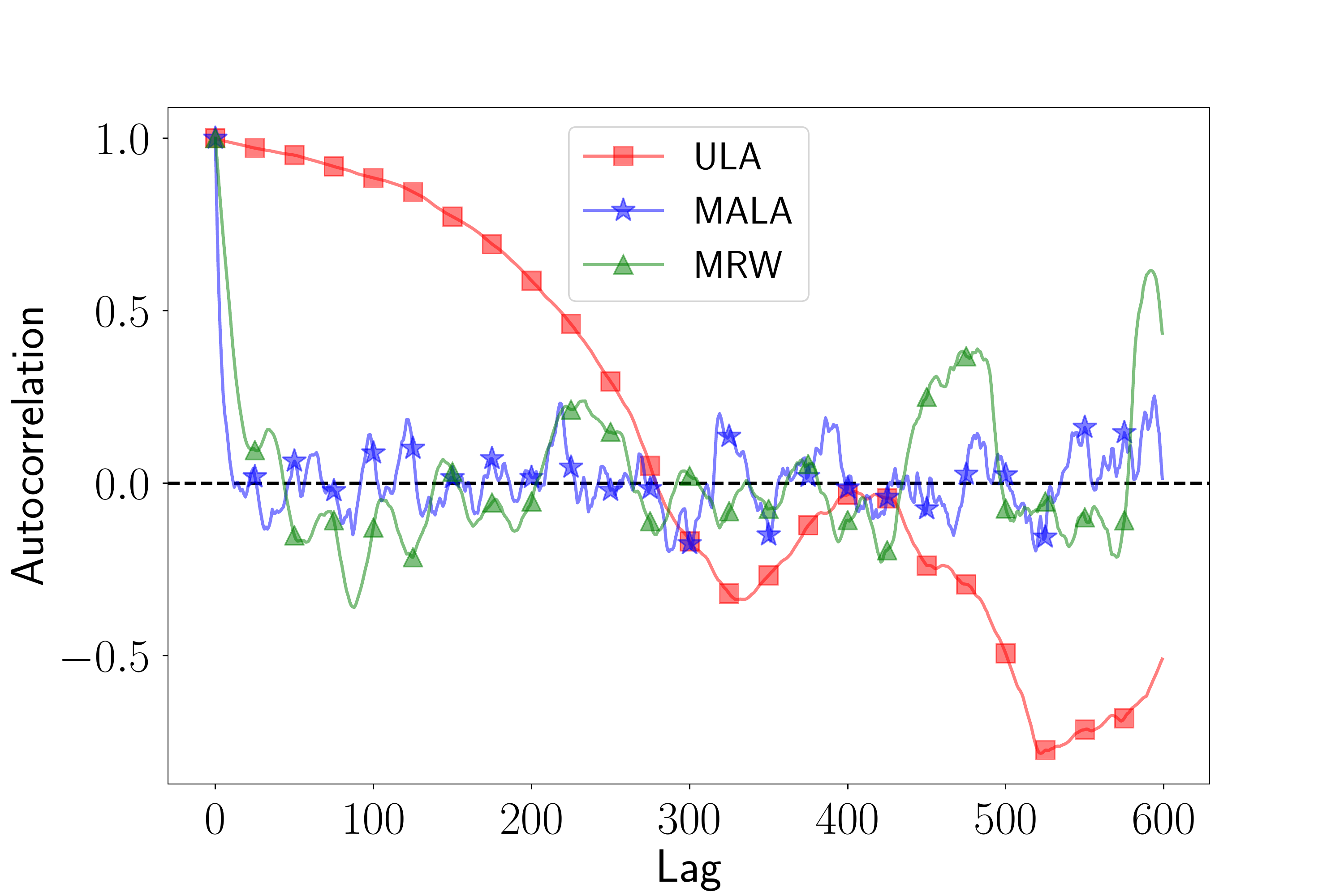}
    \caption{Markov chain autocorrelation function plot. The burn-in time for the plot is set to $300$ iterations.}
    \label{fig:gaussian_mixture_autocorrelation}
    \end{center}
\end{figure}

We accompany the study based on exact TV error computation with two
classical convergence diagnostic plots for general MCMC
algorithms. Figure~\ref{fig:gaussian_mixture_traceplot} shows the
trace plots of the three sampling algorithms in $10$ runs. Comparing
the three panels (a)---(c) in
Figure~\ref{fig:gaussian_mixture_traceplot}, we observe that the trace
plot of MALA stabilizes much faster than that of ULA and MRW.
Furthermore, to compare the efficiency of the chains in stationarity,
Figure~\ref{fig:gaussian_mixture_autocorrelation} shows the
autocorrelation function of the three chains.  In order to ensure that
these autocorrelations are computed for the stationary distribution,
we set in practice the burn-in period to be $300$ iterations.  Again,
we observe that MALA is more efficient than ULA and MRW.

\subsection{Bayesian Logistic Regression}
\label{sub:bayesian_logistic_regression}

We now consider the problem of logistic regression in a
frequentist-Bayesian setting, similar to that considered by
\cite{dalalyan2016theoretical}.  Once again, we establish
that MALA has superior performance relative to ULA.  Given a binary
variable $y \in \{0,1 \}$ and a covariate $x \in \real^\usedim$, the
logistic model for the conditional distribution of $y$ given $x$ takes
the form
\begin{align}
\label{eq:logistic}
  \Prob(y = 1 \vert x; \theta) = \frac{e^{\theta\tp x}}{1 + e^{\theta
      \tp x}},
\end{align}
for some parameter $\theta \in \real^\dims$.

In a Bayesian framework, we model the parameter $\theta$ in the
logistic equation as a random variable with a prior distribution
$\target_0$.  Suppose that we observe a set of independent samples
$\braces{(x_i, y_i)}_{i=1}^\obs$ with $(x_i, y_i) \in \real^\dims
\times \{0, 1\}$, with each $y_i$ conditioned on $x_i$ drawn from a
logistic distribution with some unknown parameter $\theta^*$.  Using
Bayes' rule, we can then compute the posterior distribution of the
parameter $\theta$ given the data.  Drawing samples from this
posterior distribution allows us to estimate and draw inferences about
the unknown parameter.  Under mild conditions, the Bernstein-von-Mises
theorem guarantees that the posterior distribution will concentrate
around the true parameter $\theta^*$, in which case we expect that the
credible intervals formed by sampling from the posterior should contain
$\theta^*$ with high probability.  This fact provides a lens for us to
assess the accuracy of our sampling procedure.

Define the vector $\yvec = \parenth{y_1, \ldots, y_\obs}\tp \in
\braces{0, 1}^\obs$ and let $\Xmat$ be the $\obs \times \dims $ matrix
with $x_i$ as $i^{\text{th}}$-row.  We choose the prior $\pi_0$ to be
a Gaussian distribution with zero mean and covariance matrix
proportional to the inverse of the sample covariance matrix
$\Sigma_\Xmat = \frac{1}{\obs} \Xmat \tp \Xmat$.  Plugging in the
formulas for the prior and likelihood, we find that the the posterior
density is given by
\begin{align*}
  \target(\theta) = \target(\theta \vert X, Y) \propto
  \exp\braces{\yvec \tp \Xmat \theta - \sum_{i=1}^\obs \log\parenth{1
      + e^{\theta \tp x_i}} - \scalar \vecnorm{\Sigma_\Xmat^{1/2}
      \theta}{2}^2},
\end{align*}
where $\scalar > 0$ is a user-specified parameter.  Writing $\target
\propto e^{-\myfun}$, we observe that the function $\myfun$ and its
derivatives are given by
\begin{align*}
  f(\theta) &= - \yvec\tp \Xmat \theta + \sum_{i=1}^\obs
  \log\parenth{1 + e^{\theta \tp x_i}} + \scalar
  \vecnorm{\Sigma_\Xmat^{1/2} \theta}{2}^2, \\ \nabla f(\theta) &= -
  \Xmat\tp \yvec + \sum_{i=1}^\obs \frac{x_i}{1 + e^{-\theta\tp x_i}}
  + \scalar \Sigma_\Xmat \theta, \quad \text{and}, \\
  \nabla^2
  f(\theta) &= \sum_{i=1}^\obs \frac{e^{-\theta\tp x_i}}{\parenth{1 +
      e^{-\theta\tp x_i}}^2} x_i x_i\tp + \scalar \Sigma_\Xmat.
\end{align*}
With some algebra, we can deduce that the eigenvalues of the Hessian
$\hessf$ are bounded between $\smoothness \defn \parenth{0.25 \obs +
  \scalar} \, \lambda_{\text{max}}(\Sigma_\Xmat)$ and $\scparam \defn
\scalar \, \lambda_{\text{min}}(\Sigma_\Xmat)$ where
$\lambda_{\text{max}}(\Sigma_\Xmat)$ and
$\lambda_{\text{min}}(\Sigma_\Xmat)$ denote the largest and smallest
eigenvalues of the matrix $\Sigma_\Xmat$.  We make use of these bounds
in our experiments.

As in~\cite{dalalyan2016theoretical}, we also consider a
preconditioned version of the method; more precisely, we first sample
from $\target_g \propto e^{-g}$ where $g(\theta) =
\myfun(\Sigma_\Xmat^{-1/2}\theta)$, and then transform the obtained
random samples $\theta_i \mapsto \Sigma_\Xmat^{1/2}\theta_i$ to obtain
samples from $\target$.  Sampling based on the preconditioned
distribution improves the condition number of the problem.  After the
preconditioning, we have the bounds $\smoothness_g \leq 0.25 \obs +
\scalar$ and $\scparam_g \geq \scalar$, so that the new condition
number is now independent of the eigenvalues of $\Sigma_\Xmat$.

\begin{figure}
  \begin{center}
    \begin{tabular}{cc}
      \widgraph{0.48\textwidth}{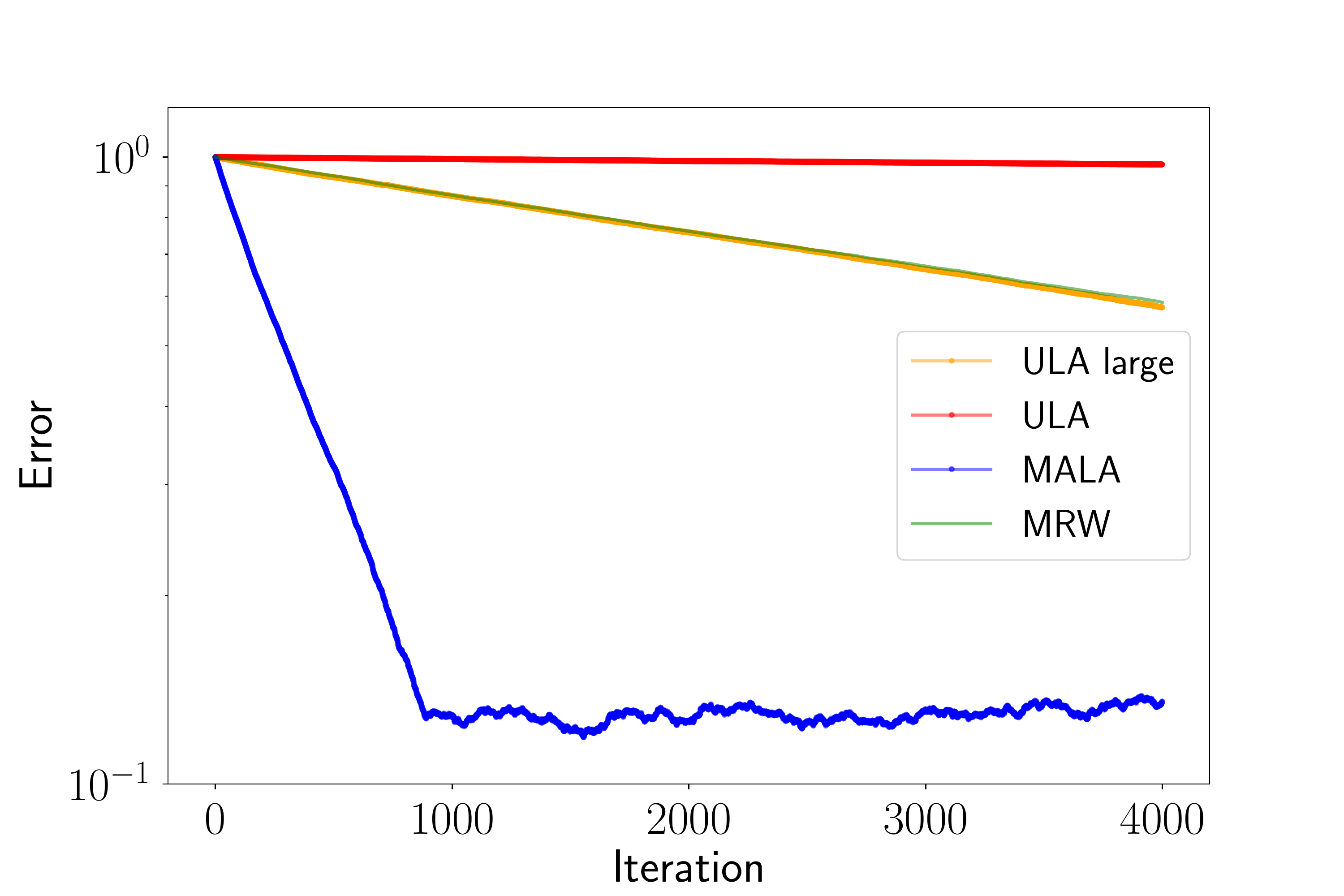} &
      \widgraph{0.48\textwidth}{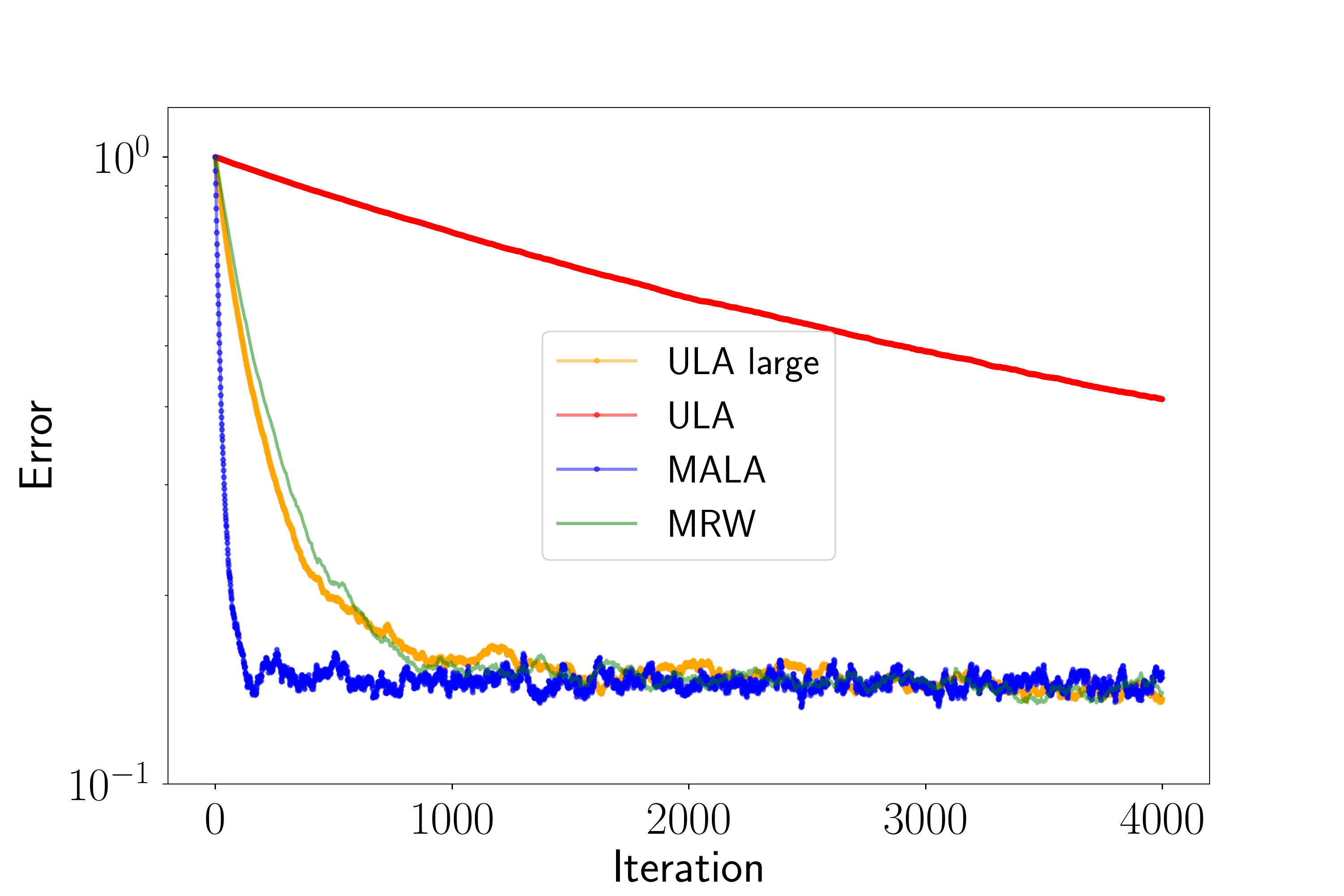}
      \\ (a) & (b)
    \end{tabular}
  \caption{Mean error as a function of iteration number.  (a) Without
    preconditioning.  \mbox{(b) With} preconditioning.}
  \label{fig:logistic_mean_error}
    \end{center}
\end{figure}

\begin{figure}[t]
  \begin{center}
    \begin{tabular}{cc}
      \widgraph{0.46\textwidth}{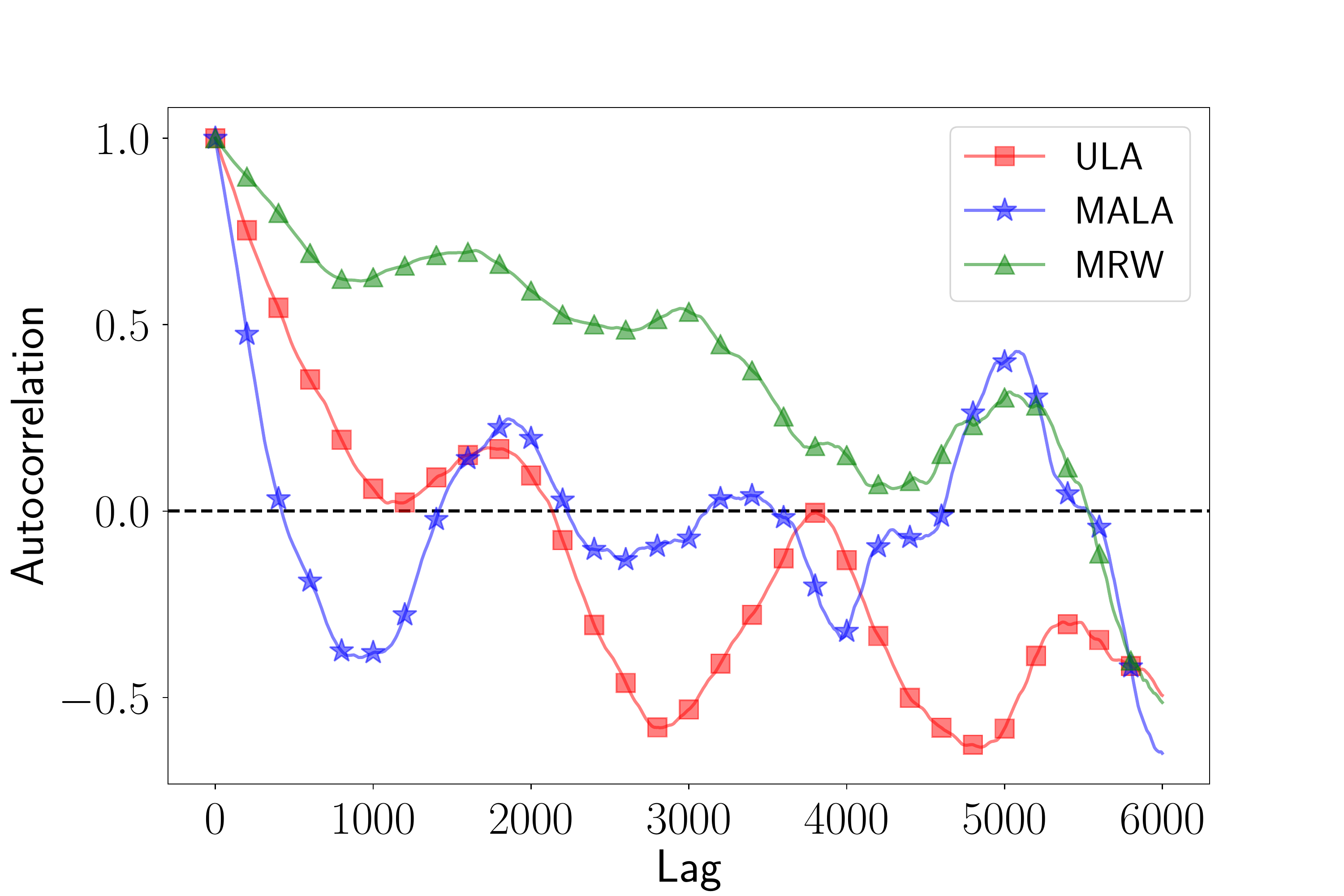} &
      \widgraph{0.46\textwidth}{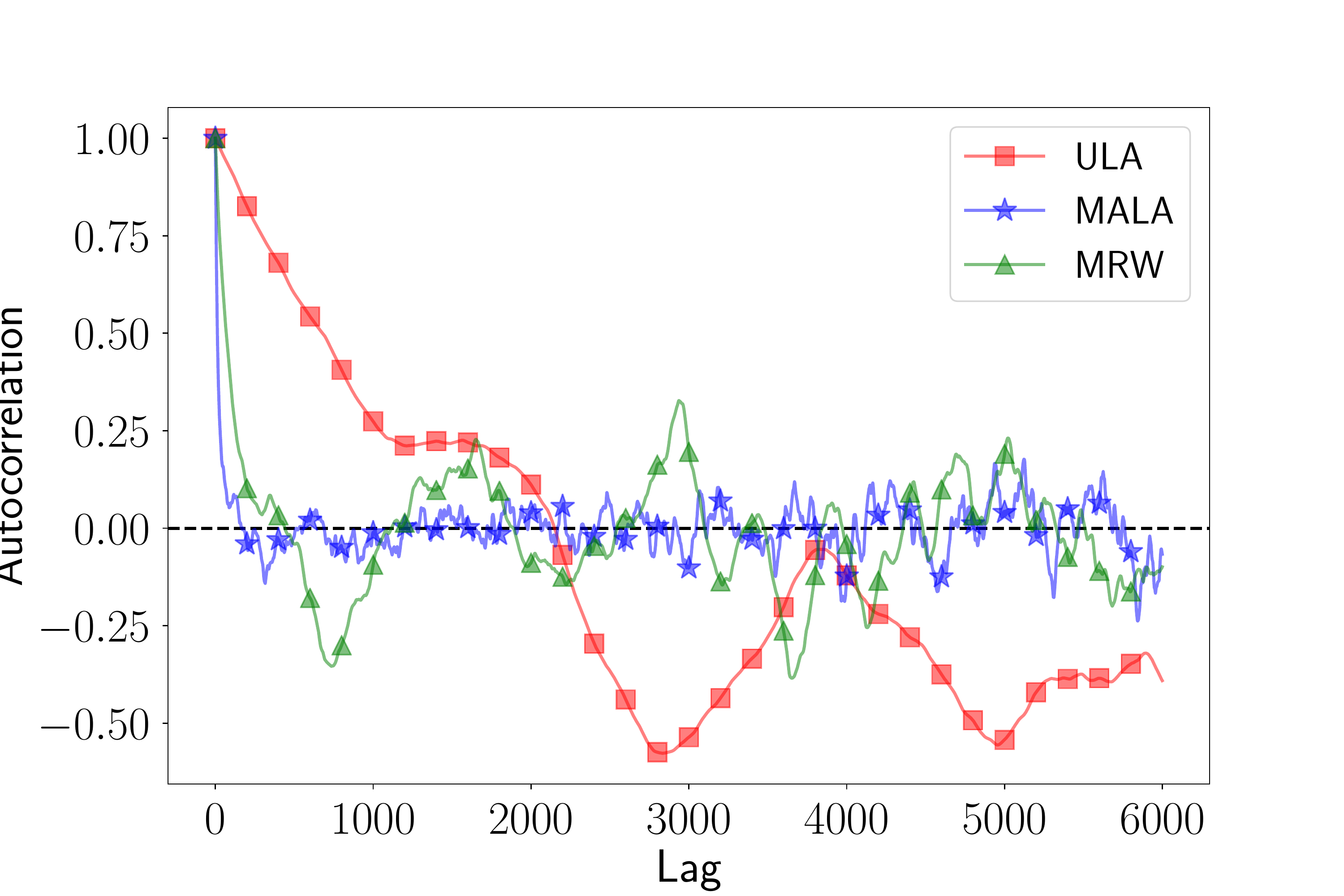}
      \\ (a) & (b)
    \end{tabular}
  \caption{Autocorrelation function plot of the first coordinate of the estimate as a function lag. The burn-in time for the plot is set to 300 iterations.  (a) Without
    preconditioning.  \mbox{(b) With} preconditioning.}
  \label{fig:autocorrelationplot_logistic_regression}
    \end{center}
\end{figure}

\begin{figure}[t]
  \begin{center}
    \begin{tabular}{cc}
      \widgraph{0.46\textwidth}{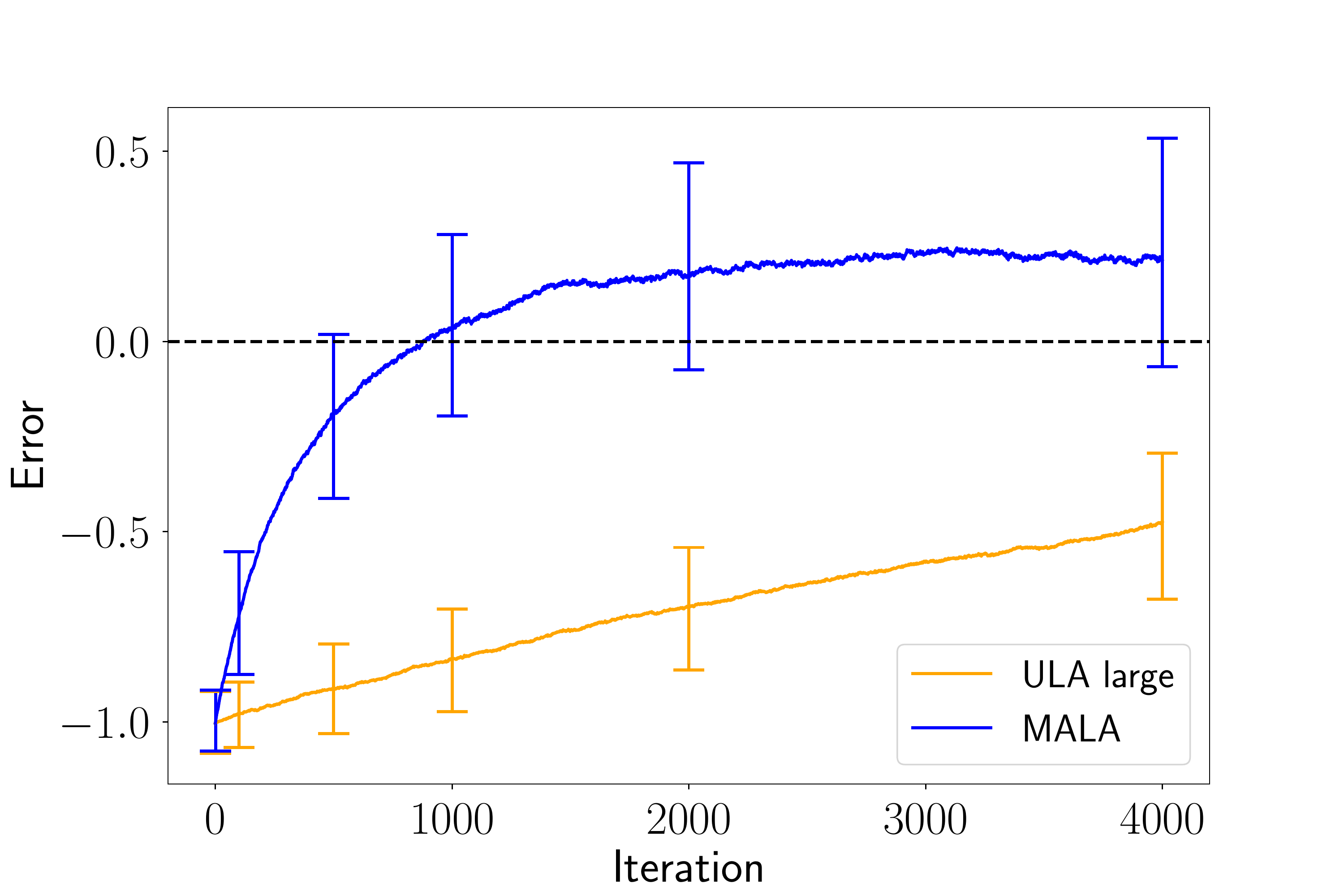} &
      \widgraph{0.46\textwidth}{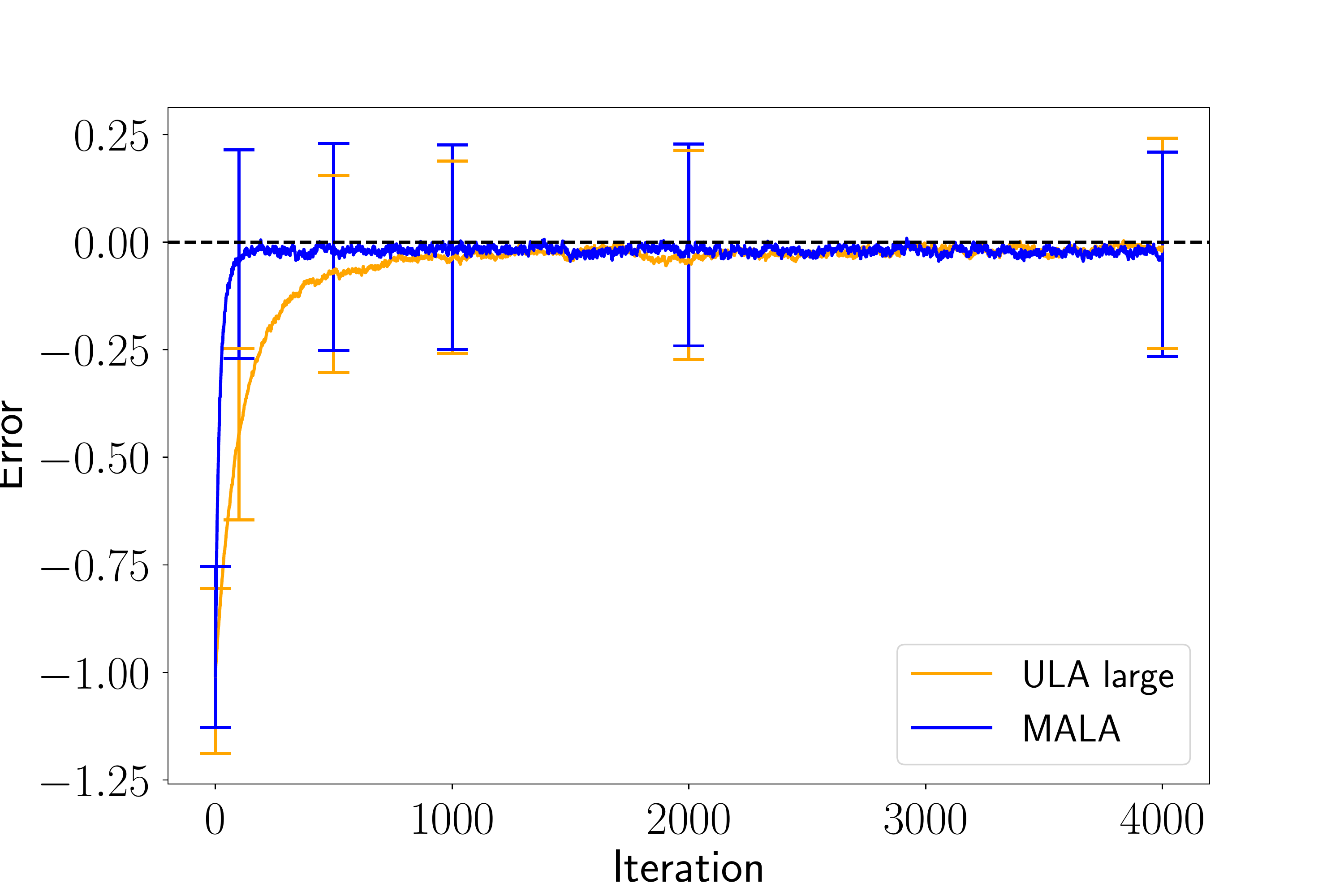}
      \\ (a) & (b)
    \end{tabular}
  \caption{Mean and $25\%$ and $75\%$ quantiles, with $\theta^*$
    subtracted, as a function of iteration number.  (a) Without
    preconditioning.  \mbox{(b) With} preconditioning.}
  \label{fig:logistic_quantile_error}
    \end{center}
\end{figure}

We randomly draw i.i.d. samples $\parenth{x_i, y_i}$ as follows.  Each
vector $x_i \in \real^\usedim$ is sampled i.i.d. Rademacher
components, and then renormalized to Euclidean norm.  given $x_i$, the
response $y_i$ is drawn from the logistic model~\eqref{eq:logistic}
with $\theta = \theta^* = \mathbf{1}_\dims = \parenth{1, \ldots,
  1}\tp$.  We fix $\dims = 2, \obs = 50$ and perform $\sampleobs =
1000$ experiments.  In order to sample from the posterior, we start
with the initial distribution as $\initial = \NORMAL(0,
\smoothness^{-1}\Ind_\dims)$.  As the first error metric, we measure
the $\ell_1$-distance between the true parameter $\theta^*$ and the
sample mean $\hat\theta_k $ of the random samples obtained from
simulating the Markov chains for $k$ iterations:
\begin{align*}
  e_k = \frac{1}{\dims} \|\hat{\theta}_k - \theta^*\|_1.
\end{align*}
Figure~\ref{fig:logistic_mean_error} provides a log-scale plot of this
error versus the iteration number.  Since there is always an
approximation error caused by the prior distribution, ULA with large
step-size ($\delta = 1.0$) can be used. However, our simulation shows
that it is still slower than MALA.  Also, the condition number
$\condition$ has a significant effect on the mixing time of ULA and
MRW.  Their convergence in the preconditioned case is significantly
better. Furthermore, the autocorrelation plots in
Figure~\ref{fig:autocorrelationplot_logistic_regression} and the plots
in Figure~\ref{fig:logistic_quantile_error} of the sample (across
experiments) mean and $25\%$ and $75\%$ quantiles, with $\theta^*$
subtracted, as a function of iterations suggest a similar story: MALA
converges faster than ULA and is less affected by the conditioning of the
problem.


\subsection{Step size vs accept-reject rate} 
\label{sub:step_size_vs_accept_reject_rate}
In this section, we provide a few simulations that highlight the effect
of step size for MALA and MRW. Note that our bounds from
Theorem~\ref{thm:mala_mixing} and \ref{thm:mrw_mixing} suggest a step size
choice of order $\dims^{-1}$ for both MALA and MRW, which in
turn led to the mixing time bounds of $\order{\dims}$.
These choices of step sizes arise
when we try to provide a worst-case control on the accept-reject step of
these algorithms. In particular, these choices ensure that the Markov chains
do not get stuck at a given state $x$, or equivalently, that the proposals
at any given state are accepted with constant probability. If instead, one
chooses a very large step size, the (worst-case) probability of acceptance
may decay exponentially with dimensions. Nonetheless, these
worst-case bounds may not hold, which would imply a faster mixing time for
these chains if a larger step size were to be used.

\begin{figure}[t]
  \begin{center}
    \begin{tabular}{cc}
      \widgraph{0.46\textwidth}{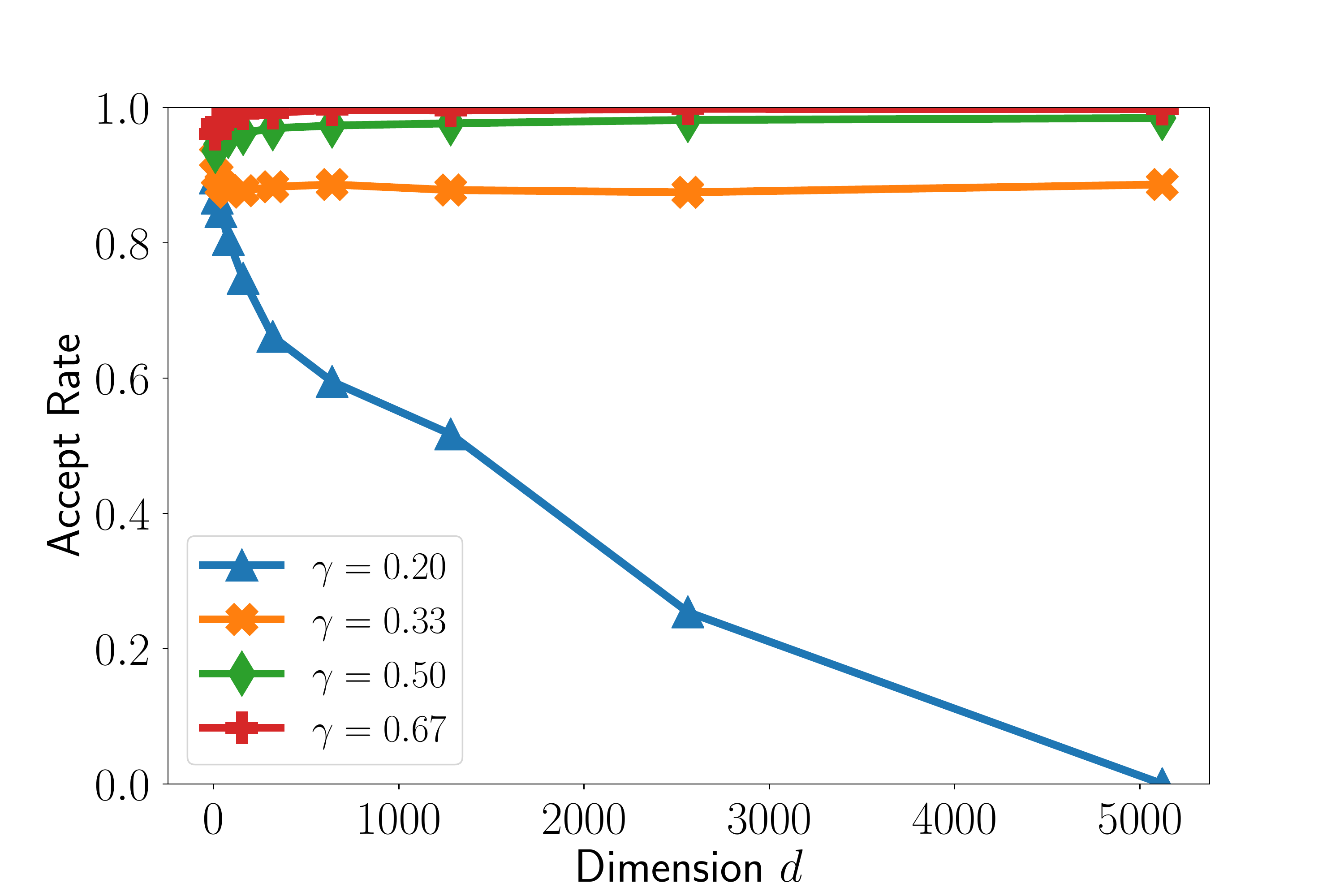}
      &
      \widgraph{0.46\textwidth}{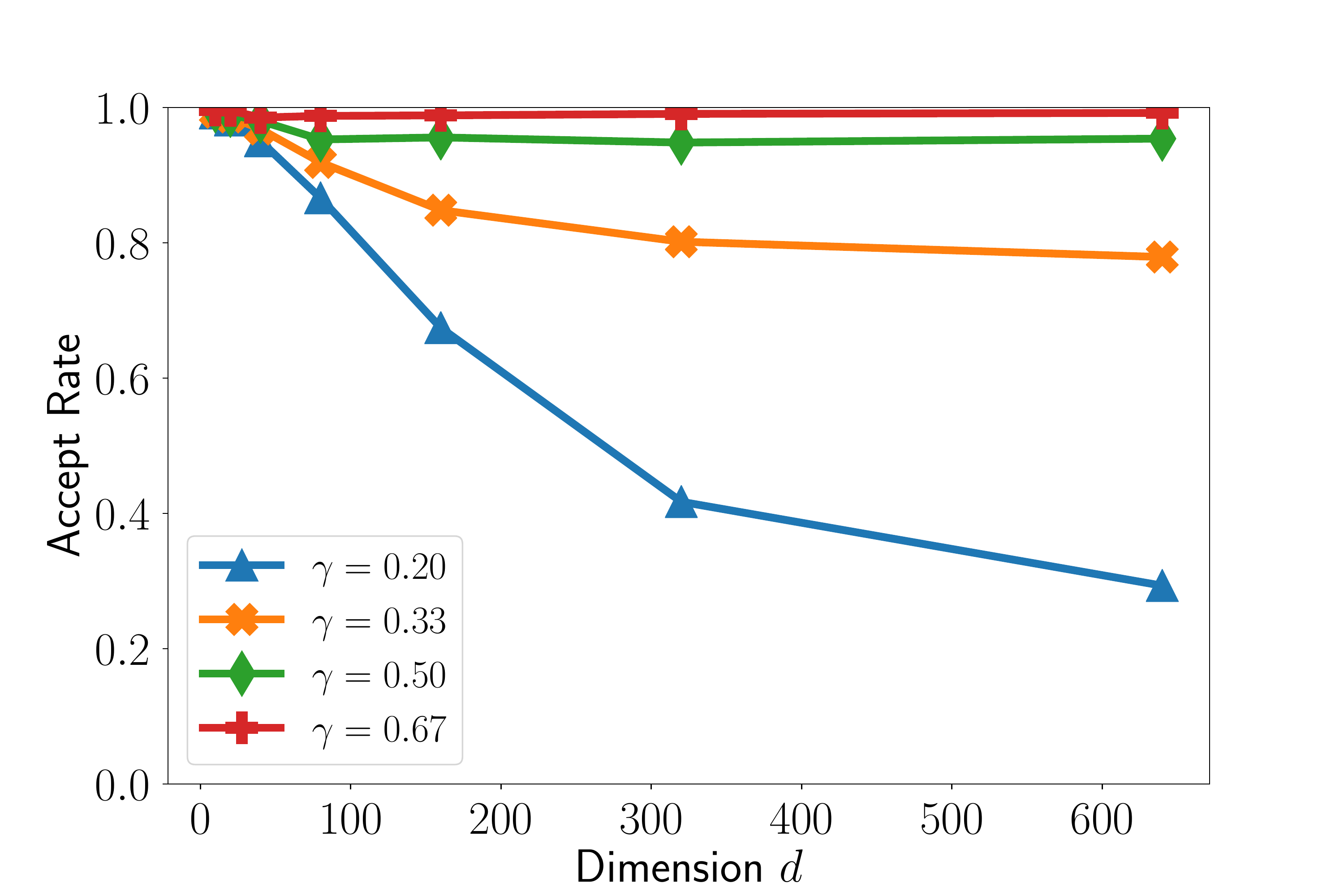}
      \\ (a) MALA: Non-isotropic Gaussian & (b) MALA: Bayesian logistic regression \\
      \widgraph{0.46\textwidth}{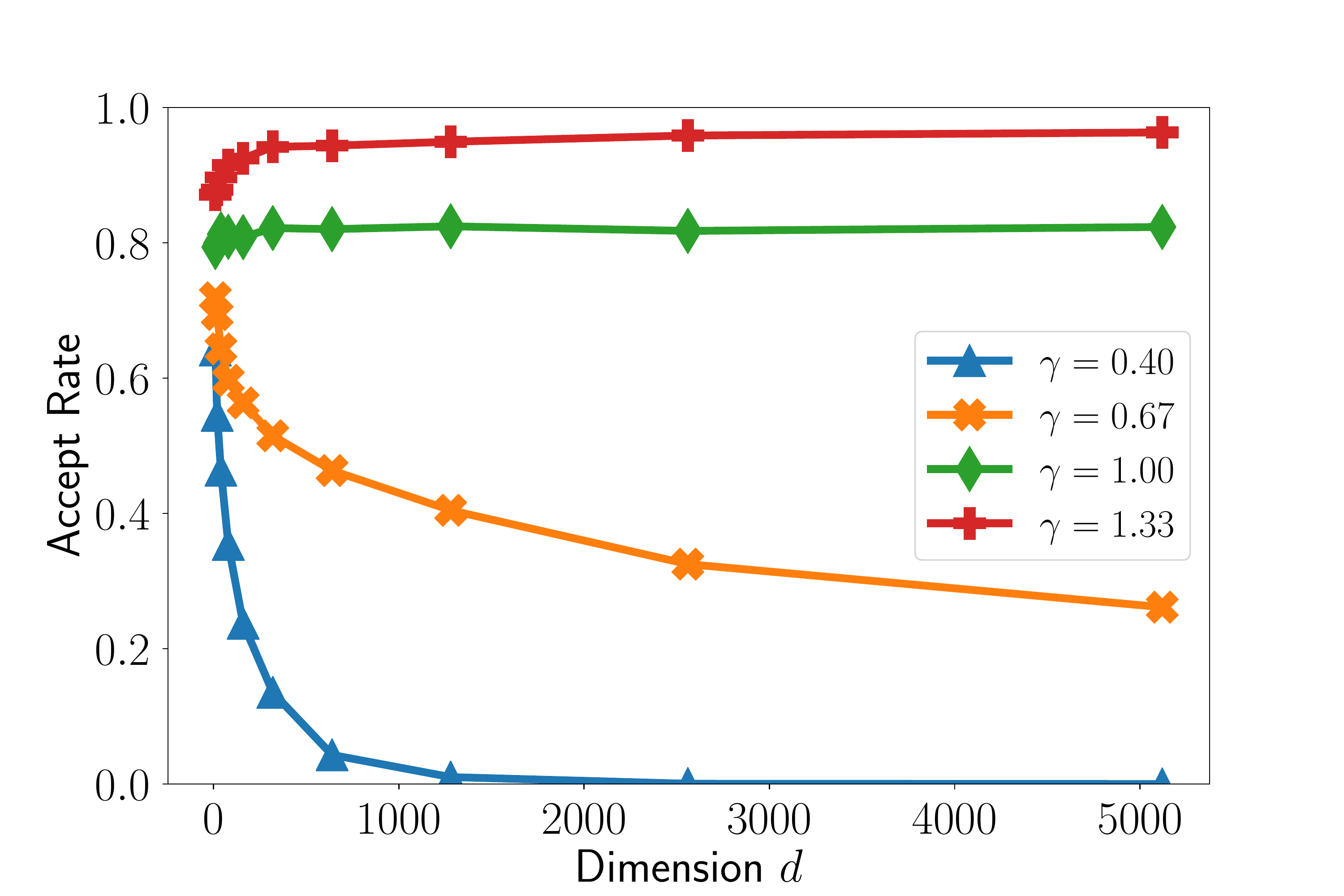}
      &
      \widgraph{0.46\textwidth}{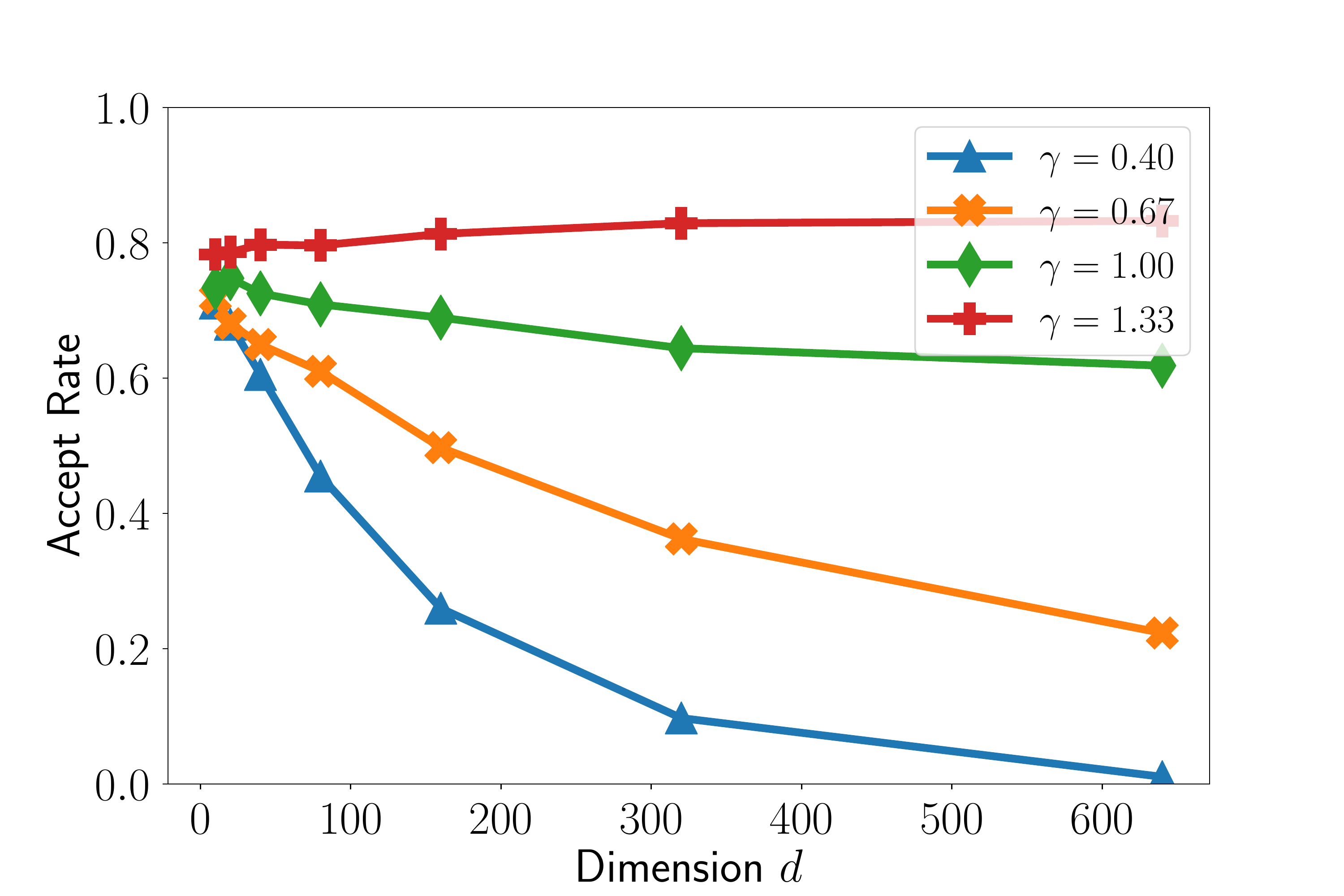}
      \\ (c) MRW: Non-isotropic Gaussian & (d) MRW: Bayesian logistic regression
    \end{tabular}
  \caption{Effect of large step size for accept-reject ratio
  for MALA and MRW. From panels (a) and (b), we see that
  for MALA the step size choice of
  $\dims^{-0.5}$ has a non-vanishing acceptance probability  rate for both
  cases.
  On the other hand, panels (c) and (d) show that for  MRW $\dims^{-1}$
  is a good choice for the step size.}
  \label{fig:step_size_role}
    \end{center}
\end{figure}

To check the validity of larger step sizes, we repeated a few experiments
discussed above, albeit with a larger
step size. In particular, we simulated the random walks for a wide-range
of step sizes $\dims^{-\gamma}$ for $\gamma \in \braces{0.2, 0.33, 0.5,
0.67}$ for MALA, and, $\gamma \in \braces{0.4, 0.67, 1, 1.33}$ for MRW. We ran these
chains for two different cases:
(a) Sampling from non-isotropic Gaussian density, discussed in
Section~\ref{sub:multivariate_gaussian_dimension_dependency},
and, (b) Posterior sampling in Bayesian logistic regression, discussed
in Section~\ref{sub:bayesian_logistic_regression}).
In Figure~\ref{fig:step_size_role}, we plot the average acceptance probability
for different step sizes $\dims^{-\gamma}$ as the dimension
$\dims$ increases.
These probabilities were computed as the average number of proposals accepted
over $100$ iterations after a manually tuned burn-in period, and further
averaged across $50$ independent runs.

We now remark on the observations from Figure~\ref{fig:step_size_role}.
We see that for MALA the acceptance probability for the step size choice
of $\dims^{-0.2}$ vanishes as $\dims$ increases. Indeed, the choice of $d^
{-0.5}$ appears to be a safe choice for both cases. In contrast, for MRW,
we need a smaller step size. From panels (c) and (d), we see that $\dims^
{-1}$ appears to be the correct choice to ensure that the
proposals are accepted with a constant probability when the dimension $\dims$
is large.

Informally, if a
step size choice of $\dims^{-\gamma}$ was to guarantee a non-vanishing
acceptance probability for MALA or MRW, our proof techniques imply a mixing
time bound of $\order{\dims^{\gamma}}$. Combining this argument with the
observations above, we suspect that the bounds for MALA from
Theorem~\ref{thm:mala_mixing} may not be tight, while for MRW the bounds
from Theorem~\ref{thm:mrw_mixing} are very likely to be tight.
Deriving a faster mixing time for MALA or establishing that the current
dimension dependency for MRW is tight, are interesting research
directions and we leave the further investigation of these questions for future work.


\section{Proofs}
\label{sec:proofs}

We now turn to the proofs of our main results.  In
Section~\ref{sub:auxiliary_results}, we begin by introducing some
background on conductance bounds, before stating three auxiliary
lemmas that underlie the proofs of our main theorems.  Taking these
three lemmas as given, we then provide the proof of
Theorem~\ref{thm:mala_mixing} in
Section~\ref{sub:proof_of_theorem_thm:mala_mixing}.
Sections~\ref{sub:proof_of_lemma_lem:hpb_region}
through~\ref{sub:proof_of_lemma_lemma:distribution_distance} are
devoted to the proofs of our three key lemmas, and we conclude with
the proof of Theorem~\ref{thm:mrw_mixing} in
Section~\ref{sub:proof_of_theorem_thm:mrw_mixing}.


\subsection{Conductance bounds and auxiliary results}
\label{sub:auxiliary_results}

Our proofs exploit standard conductance-based arguments for
controlling mixing times.  Consider an ergodic Markov chain defined by
a transition operator $\transition$, and let $\Target$ be its
stationary distribution.  For each scalar $\res \in (0, 1/2)$, we
define the $\res$-conductance
\begin{align}
  \label{eq:s_conductance}
  \conductance_\res \defn \inf_{\Target(\set) \in (\res, 1-\res)}
  \frac{\int_{\set}\transition_\myvec(\set^c)
    \target(\myvec)d\myvec}{\min\braces{\Target(\set)-\res,
      \Target(\set^c)-\res}}.
\end{align}
In this formula, the notation $\transition_u$ is shorthand for the
distribution $\transition(\diracdelta_u)$ obtained by applying the
transition operator to a dirac distribution concentrated on $u$.  In
words, the $\res$-conductance measures how much probability mass flows
across disjoint sets relative to their stationary mass.  By a
continuity argument, it can be seen that limiting conductance of the
chain is equal to the limiting value of $\res$-conductance---that is,
$\conductance = \lim_{\res\rightarrow 0}\conductance_\res$.

For a reversible lazy Markov chain with $\warmparam$-warm start,
Lov{\'a}sz~\citeyear{lovasz1999hit} (see
also~\cite{kannan1995isoperimetric}) proved that
\begin{align}
\label{eq:lovasz_tv_bound}
    \tvnorm{\transition^k( \initial)-\Target} \leq \warmparam\res +
     \warmparam \parenth{1- \frac{\sconductance^2}{2}}^k \leq
    \warmparam \res + {\warmparam} e^{-k {\sconductance^2}/{2}} \quad
    \mbox{for any $\res \in \big(0, \frac{1}{2} \big)$.}
\end{align}
In order to make effective use of this lower bound, we need to lower
bound the $s$-conductance $\sconductance$, and then choose the
parameter $s$ so as to optimize the tradeoff between the two terms
in the bound.  We now state some auxiliary results that are useful.

We start with a result that shows that the probability mass of any
strongly log concave distributions is concentrated in a Euclidean ball
around the mode.  For each $\res \in (0, 1)$, we introduce the
Euclidean ball
\begin{align}
\label{eq:defn_truncall}
\truncball_\res = \mathbb{B}\parenth{\xstar,
  \radius(\res)\sqrt\frac{\dims}{\scparam}}
\end{align}
where the function $\radius$ was previously defined in
equation~\eqref{eq:defn_r}, and $\xstar \defn \arg \max \limits_{x \in
  \real^\usedim} \target(x)$ denotes the mode.


\begin{lemma}
\label{lem:hpb_region}
For any $\res \in \big(0, \frac{1}{2} \big)$, we have
$\Target(\truncballres) \geq 1 - \res$.
\end{lemma}
\noindent See Section~\ref{sub:proof_of_lemma_lem:hpb_region} for the
proof of this claim.

\vspace*{.1in}

In order to establish the conductance bounds inside this ball, we
first prove an extension of a result by~\cite{lovasz1999hit}.
The next result provides a lower bound on the
flow of Markov chain with transition distribution
$\transition_x$ and strongly log concave target
distributions~$\Target$.
Similar results have been used in several
prior works to establish fast mixing of several random walks like ball walk,
Hit and run~\citep{lovasz1999hit,lovasz2006hit,lovasz2007geometry}, Dikin
walk~\citep{narayanan2016randomized} and Vaidya and John walks~\citep{chen2018fast}.
\begin{lemma}
\label{lem:conductance_isoperimetry}
Let $\convex$ be a convex set such that
$\tvnorm{\transition_x-\transition_y} \leq 1-\lovtv$ whenever $x, y
\in \convex$ and $\enorm{x-y}\leq \lovdis$.  Then for any measurable
partition $\set_1$ and $\set_2$ of $\realdim$, we have
\begin{align}
  \label{eq:conductance_isoperimetry}
  \int_{\set_1}\transition_{u}(\set_2) \target(u)du \geq
  \frac{\lovtv}{4}\min \braces{1, \frac{\log 2 \cdot \lovdis \cdot
      \Target^2(\convex) \cdot \sqrt{\scparam}}{8}}
  \min\braces{\Target(\set_1 \cap \convex), \Target(\set_2 \cap
    \convex)}.
\end{align}
\end{lemma}
\noindent See
Section~\ref{sub:proof_of_lemma_lem:conductance_isoperimetry} for the
proof of this lemma.

We next introduce a few pieces of notations to state a MALA specific
result.  Define a function $\stepext: (0, 1) \times (0, 1) \rightarrow
\real_{+}$ as follows:
\begin{subequations}
  \begin{align}
\label{eq:step_size_function}
\stepext(\res, \tvscalar) &\defn \min\bigg\{
\frac{\sqrt\tvscalar}{8\sqrt{2} \radius(\res)}
\frac{\sqrt{\scparam}}{\smoothness\sqrt{\dims\smoothness}},
\frac{\tvscalar}{64\scalar_\tvscalar}\frac{1}{\smoothness\dims},
\frac{\tvscalar^{2/3}}{26(\scalar_\tvscalar \radius^2(\res) )^{1/3}}
\frac{1}{\smoothness}\parenth{\frac{\scparam}{\smoothness\dims^2}}^{1/3}
\bigg\},\\
\text{where}\quad \scalar_\tvscalar &\defn 1+2\sqrt{\log(16/\tvscalar)} +
2\log(16/\tvscalar),
\label{eq:alpha_epsilon}
\end{align}
and the function $\radius$ was defined in equation~\eqref{eq:defn_r}.
\end{subequations}

In the next lemma, we show two important
properties for MALA: (1) the proposal distributions of MALA at two
different points are close if the two points are close, and (2) the
accept-reject step of MALA is well behaved inside the ball
$\truncballres$ provided the step size is chosen carefully.  Note that
for MALA, the proposal distribution of the chain at $x$ is given by
\begin{align}
\label{eq:proposal_distribution}
    \proposal_x^{\tagmala(\step)} = \NORMAL(\mean_x, 2\step
    \Ind_\dims), \quad \text{where }\quad \mean_x = x-\step \gradf(x).
\end{align}
We use $\transition_x^{\tagmala(\step)}$ to denote the
transition distribution of MALA.
\begin{lemma}
 \label{lemma:distribution_distance}
\begin{subequations}
For any step size $\step \in \big(0, \frac{2}{\smoothness} \big]$, the
  MALA proposal distribution satisfies the bound
\begin{align}
  \label{eq:proposal_difference}
\sup_{ \substack{x,y \in \real^\usedim \\ x \neq y}}
\frac{\tvnorm{\proposal_x^{\tagmala(\step)} -
    \proposal_y^{\tagmala(\step)}}}{\enorm{x -y}} &\leq
\frac{1}{\sqrt{2\step}}.
\end{align}
Moreover, given scalars $\res \in (0, 1/2)$ and $\tvscalar \in (0,
1)$, then the MALA proposal distribution for any step size $\step \in
\big(0, \stepext(\res, \tvscalar) \big]$ satisfies the bound
\begin{align}
\label{eq:transition_difference}
\sup_{x \in \truncballres} \tvnorm{\proposal_x^{\tagmala(\step)} -
  \transition_x^{\tagmala(\step)}} \leq \frac{\tvscalar}{8},
\end{align}
where the truncated ball $\truncballres$ was defined in
equation~\eqref{eq:defn_truncall}.
\end{subequations}
\end{lemma}
\noindent See
Section~\ref{sub:proof_of_lemma_lemma:distribution_distance} for the
proof of this claim. With these results in hand, we are now equipped
to prove the mixing time bound for MALA.


\subsection{Proof of Theorem~\ref{thm:mala_mixing}}
\label{sub:proof_of_theorem_thm:mala_mixing}

At a high level, the proof involves three key steps.  Our first step
is to use Lemma~\ref{lemma:distribution_distance} to establish that
for an appropriate choice of step size, the MALA update has nice
properties inside a high probability region given by
Lemma~\ref{lem:hpb_region}.  The second step is to apply
Lemma~\ref{lem:conductance_isoperimetry} so as to obtain a lower bound
on the $\res$-conductance $\conductance_\res$ of the MALA update.
Finally, by making an appropriate choice of parameter $\res$, we
establish the claimed convergence rate.

In order to simplify notation, we drop the superscripts
$\tagmala(\step)$ from our notation---that is, we use $\transition_x$
and $\proposal_x$, respectively, to denote the transition and proposal
distributions at $x$ for MALA, each with step size $\step$.  By
applying the triangle inequality, we obtain the upper bound
\begin{align}
\label{eq:triangle_inequality_tv_distance}
 \tvnorm{\transition_x-\transition_y} \leq
 \tvnorm{\proposal_x-\transition_x} + \tvnorm{\proposal_x-\proposal_y}
 + \tvnorm{\proposal_y-\transition_y}.
\end{align}
Now applying claim~\eqref{eq:proposal_difference} from
Lemma~\ref{lemma:distribution_distance} guarantees that
\begin{align*}
  \tvnorm{\proposal_x-\proposal_y} \leq \tvscalar/\sqrt{2} \qquad
  \mbox{ for all $x, y \in \realdim$ such that $\enorm{x-y} \leq
    \tvscalar\sqrt\step$.}
\end{align*}
Furthermore, for any $\step \leq \stepext(\res, \tvscalar)$, the
bound~\eqref{eq:transition_difference} from
Lemma~\ref{lemma:distribution_distance} implies that
\mbox{$\tvnorm{\proposal_x - \transition_x} \leq \tvscalar/8$} for any
$x \in \truncballres$.  Plugging in these bounds in the
inequality~\eqref{eq:triangle_inequality_tv_distance}, we find that
\begin{align*}
\tvnorm{\transition_x - \transition_y} \leq 1 - (1-\tvscalar) \quad
\mbox{ $\forall \ x, y \in \truncballres$ such that $\enorm{x-y} \leq
  \tvscalar\sqrt{\step}$.}
\end{align*}
Thus, the transition distribution $\transition_x$ satisfies the
assumptions of \mbox{Lemma~\ref{lem:conductance_isoperimetry}} for
\begin{align}
\label{eq:values_of_rho_delta_rs}
 \convex = \truncballres,\quad \lovtv = (1-\tvscalar)\quad\text{ and
 }\quad\lovdis = \tvscalar\sqrt{\step}.
\end{align}

We now derive a lower bound on the $\res$-conductance of MALA.
Choosing a measurable set $\set$ such
that $\Target(\set) >\res$ and substituting the terms from
equation~\eqref{eq:values_of_rho_delta_rs} in the
inequality~\eqref{eq:conductance_isoperimetry}, we find that
\begin{align*}
\int_{\set}\transition_{u}(\set^c) \target(u) du & \geq
\frac{(1-\tvscalar)}{4}\min \braces{1, \frac{\log2 \cdot
    \tvscalar\sqrt{\step} \cdot \Target^2(\truncballres) \cdot
    \sqrt{\scparam}}{8}}\cdot \\
    &\quad\quad\quad\cdot \min\braces{\Target(\set \cap
  \truncball_\res), \Target(\set^c \cap \truncball_\res)} \\
& \stackrel{(i)}{\geq} \frac{(1-\tvscalar)\tvscalar\sqrt{\step} \cdot
  \Target^2( \truncballres) \cdot \sqrt{\scparam}}{64} \min \braces{
  \Target(\set) - \res, \Target(\set^c) - \res}.
\end{align*}
In this argument, inequality (i) follows from the facts that $\log 2
\geq 1/2$ and $\Target(\set), \Target(\set^c) > \res$.  Moreover, we
have applied Lemma~\ref{lem:hpb_region} to find that
$\Target(\truncballres) \geq 1-\res$ and hence
\begin{align*}
  \Target(\myset\cap\truncballres) = \Target(\myset) - \Target(\myset
  \cap \truncballres^c) \geq \Target(\myset)-\res \quad \mbox{for
    $\myset \in \braces{\set, \set^c}$.}
\end{align*}
We have also assumed that the second argument of the minimum is less
than $1$.  Applying the definition~\eqref{eq:s_conductance} of
$\conductance_\res$ for MALA, we find that
\begin{align}
\label{eq:cond_bound_epsilon}
\conductance_\res^{\tagmala\footnotesize{(\step)}} \geq
\frac{(1-\tvscalar) \tvscalar \cdot \Target^2(\truncballres) \cdot
  \sqrt{\step \scparam}}{64}, \quad \text{ for any } \step \leq
\stepext(\res, \tvscalar).
\end{align}

By making a suitable choice of $\res$, we can now complete the proof.
Using Lemma~\ref{lem:hpb_region}, we have that
$\Target(\truncball_{\threshold/2}) \geq 1- \threshold/2 \geq 1/2$ for
any $\threshold \in (0, 1)$.  Applying the
definition~\eqref{eq:alpha_epsilon} of $\scalar_\tvscalar$, we obtain
that $\scalar_{1/2} \leq 12$.  Using this fact and the
definitions~\eqref{eq:defn_h} and \eqref{eq:step_size_function} for
the functions $\stepfun(\cdot)$ and $\stepext(\cdot, \cdot)$, it is
straightforward to verify that $\UNICON\stepfun(\threshold/(2\warmparam))
\leq \stepext(\threshold/(2\warmparam), 1/2)$, for an appropriate
choice of universal constant $\UNICON$.  Substituting in $\res =
\threshold/(2\warmparam)$, $\tvscalar = 1/2$, and $\step =
\UNICON\stepfun(\threshold/(2\warmparam))$, and also making use of the
lower bound $\Target(\truncball_{\threshold/2\warmparam}) \geq 1/2$ in
the bound~\eqref{eq:cond_bound_epsilon}, we find that
$\conductance_{\threshold/2\warmparam}^{\tagmala\footnotesize{(\step)}}
\geq \unicontwo \sqrt{\scparam\step}$ for some universal constant
$\unicontwo$.  Using the convergence rate~\eqref{eq:lovasz_tv_bound},
we obtain that
\begin{align}
\label{eq:mh_bound_mixing_time}
\tvnorm{\transition_{\tagmala(\step)}^k(\initial)-\Target}
\leq \warmparam \frac{\threshold}{2\warmparam}
  + {\warmparam} e^{-k \scparam\step /\unicontwo}
\leq \threshold \quad\text{ for all }\quad k \geq
\frac{\unicontwo}{\scparam\step} \cdot
\log\parenth{\frac{2{\warmparam}}{\threshold}},
\end{align}
for a suitably large constant $\unicontwo$.  Substituting the
expression~\eqref{eq:defn_h} for \mbox{$\step = \UNICON
\stepfun(\threshold/(2\warmparam))$}, yields the claimed bound on mixing
time.


\subsection{Proof of Lemma~\ref{lem:hpb_region}}
\label{sub:proof_of_lemma_lem:hpb_region}

The proof consists of two main steps.  First, we establish that the
distribution $\Target$ is sub-Gaussian, which then guarantees
concentration around the mean.  Second, we show that the mean and the
mode of the distribution $\Target$ are not far apart.  Combining these
two claims yields a high probability region around the mode $\xstar$.

Let $x$ denote the random variable with distribution $\Target$ and
mean $\xbar = \Exs_{x\sim\Target}\brackets{x}$.  We claim that
$x-\xbar$ is a sub-Gaussian random vector with parameter
$1/\sqrt\scparam$, meaning that
\begin{align*}
\Exs_x\brackets{e^{\myvec\tp(x-\xbar)}}\leq
e^{{\enorm{\myvec}^2}/(2\scparam)} \quad \text{ for any vector }
\myvec \in \realdim.
\end{align*}
In order to prove this claim, we make use of a result due to~\cite{harge2004convex}
(Theorem 1.1), which we restate here.  Let
\mbox{$y \sim \NORMAL(\mu, \Sigma)$} with density $e$ and $x$ be
a random variable with density function $ q \cdot e $ where $q$
is a log-concave function.  Then for any convex function $g:\realdim
\mapsto \real$ we have
\begin{align}
\label{eq:harge_bound}
\Exs_x\brackets{g(x-\Exs[x])} \leq \Exs_y\brackets{g(y-\Exs[y])}.
\end{align}

From Lemma~\ref{lemma:strong_convexity}\ref{item:sc_convex_quadratic}
we have that $x \mapsto \myfun(x) -
\frac{\scparam}{2}\enorm{x-\xstar}^2$ is a convex function.  Thus we
can express the density $\target$ as the
product of a log concave function
and the density of a random variable with distribution
$\NORMAL(\xstar, \Ind_\dims/\scparam)$.  Letting $ y \sim
\NORMAL(\xstar, \Ind_\dims/\scparam)$ and noting that $g(z) \defn
e^{\myvec\tp z}$ is a convex function for each fixed vector $u$,
applying the Harg\'{e} bound~\eqref{eq:harge_bound} yields
\begin{align*}
\Exs_x\brackets{e^{\myvec\tp(x-\xbar)}} \leq
\Exs_y\brackets{e^{\myvec\tp(y-x^*)}} \stackrel{(i)}{\leq}
e^{\enorm{\myvec}^2/2\scparam}.
\end{align*}
Here inequality~(i) follows from the fact that the random vector
$y-\xstar$ is sub-Gaussian with parameter $1/\sqrt{\scparam}$.

Using the standard tail bounds for quadratic forms for sub-Gaussian
random vectors~(e.g., Theorem~1 by~\citealt{hsu2012tail}), we find that
\begin{align}
\label{eq:subgaussian_bound}
\Prob_{x\sim \Target}\brackets{\enorm{x-\xbar}^2 > \frac{\dims
  }{\scparam} \parenth{1 + 2\sqrt{\frac{\tailboundscalar}{\dims}} +
    2\frac{\tailboundscalar}{\dims}} } \leq e^{-\tailboundscalar}.
\end{align}
Define $\mathcal{B}_1 \defn \mathbb{B}\parenth{\xbar,
  \sqrt{\frac{\dims}{\scparam}} \cdot \tilde{\radius}(\res)}$ where
$\tilde{\radius}(\res) = 1 +
2\max\braces{\parenth{\frac{\log(1/\res)}{\dims}}^{0.25},
  \sqrt{\frac{\log(1/\res)}{\dims}}} $.  Observe that
$\tilde{\radius}(\res)^2\ {\geq}\ {{1 +
    2\sqrt{\frac{\log(1/\res)}{\dims}} +
    2\frac{\log(1/\res)}{\dims}}}$ and consequently the
bound~\eqref{eq:subgaussian_bound} implies that
\mbox{$\Target\parenth{\mathcal{B}_1} = \Prob_{x\sim
    \Target}\brackets{x \in \mathcal{B}_1} \geq 1-\res$}.  Now
applying triangle inequality, we obtain that
\begin{align*}
\mathcal{B}_1 \subseteq \mathbb{B}\parenth{\xstar,
  \enorm{\xbar-\xstar} +\sqrt{\frac{\dims}{\scparam}}\cdot
  \tilde{\radius}(\res)} \rdefn \mathcal{B}_2
\end{align*}
From Theorem~1~by~\cite{durmus2016high}, we have that
$\Exs_{x\sim\Target}\enorm{x-\xstar}^2\leq {\dims/\scparam}$.  Using
Jensen inequality twice, we find that
\begin{align}
\label{eq:var_target}
    {\enorm{\xbar - \xstar}} =
    \enorm{\Exs_{x\sim\Target}\brackets{x}-\xstar} \leq
          {\Exs_{x\sim\Target}\enorm{x-\xstar}} \leq
          \sqrt{\Exs_{x\sim\Target}\enorm{x-\xstar}^2} \leq
          \sqrt{\frac{\dims}{\scparam}}.
\end{align}
Noting the relation $\radius(\res) = 1 + \tilde{\radius}(\res)$, we
thus obtain that $ \enorm{\xbar-\xstar}
+\sqrt{\frac{\dims}{\scparam}}\cdot \tilde{\radius}(\res)
\leq\radius(\res)\sqrt{\frac{\dims}{\scparam}} $ and consequently
$\mathcal{B}_1 \subseteq \mathcal{B}_2 \subseteq \truncballres$.  As a
result, we obtain $\Target\parenth{\truncballres} \geq
\Target\parenth{\mathcal{B}_1} \geq 1-\res$ as claimed.


{\comment{
\paragraph{Older Proof}
\label{par:older_proof}

Without loss of generality by recentering as needed, we can assume
that $\myfun(\xstar) = 0$.  Such an assumption is possible because
substituting $\myfun(\cdot)$ by $\myfun(\cdot) + \scalar$ for any
scalar $\scalar$ leaves the distribution $\Target$ unchanged.  Since
$\myfun$ is $\scparam$-strongly convex and $\smoothness$-smooth,
applying
Lemma~\ref{lemma:strong_convexity}\ref{item:sc_quadratic_bound} and
Lemma~\ref{lemma:smoothness}\ref{item:smooth_quadratic_bound}, we
obtain that
\begin{align*}
    \frac{\smoothness}{2} \enorm{x-\xstar}^2 \geq \myfun(x)
    \geq \frac{\scparam}{2} \enorm{x-\xstar}^2.
\end{align*}
Let $C_\Target$ denote the normalization
constant for the density $\target$, i.e., $\target(x) = e^{-\myfun(x)}/C_\Target$.
Then we have
\begin{align}
\label{eq:lower_bound_normalization}
    C_\Target = \int_{\realdim} e^{-\myfun(x)} dx
    \geq \int_{\realdim} e^{-\frac{\smoothness}{2}
     \enorm{x-\xstar}^2} dx
     = \parenth{\frac{\smoothness}{2\pi}}^{-\dims/2}.
\end{align}
Let $\gaussiandistribution$ denote the
multivariate Gaussian distribution $\NORMAL(\xstar, \scparam^{-1}\Ind_\dims)$.
Observe that
\begin{align*}
    \Target(\myset)
    = \frac{\int_\myset e^{-\myfun(x)dx}}{C_\Target}
    &\leq \frac{1}{C_\Target} \displaystyle
    \int_\myset e^{-\frac{\scparam\enorm{x-\xstar}^2}{2}}dx\\
    &= \frac{1}{C_\Target}\parenth{\frac{\scparam}{2\pi}}^{-\dims/2}
    \int_\myset \frac{1}{(2\pi\scparam^{-1})^{\dims/2}}
    e^{-\frac{\scparam\enorm{x-\xstar}^2}{2}}dx\\
    &\leq \parenth{\frac{\smoothness}{\scparam}}^{-\dims/2}
     \gaussiandistribution(\myset),
\end{align*}
where the last inequality follows from the
bound~\eqref{eq:lower_bound_normalization}.
Define \mbox{$\myset_{1+\tailboundscalar} =
\braces{x\big\vert \enorm{x-\xstar}^2
\geq \frac{\dims}{\scparam}(1+\tailboundscalar)}$}.
Then, using chi-square tail bounds we obtain
that $\gaussiandistribution\parenth{\myset_{1+\tailboundscalar}}
\leq 2e^{-\dims\tailboundscalar/4}$ for
 all $\tailboundscalar \geq 1$, and consequently
\begin{align*}
    \Target(\myset_{1+\tailboundscalar})
     \leq 2\parenth{\frac{\smoothness}{\scparam}}^{-\dims/2}
     e^{-\dims\tailboundscalar/4}
    \quad \text{ for all } t \geq 1.
\end{align*}
Setting $\tailboundscalar = \tailboundscalar_\res ={2
  \log\frac{\smoothness}{\scparam} + \frac{4}{\dims} \log
  \parenth{\frac{2}{\res}}}$ in this inequality we find that
$\Target(\myset_{1+\tailboundscalar_\res}) \leq \res$.  Using the fact
that $\sqrt{a+b} \leq \sqrt{a} + \sqrt{b}$ for all $a, b \geq 0$, we
obtain that $\radius(\res) \geq 1+ \sqrt{\tailboundscalar_\res}$
(where $\radius(\res)$ is defined in the equation~\eqref{eq:defn_r}).
Observing that $ \myset_{1+\tailboundscalar_\res}^c \subseteq
\truncballres$ implies the claim.  } }  


\subsection{Proof of Lemma~\ref{lem:conductance_isoperimetry}}
\label{sub:proof_of_lemma_lem:conductance_isoperimetry}
The proof of this lemma is based on the ideas employed in prior works to establish conductance
bounds, first for Hit-and-run~\citep{lovasz1999hit}, and since then for several
other random walks~\citep{lovasz2007geometry,narayanan2016randomized,chen2018fast}.
See the survey by~\cite{vempala2005geometric} for further details.

For our setting, a key ingredient is the following isoperimetric
inequality for log-concave distributions.  Let $\realdim = \setiso_1
\cup \setiso_2 \cup \setiso_3$ be a partition.
Let $y \sim \NORMAL(0, \sigma^2\Ind_\dims)$ with density $e$
and let $\Target$ be a distribution with a density given by
$q \cdot e$ where $q$ is a log-concave function.
Then \cite{cousins2014cubic} (Theorem~4.4) proved that
\begin{align}
    \label{eq:isoperimetry}
    \Target(\setiso_3) \geq \frac{\log 2 \cdot d(\setiso_1,
      \setiso_2)}{\sigma}
      \Target(\setiso_1) \Target(\setiso_2)
\end{align}
where $d(\setiso_1, \setiso_2) \defn \inf\braces{\enorm{x-y}\big \vert
  x \in \setiso_1, y \in \setiso_2}$.

We invoke this result for the truncated distribution $\truncTarget$
 with the density $\trunctarget$ defined as
\begin{align}
\label{eq:defn_trunc_target}
    \trunctarget(x) \defn
    \frac{1}{\displaystyle\int_{\convex}\target(y)dy}
    \target(x)\indicator_{\convex}(x) =
    \frac{1}{\displaystyle\int_{\convex}e^{-\myfun(y)}dy}
    e^{-\myfun(x)}\indicator_{\convex}(x),
\end{align}
where $\indicator_{\convex}(\cdot)$ denotes the indicator function for
the set $\convex$, i.e., we have $\indicator_{\convex}(x) = 1$ if $x
\in \convex$, and $0$ otherwise.
Let $\xstar = \arg\max \target(x) = \arg\min \myfun(x)$.
Observe that $\scparam$-strong-convexity of $\myfun$ implies that
$x \mapsto \myfun(x) - \frac{\scparam}{2}\enorm{x-\xstar}^2$ is a convex
function (Lemma~\ref{lemma:strong_convexity}\ref{item:sc_convex_quadratic}).
Noting that the function $\indicator_{\convex}(\cdot)$ is log-concave
and that log-concavity is closed under multiplication, we conclude that
$\trunctarget$ can be expressed as a product of log-concave function
and density of the Gaussian distribution
$\NORMAL\parenth{\xstar, \frac{1}{\scparam}\Ind_\dims}$.
Consequently, we can apply the
result~\eqref{eq:isoperimetry} with $\Target$ replaced
by $\truncTarget$ and $\sigma = 1/\sqrt{\scparam}$.

\begin{figure}[h]
 \begin{center}
    \widgraph{0.7\linewidth}{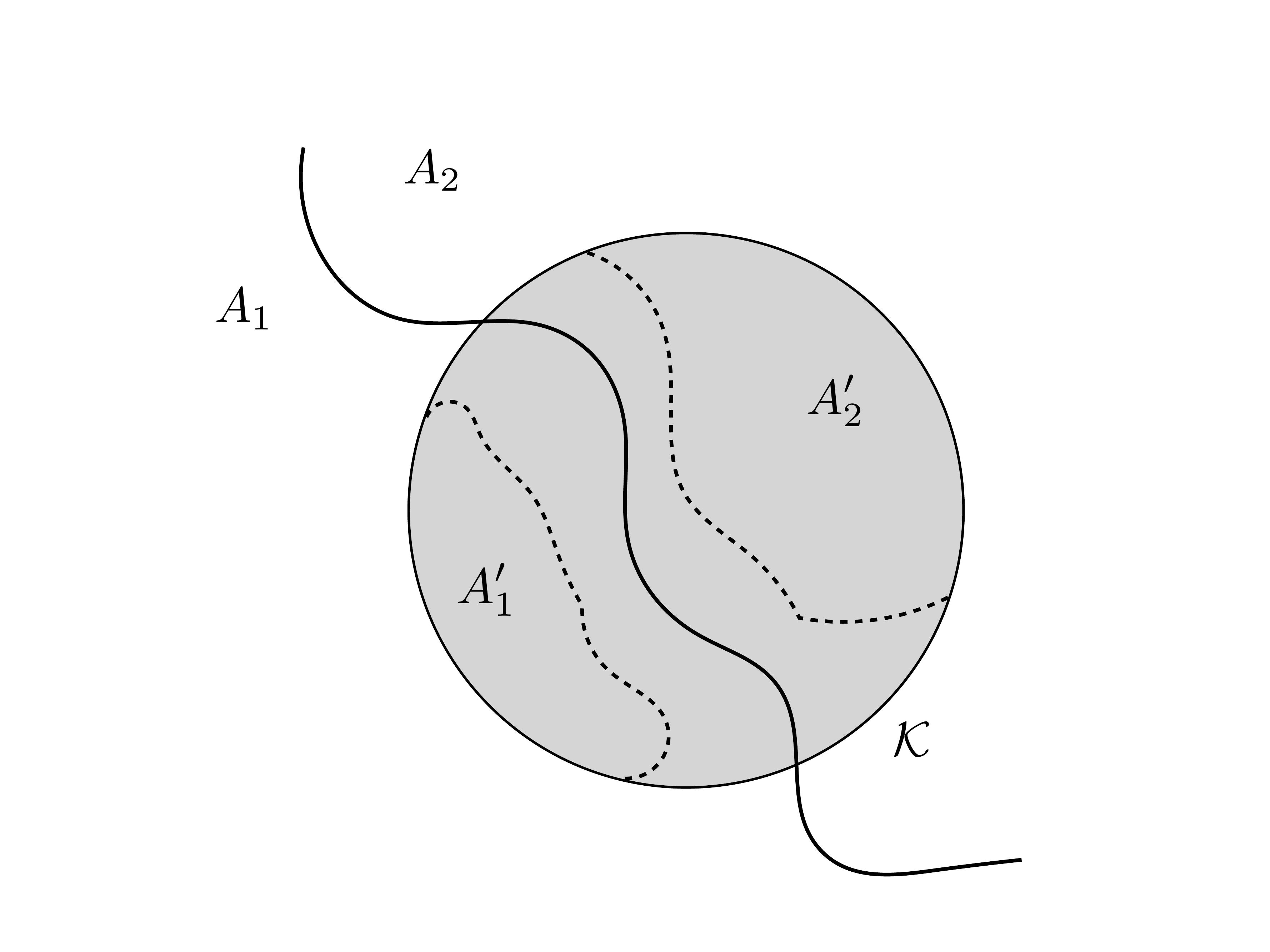}
    \caption{The sets $\set_1$ and $\set_2$ form a partition of
      $\realdim$, and we use $\convex$ to denote a compact convex
      subset.  The sets $\set_1'$ and $\set_2'$ are defined in
      equation~\eqref{eq:define_three_sets}.  }
    \label{fig:s_123}
 \end{center}
\end{figure}

We now prove the claim of the lemma.
Define the sets
\begin{align}
 \label{eq:define_three_sets}
  \set_1' \defn \braces{ u \in \set_1 \cap \convex \; \mid \;
    \transition_u(\set_2) < \frac{\lovtv}{2} \!}, \quad \set_2' \defn
  \braces{ v \in \set_2 \cap \convex \; \mid \;
    \transition_v(\set_1) < \frac{\lovtv}{2} },
\end{align}
along with the complement $\set_3' \defn \convex \backslash (\set_1'
\cup \set_2')$.  See Figure~\ref{fig:s_123} for an illustration.
Based on these three sets, we split our proof of the
claim~\eqref{eq:conductance_isoperimetry} into two distinct cases:
\begin{itemize}
  \item Case 1: $\Target(\set_1') \leq \Target(\set_1 \cap \convex)/2$
    or $\Target(\set_2') \leq \Target(\set_2 \cap \convex)/2$.
  \item Case 2: $\Target(\set_i') \geq \Target(\set_i \cap \convex)/2$
    for $i = 1, 2$.
\end{itemize}
Note that these cases are mutually exclusive, and cover all
possibilities.


\paragraph*{Case 1}

We have $\Target(\set_1\cap\convex \backslash \set_1') \geq
\Target(\set_1\cap\convex)/2$, then
\begin{align*}
  \int_{\set_1} \transition_u(\set_2) \target(u)du
  \stackrel{(i)}{\geq}\int_{\set_1\cap\convex \backslash
    \set_1'} \transition_u(\set_2) \target(u)du
    &\stackrel{(ii)}{\geq}
  \frac{\lovtv}{2} \Target(\set_1\cap\convex \backslash \set_1')\\
  &\stackrel{(iii)}{\geq} \frac{\lovtv}{4} \Target(\set_1 \cap
  \convex),
\end{align*}
which implies the claim~\eqref{eq:conductance_isoperimetry}.  In the
above sequence of inequalities, step (i) is trivially true;
step (ii) from the
definition~\eqref{eq:define_three_sets} of the set $\set_1'$, and step
(iii) from the assumption for this case.

A similar argument with the roles of $\set_1$ and $\set_2$ switched,
establishes the claim when $\Target(\set_2') \leq \Target(\set_2 \cap
\convex)/2$.


\paragraph{Case 2}
\label{par:case_2}

We have $\Target(\set_i') \geq \Target(\set_i \cap \convex)/2$ for
both $i=1$ and $2$.  For any $\myvec \in \set_1'$ and $\myvectwo \in
\set_2'$, we have that
\begin{align*}
  \tvnorm{\transition_\myvec- \transition_\myvectwo} \geq
  \transition_\myvec(\set_1) - \transition_\myvectwo(\set_1)
  \stackrel{(i)}{=} 1 - \transition_\myvec(\set_2) -
  \transition_\myvectwo(\set_1) > 1 - \lovtv,
\end{align*}
where step (i) follows from the fact that $\set_1 =
\realdim\backslash\set_2$ and thereby $\transition_\myvec(\set_1) = 1
- \transition_\myvec(\set_2)$.  Since $\myvec, \myvectwo \in \convex$,
the assumption of the lemma implies that $\enorm{\myvec-\myvectwo}
\geq \lovdis$ and consequently
\begin{align}
    \label{eq:d_S1_S2}
    d(\set_1', \set_2') \geq \lovdis.
\end{align}
We claim that
\begin{align}
  \label{eq:equality_of_T1_T2}
  \int_{\set_1}
  \transition_\myvec(\set_2)\target(\myvec)d\myvec =
  \int_{\set_2}
  \transition_\myvectwo(\set_1)\target(\myvectwo)d\myvectwo
\end{align}
We provide the proof of this claim at the end.
Assuming this claim as given, we now complete the proof.
Using equation~\eqref{eq:equality_of_T1_T2}, we have
\begin{align}
  \int_{\set_1}
  \transition_\myvec(\set_2)\target(\myvec)d\myvec &=
  \frac{1}{2}\parenth{ \int_{\set_1}
    \transition_\myvec(\set_2)\target(\myvec)d\myvec +
    \int_{\set_2}
    \transition_\myvectwo(\set_1)\target(\myvectwo)d\myvectwo}
  \notag \\
%
%
& \geq \frac{1}{4}\parenth{ \int_{\set_1\cap\convex\backslash\set_1'}
    \transition_\myvec(\set_2)\target(\myvec)d\myvec +
    \int_{\set_2\cap\convex\backslash\set_2'}
    \transition_\myvectwo(\set_1)\target(\myvectwo)d\myvectwo} \notag
  \\
& \stackrel{(i)}{\geq} \frac{\lovtv}{8}
  \Target(\convex\backslash(\set_1'\cup\set_2')),
        \label{eq:case_2_intermediate_1}
\end{align}
where step (i) follows from
the definition~\eqref{eq:define_three_sets} of the set
$\set_3'=\convex\backslash(\set_1'\cup\set_2')$.  Further, we have
\begin{align}
    \Target(\convex\backslash(\set_1'\cup\set_2')) & \stackrel{(i)}{=}
    \Target(\convex) \cdot
    \truncTarget(\convex\backslash\set_1'\backslash\set_2')
    \notag\\
& \stackrel{(ii)}{\geq} \Target(\convex) \cdot \frac{\log 2 \cdot
      d(\set_1',
      \set_2')}{1/\sqrt{\scparam}} \cdot
    \truncTarget(\set_1') \cdot \truncTarget(\set_2')
    \notag \\
    & \stackrel{(iii)}{\geq} \Target(\convex) \cdot \log 2 \cdot
      d(\set_1', \set_2') \cdot \sqrt{\scparam} \cdot
    \Target(\set_1') \cdot \Target(\set_2')
    \notag\\ &\stackrel{(iv)}{\geq} \Target(\convex) \cdot \log
      2\cdot\lovdis \cdot \sqrt{\scparam} \cdot \frac{1}{4} \cdot
    \Target(\set_1 \cap \convex)\cdot \Target(\set_2 \cap
      \convex).
    \label{eq:case_2_intermediate_2}
\end{align}
where step (i) follows from the
definition~\eqref{eq:defn_trunc_target} of the truncated
distribution~$\truncTarget$, step (ii) follows from applying the
isoperimetry~\eqref{eq:isoperimetry} for the distribution
$\truncTarget$ with $\sigma = 1/\sqrt{\scparam}$,
step (iii) from the definition of $\truncTarget$
and step (iv) from
inequality~\eqref{eq:d_S1_S2} and the assumption for this case.
Let $\scalar \defn \Target(\set_1 \cap \convex)/\Target(\convex)$.
Note that $\scalar \in [0, 1]$ and $\Target(\set_2\cap\convex)
/\Target(\convex) = 1 - \scalar$.
We have
\begin{align}
  \Target(\set_1 \cap \convex)\cdot \Target(\set_2 \cap
      \convex)
  &= \Target^2(\convex) \cdot \scalar (1-\scalar) \notag\\
  &\geq \Target^2(\convex) \cdot  \frac{1}{2}\min\braces{\scalar, 1-\scalar} \notag\\
  &= \Target(\convex) \cdot  \frac{1}{2}\min\braces{\Target(\set_1\cap\convex),
  \Target(\set_2\cap\convex)}
  \label{eq:product_to_min}
\end{align}
Putting the inequalities~\eqref{eq:case_2_intermediate_1},
\eqref{eq:case_2_intermediate_2} and \eqref{eq:product_to_min}
together, establishes the
claim~\eqref{eq:conductance_isoperimetry} of the lemma for this case.

We now prove our earlier claim~\eqref{eq:equality_of_T1_T2}.
Note that it suffices to prove that
\begin{align*}
    \int_{\set_1} \transition_\myvec(\set_2)\target(\myvec)d\myvec =
    \int_{\set_2}
    \transition_\myvectwo(\set_1)\target(\myvectwo)d\myvectwo.
\end{align*}
We have
\begin{align*}
  \int_{\set_2} \transition_u(\set_1) \target(u)du &
  \stackrel{\mathmakebox[\widthof{==}]{(i)}}{=} \int_{\realdim}
  \transition_u(\set_1) \target(u)du - \int_{\set_1}
  \transition_u(\set_1) \target(u)du \\
& \stackrel{\mathmakebox[\widthof{==}]{(ii)}}{=} \stationary(\set_1) -
  \int_{\set_1} \transition_u(\set_1)\target(u)du \\
& \stackrel{\mathmakebox[\widthof{==}]{}}{=} \int_{\set_1} \target(u)
  du - \int_{\set_1} \transition_u(\set_1) \target(u) du
  \\
& \stackrel{\mathmakebox[\widthof{==}]{(iii)}}{=}
  \int_{\set_1}\transition_u(\set_2) \target(u) du,
\end{align*}
where steps~(i) and (iii) (respectively)
follow from the fact that
$\set_1 = \realdim\backslash\set_2$ and the consequent
fact that $1-\transition_u(\set_1) =\transition_u(\set_2)$,
and step~(ii) follows from the fact that
$\target$ is the stationary density for
the transition distribution $\transition_x$ and thereby
$\int_{\realdim} \transition_u(\set_1) \target(u)du = \Target(\set_1)$.





\subsection{Proof of Lemma~\ref{lemma:distribution_distance}}
\label{sub:proof_of_lemma_lemma:distribution_distance}

We prove each claim of the lemma separately.  To simplify notation, we
drop the superscript from our notations of distributions
$\transition_x^{\tagmala(\step)}$ and $\proposal_x^{\tagmala(\step)}$.


\subsubsection{Proof of claim~\eqref{eq:proposal_difference}}
\label{ssub:proof_of_claim_eq:proposal_difference}

In order to bound the total variation distance
$\tvnorm{\proposal_x-\proposal_y}$, we apply Pinsker's
inequality~\citep{Cover}, which guarantees that $\tvnorm{\proposal_x -
  \proposal_y} \leq \sqrt{2 \kldiv{ \proposal_x}{ \proposal_y}}$.
Given multivariate normal distributions $\mathcal{G}_1 =
\NORMAL\parenth{\mu_1, \Sigma}$ and $\mathcal{G}_2 =
\NORMAL\parenth{\mu_2, \Sigma}$, the Kullback-Leibler divergence
between the two is given by
\begin{align}
\label{eq:kl_divergence_expression}
    \kldiv{\mathcal{G}_1}{\mathcal{G}_2} =
    \frac{1}{2}\parenth{\mu_1-\mu_2}^\top
    \Sigma^{-1}\parenth{\mu_1-\mu_2}.
\end{align}
Substituting $\mathcal{G}_1 = \proposal_x$ and $\mathcal{G}_2 =
\proposal_y$ into the above expression and applying Pinsker's
inequality, we find that
\begin{align*}
 \tvnorm{\proposal_x-\proposal_y} \leq
 \sqrt{2\kldiv{\proposal_x}{\proposal_y}} & =
 \frac{\enorm{\mean_x-\mean_y}}{\sqrt{2\step}} \\
& \stackrel{(i)}{=}
 \frac{\enorm{(x-\step\gradf(x))-(y-\step\gradf(y))}}{\sqrt{2\step}},
\end{align*}
where step~(i) follows from the
definition~\eqref{eq:proposal_distribution} of the mean $\mean_x$.
Consequently, in order to establish the
claim~\eqref{eq:proposal_difference}, it suffices to show that
\begin{align*}
  \enorm{(x-\step\gradf(x))-(y-\step\gradf(y))} \leq \enorm{x-y}.
\end{align*}
Recalling that $\opnorm{\mymat}$ denotes the $\ell_2$-operator norm of
a matrix $\mymat$ (equal to the maximum singular value), we have
\begin{align*}
  \enorm{(x-\step\gradf(x))-(y-\step\gradf(y))} &= \enorm{\int_0^1
    \brackets{\Ind-\step \hessf(x+t(x-y))} (x-y) dt}\\ &\leq \int_0^1
  \enorm{\brackets{\Ind-\step \hessf(x+t(x-y))}
    (x-y)}dt\\ &\stackrel{(i)}{\leq} \sup_{z \in \realdim}
  \opnorm{\Ind_\dims -\step\hessf(z)} \; \enorm{x-y},
\end{align*}
where step~(i) follows from the definition of the operator norm.
Lemma~\ref{lemma:strong_convexity}\ref{item:sc_hessian_bound} and
Lemma~\ref{lemma:smoothness}\ref{item:smooth_hessian_bound} guarantee
that the Hessian is sandwiched as $\scparam\Ind_\dims\preceq \hessf(z)
\preceq \smoothness\Ind_\dims$ for all $z \in \realdim$, where
$\Ind_\dims$ denotes the $\dims$-dimensional identity matrix.  From
this Hessian sandwich, it follows that
\begin{align*}
  \opnorm{\Ind_\dims - \step \hessf(x)} = \max\braces{\abss{1 - \step
      \smoothness}, \abss{1-\step\scparam}} \; < \; 1.
\end{align*}
Putting together the pieces yields the claim.


\subsubsection{Proof of claim~\eqref{eq:transition_difference}}
\label{ssub:proof_of_claim_eq:transition_difference}

Let $\proposal_1$ be a distribution admitting a density $\density_1$
on $\realdim$, and let $\proposal_2$ be a distribution which has an
atom at $x$ and admitting a density $\density_2$ on
$\realdim\backslash\braces{x}$.  The total variation distance between
the distributions $\proposal_1$ and $\proposal_2$ is given by
\begin{align}
  \label{EqnTVFormula}
\tvnorm{\proposal_1 - \proposal_2} =
\frac{1}{2}\parenth{\proposal_2(\braces{x}) +
  \int_{\realdim}\abss{\density_1(z)-\density_2(z)}dz}.
\end{align}
The accept-reject step for MALA implies that
\begin{align}
  \transition_x(\braces{x}) = 1-\int_{\realdim} \min\braces{1,
    \frac{\target(z)\cdot \density_z(x)}{\target(x) \cdot
      \density_x(z)}} \density_x(z) dz,
  \label{eq:remain_at_x}
\end{align}
where $p_x$ denotes the density corresponding to the proposal
distribution~$\proposal_x = \NORMAL(x-\step\gradf(x), 2\step\Ind_\dims)$.
From this fact and the formula~\eqref{EqnTVFormula}, we find that
\begin{align}
 \tvnorm{\proposal_x - \transition_x} & = \frac{1}{2}
 \parenth{\transition_x(\braces{x}) + \int_{\realdim} \density_x(z) dz
   - \int_{\realdim} \min\braces{1, \frac{\target(z)\cdot
       \density_z(x)}{\target(x) \cdot \density_x(z)}} \density_x(z)
   dz} \notag \\
& = \frac{1}{2} \parenth{2 - 2 \int_{\realdim} \min\braces{1,
     \frac{\target(z)\cdot \density_z(x)}{\target(x)\cdot
       \density_x(z)}} \density_x(z) dz }\notag \\
& = 1 - \Exs_{z \sim \proposal_x} \brackets{\min\braces{1,
     \frac{\target(z) \cdot \density_z(x)}{\target(x)\cdot
       \density_x(z)}}}.
  \label{eq:intermediate_transition_difference}
\end{align}
By applying Markov's inequality, we obtain
\begin{align}
\label{eq:markov_inequality}
  \Exs_{z \sim \proposal_x} \brackets{\min\braces{1,
      \frac{\target(z)\cdot \density_z(x)}{\target(x)\cdot
        \density_x(z)}}} \geq \alpha \; \Prob\brackets{\frac{\target(z)
      \cdot \density_z(x)}{\target(x)\cdot \density_x(z)} \geq \alpha}
  \quad \mbox{for all $\alpha \in (0, 1]$}.
\end{align}
We now derive a high probability lower bound for the ratio
$\brackets{\target(z) \density_z(x)}/\brackets{\target(x)
  \density_x(z)}$.  Noting that \mbox{$\target(x) \propto
  \exp(-\myfun(x))$} and $\density_x(z) \propto
\exp\parenth{-\enorm{x-\step\gradf(x)-z}^2/(4\step)}$, we have
\begin{align}
\displaystyle\frac{\target(z)\cdot \density_z(x)}{\target(x)\cdot
  \density_x(z)} &= \frac{\exp\biggparenth{-\myfun(z) -
    \frac{\enorm{x-z+\step\gradf(z)}^2}{4\step}}}
                  {\exp\biggparenth{-\myfun(x)
                      -\frac{\enorm{z-x+\step\gradf(x)}^2}{4\step}}}
                  \notag \\
& = {\exp \biggparenth{\frac{4\step(\myfun(x)-\myfun(z))+
                        \enorm{z-x+\step\gradf(x)}^2 -
                        \enorm{x-z+\step\gradf(z)}^2}{4\step}}}.
    \label{eq:density_ratio_expression}
\end{align}
Keeping track of the numerator of this exponent, we find that
\begin{align}
& 4\step(\myfun(x)-\myfun(z)) + \enorm{z-x+\step\gradf(x)}^2 -
  \enorm{x-z+\step\gradf(z)}^2\notag \\
  & \quad= 4\step(\myfun(x)-\myfun(z)) + \enorm{z-x}^2 +
  \enorm{\step\gradf(x)}^2+
  2\step(z-x)\tp\gradf(x)\notag \\
& \quad\quad\quad- \enorm{x-z}^2 -\enorm{\step\gradf(z)}^2 -
  2\step(x-z)\tp\gradf(z)\notag\\ &\quad= 2\step
  \underbrace{(\myfun(x)-\myfun(z)-(x-z)\tp\gradf(x))}_{\term_1} +
  2\step\underbrace{(\myfun(x)-\myfun(z)-(x-z)\tp\gradf(z))}_{\term_2}
  \notag \\
& \quad\quad\quad+ \step^2
  \underbrace{\parenth{\enorm{\gradf(x)}^2-\enorm{\gradf(z)}^2}}_{\term_3}.
    \label{eq:density_ratio_numerator}
\end{align}
Now we provide lower bounds for the terms $\term_i$, $i = 1, 2, 3$
defined in the above display.  Since $\myfun$ is strongly convex and
smooth, applying
Lemma~\ref{lemma:strong_convexity}\ref{item:sc_quadratic_bound} and
Lemma~\ref{lemma:smoothness}\ref{item:smooth_quadratic_bound} yields
\begin{align}
  \term_1 \geq -\frac{\smoothness}{2}\enorm{x-z}^2,\quad\text{and}\quad
  \term_2 \geq \frac{\scparam}{2}\enorm{x-z}^2.
    \label{eq:m1_m2_bounds}
\end{align}
In order to lower bound $\term_3$, we observe that
\begin{align}
    \term_3 = \enorm{\gradf(x)}^2-\enorm{\gradf(z)}^2 &=
    \inprod{\gradf(x)+\gradf(z)}{\gradf(x)-\gradf(z)}\notag\\ &\stackrel{(i)}{\geq}
    -\enorm{\gradf(x)+\gradf(z)}
    \enorm{\gradf(x)-\gradf(z)}\notag\\ &\stackrel{(ii)}{\geq}
    -\parenth{2\enorm{\gradf(x)}+\smoothness \enorm{x-z}}
    \smoothness\enorm{x-z},
     \label{eq:bound_on_grad_square_difference}
\end{align}
where step~(i) follows from the Cauchy-Schwarz's inequality and
step~(ii) from the triangle inequality and $\smoothness$-smoothness of
the function $\myfun$
(cf. Lemma~\ref{lemma:smoothness}\ref{item:smooth_lipschitz_bound}).

Combining the bounds~\eqref{eq:m1_m2_bounds}
and~\eqref{eq:bound_on_grad_square_difference} with
equations~\eqref{eq:density_ratio_numerator}
and~\eqref{eq:density_ratio_expression}, we have established that
\begin{align}
\label{eq:density_ratio_bounds}
\displaystyle\frac{\target(z)\cdot \density_z(x)}{\target(x)\cdot
  \density_x(z)} \geq
\exp\parenth{\underbrace{-\frac{1}{4}(\smoothness-\scparam)\enorm{x-z}^2
    -\frac{\step}{4}\parenth{ 2\smoothness\enorm{x-z}\enorm{\gradf(x)}
      + \smoothness^2 \enorm{x-z}^2 }}_{\rdefn \bigterm}}.
\end{align}
Now to provide a high probability lower bound for the term $\bigterm$,
we make use of the standard chi-squared tail bounds and the following
relation between $x$ and $z$:
\begin{align*}
z \stackrel{(d)}{=} x - \step \gradf(x) + \sqrt{2\step} \rvg,
\end{align*}
where $\rvg\sim\NORMAL(0, \Ind_\dims)$ and $\stackrel{(d)}{=}$ denotes
equality in distribution.  We have
\begin{align*}
\enorm{x-z} = \enorm{\step \gradf(x) + \noise} \leq
\step\enorm{\gradf(x)} + \sqrt{2\step}\enorm{\rvg},
\end{align*}
which also implies
\begin{align*}
\enorm{x-z}^2 \leq 2\step^2 \enorm{\gradf(x)}^2 +
4\step\enorm{\rvg}^2.
\end{align*}
Using these two inequalities, we find that
\begin{multline*}
\bigterm \geq -\frac{(\smoothness-\scparam)\step^2}{2}
\enorm{\gradf(x)}^2 - (\smoothness-\scparam)\step \enorm{\rvg}^2
-\frac{\smoothness\step^2}{2}\enorm{\gradf(x)}^2  \\
-
\frac{\smoothness\step\sqrt{\step}}{\sqrt{2}}
\enorm{\gradf(x)}\enorm{\rvg}
- \frac{\smoothness^2\step^3}{2}\enorm{\gradf(x)}^2
-\smoothness^2\step^2\enorm{\rvg}^2.
\end{multline*}
Simplifying and using the fact that $\smoothness\step \leq 1$, we obtain that
\begin{align*}
\bigterm & \geq -2 \parenth{\smoothness\step^2\enorm{\gradf(x)}^2 +
  \smoothness\step \enorm{\rvg}^2 + \smoothness\step\sqrt{\step}
  \enorm{\gradf(x)}\enorm{\rvg} }.
\end{align*}
Since $x \in \truncball_\res$, we have
\begin{align}
\label{eq:bound_on_gradf}
  \enorm{\gradf(x)} = \enorm{\gradf(x)-\gradf(\xstar)}
  \stackrel{(i)}{\leq} \smoothness \enorm{x-\xstar} \leq \smoothness
  \sqrt{\frac\dims\scparam} \radius(\res) \rdefn \maxgrad_\res,
\end{align}
where inequality~(i) follows from the
property~\ref{item:smooth_lipschitz_bound} of
Lemma~\ref{lemma:smoothness}.  Thus, we have shown that
 \begin{align}
 \label{eq:final_bigterm}
  \bigterm & \geq -2 \parenth{\smoothness\step^2\maxgrad_\res^2 +
    \smoothness\step \enorm{\rvg}^2 + \smoothness\step\sqrt{\step}
    \maxgrad_\res\enorm{\rvg} }.
 \end{align}
Standard tail bounds for $\chi^2$-variables guarantee that
\begin{align*}
  \mbox{$\Prob
\brackets{\enorm{\rvg}^2 \leq \dims\scalar_\tvscalar} \geq
(1-\tvscalar/16)$ for $\scalar_\tvscalar =
1+2\sqrt{\log(16/\tvscalar)} + 2\log(16/\tvscalar)$.}
\end{align*}
A simple
observation reveals that the function $\stepext$ defined in
equation~\eqref{eq:step_size_function} was chosen such that for any
$\step \leq \stepext(\res, \tvscalar)$, we have
\begin{align*}
    \smoothness \step^2 \maxgrad_\res^2 \leq \frac{\tvscalar}{128},
    \quad \smoothness\step\dims\scalar_\tvscalar \leq
    \frac{\tvscalar}{64}, \quad\text{and},\quad
    \smoothness\step\sqrt{\step}\maxgrad_\res\sqrt{\dims\scalar_\tvscalar}\leq
    \frac{\tvscalar}{128}.
\end{align*}
Combining this observation with the high probability bound on
$\enorm{\rvg}$ and using the inequality~\eqref{eq:final_bigterm} we
obtain that $\bigterm \geq -\tvscalar/16$ with probability at least
$1-\tvscalar/16$.  Plugging this bound in the
inequality~\eqref{eq:density_ratio_bounds}, we find that
\begin{align*}
\Prob \brackets{\displaystyle\frac{\target(z)\cdot
    \density_z(x)}{\target(x)\cdot \density_x(z)} \geq
  \exp\parenth{-\frac{\tvscalar}{16}}} \geq (1-\tvscalar/16).
\end{align*}
Thus, we have derived a desirable high probability lower bound on the
accept-reject ratio.  Substituting $\scalar = \exp(-\tvscalar/16)$ in
the inequality~\eqref{eq:markov_inequality} and using the fact that
\mbox{$e^{-\tvscalar/16} \geq 1 -\tvscalar/16$} for any scalar
$\tvscalar > 0$, we find that
\begin{align*}
\Exs_{z \sim \proposal_x} \brackets{\min\braces{1,
    \frac{\target(z)\cdot \density_z(x)}{\target(x) \cdot
      \density_x(z)}}} \geq 1 - \frac{\tvscalar}{8}, \quad\text{ for
  any } \tvscalar \in (0, 1) \text{ and } \step \leq
\stepext(\res,\tvscalar).
\end{align*}
Substituting this bound in the
inequality~\eqref{eq:intermediate_transition_difference} completes the
proof.


\subsection{Proof of Theorem~\ref{thm:mrw_mixing}}
\label{sub:proof_of_theorem_thm:mrw_mixing}

The proof of this theorem is similar to the proof of
Theorem~\ref{thm:mala_mixing}.  We begin by claiming that
\begin{subequations}
  \begin{align}
  \label{eq:mrw_proposal}
  \tvnorm{\proposal_x^{\tagmrw(\step)}-\proposal_y^{\tagmrw(\step)}}
  &= \frac{\tvscalar}{\sqrt{2}} \ \quad \text{for all } x, y
  \text{ such that } \enorm{x-y} \leq \tvscalar\sqrt{\step}\\
  \label{eq:mrw_transition}
  \tvnorm{\proposal_x^{\tagmrw(\step)}-\transition_x^{\tagmrw(\step)}}
  &= \frac{\tvscalar}{8} \quad\quad \text{for all } x \in \truncballres,
  \end{align}
\end{subequations}
for any $\step \leq \UNICON\tvscalar^2\scparam/(
\scalar_\tvscalar \dims \smoothness^2 \radius(\res))$
for some universal constant $\UNICON$.
Plugging $\res = \threshold/(2\warmparam)$, $\tvscalar = 1/2$
and arguing as in Section~\ref{sub:proof_of_theorem_thm:mala_mixing},
we find that
\mbox{$\conductance_{\threshold/2\warmparam}^{\tagmrw{(\step)}}
\geq \unicontwo \sqrt{\scparam\step}$}
for some universal constant $\unicontwo$.
Using the convergence rate~\eqref{eq:lovasz_tv_bound}, we obtain that
\begin{align}
\label{eq:mrw_mh_bound_mixing_time}
\tvnorm{\transition_{\tagmrw(\step)}^k(\initial)-\Target}
\leq \warmparam \frac{\threshold}{2\warmparam}
  + {\warmparam} e^{-k \scparam\step /\unicontwo}
\leq \threshold \quad\text{ for all }\quad k \geq
\frac{\unicontwo}{\scparam\step} \cdot
\log\parenth{\frac{2{\warmparam}}{\threshold}},
\end{align}
for a suitably large constant $\unicontwo$.
Substituting $\step \leq
\UNICON\scparam/\parenth{\dims
\smoothness^2\radius(\threshold/2\warmparam)}$,
yields the claimed bound on mixing
time of MRW.

It is now left to establish our earlier claims~\eqref{eq:mrw_proposal}
and \eqref{eq:mrw_transition}.  Note that
\mbox{$\proposal_x^{\tagmrw(\step)} = \NORMAL(x, 2\step \Ind_\dims)$.}
For brevity, we drop the superscripts from our notations.  Using the
expression~\eqref{eq:kl_divergence_expression} for the KL-divergence
and applying Pinsker's inequality leads to the upper bound
\begin{align*}
  \tvnorm{\proposal_x-\proposal_y} {\leq}
  \sqrt{2\kldiv{\proposal_x}{\proposal_y}} =
  \frac{\enorm{x-y}}{\sqrt{2\step}},
\end{align*}
which implies the claim~\eqref{eq:mrw_proposal}.

We now prove the bound~\eqref{eq:mrw_transition}.  Letting
$\density_x$ to denote the density of the proposal distribution
$\proposal_x$ and using the
bounds~\eqref{eq:intermediate_transition_difference} and
\eqref{eq:markov_inequality}, it suffices to prove that
\begin{align}
\label{eq:mrw_intermediate_bound}
  \Prob_{z \sim
    \proposal_x}\brackets{\displaystyle\frac{\target(z)}{\target(x)}
    \geq \exp\parenth{-\frac{\tvscalar}{16}}} \stackrel{(i)}{=}
  \Prob_{z \sim \proposal_x}\brackets{ \myfun(x) - \myfun(z) \geq
    -\frac{\tvscalar}{16}} \geq (1-\tvscalar/16),
\end{align}
where step~(i) follows from the fact that \mbox{$\target(x)
  \propto e^{-\myfun(x)}$.}  We have
\begin{align}
  \myfun(x) - \myfun(z)
  \stackrel{(i)}{\geq} \gradf(z)\tp(x-z)
  &= \parenth{\gradf(z)-\gradf(x)}\tp(x-z) + \gradf(x)\tp(x-z)\notag\\
  &\stackrel{(ii)}{\geq} -\smoothness\enorm{x-z}^2 + \gradf
  (x)\tp(x-z)\notag\\
  &=  - 2\smoothness \step \enorm{\rvg}^2
  +\sqrt{2\step} \gradf(x)\tp\rvg \label{eq:mrw_last_bound}
\end{align}
where the step~(i) follows from the convexity of the function
$\myfun$, step~(ii) the smoothness of the
function~$\myfun$ (Lemma~\ref{lemma:smoothness}\ref{item:smooth_gradient_x_bound}).
Note that the random variable
$\chi \defn \gradf(x)\tp\rvg \sim \NORMAL(0, \enorm{\gradf(x)}^2)$
and that $\enorm{\gradf(x)} \leq \maxgrad_\res$ for any \mbox{$x \in \truncballres$}.
Consequently, we have $\chi \geq -\maxgrad_\res \cdot 2\sqrt{\log(32/\tvscalar)}$
with probability
at least $1-\tvscalar/32$.
On the other hand, using the standard tail bound for a Chi-squared random
variable, we obtain that
\mbox{$\Prob\brackets{\enorm{\rvg}^2 \geq \dims
   \scalar_\tvscalar} \leq \tvscalar/32$} for $\scalar_\tvscalar =
1+2\sqrt{\log(32/\tvscalar)} + 2\log(32/\tvscalar)$.
Recalling that $\maxgrad_\res = \smoothness
  \sqrt{\frac\dims\scparam} \radius(\res)$ and doing straightforward
calculation reveals that for
$ \step \leq \frac{\tvscalar^2}{(8192 \scalar_\tvscalar \dims
  \frac{\smoothness^2}{\scparam}\radius(\res))}$,
we have
\begin{align*}
2\smoothness\step
  \dims\scalar_\tvscalar \leq \frac{\tvscalar}{64}
  \quad \text{ and } \quad
  \sqrt{2\step} \maxgrad_\res 2\sqrt{\log(32/\tvscalar)} \leq
  \frac{3\tvscalar}{64}
\end{align*}
Combining these bounds with the high probability statements above and plugging
in the inequality~\eqref{eq:mrw_last_bound}, we find that
\mbox{$\myfun(x)-\myfun(z) \geq -\tvscalar/16$} with probability
at least $1-\tvscalar/16$, which yields the claim~\eqref{eq:mrw_intermediate_bound}.


\section{Discussion}
\label{sec:discussion}
In this paper, we derived non-asymptotic bounds on the mixing time of
the Metropolis adjusted Langevin algorithm and Metropolized random
walk for log-concave distributions. These algorithms are based on a
two-phase scheme: (1) a proposal step followed by (2) an accept-reject
step.  Our results show that the accept-reject step while leading to
significant complications in the analysis is practically very useful:
algorithms applying this step mix significantly faster than the ones
without it. In particular, we showed that for a strongly log-concave
distribution in $\realdim$ with condition number $\condition$, the
$\threshold$-mixing time for MALA is of $\calo\parenth{\dims
  \condition\log(1/\threshold)}$.  This guarantee significantly better
than the $\calo\parenth{\dims \condition^2/\threshold^2}$ mixing time
for ULA established in the literature.  We also proposed a modified
version of MALA to sample from non-strongly log-concave distributions
and showed that it mixes in $\calo\parenth{\dims^2/\threshold^{1.5}}$;
thus, this algorithm dependency on the desired accuracy $\threshold$
when compared to the $\calo\parenth{\dims^3/\threshold^{4}}$ mixing
time for ULA for the same task.  Furthermore, we established
$\calo\parenth{\dims \condition^2\log(1/\threshold)}$ mixing time
bound for the Metropolized random walk for log-concave sampling.

Several fundamental questions arise from our work.  All of our results
are upper bounds on mixing time, and our simulation results suggest
that they are tight for the choice of step size used in the Theorem
statements.  However, simulations from
Section~\ref{sub:step_size_vs_accept_reject_rate} suggest that warmness
parameter $\warmparam$ should not affect the choice of step size too much
and hence potentially larger choices of step sizes (and thereby faster mixing)
are possible. To this end, in a recent pre-print~\citep{CheDwiWaiYu19},
we established faster mixing time bounds for MALA and MRW from a non-warm
start where we show that the dependence on warmness can be improved from
$\log\warmparam$ to $\log\log \warmparam$. 
Moreover, for a deterministic start, one may consider running ULA
for a few steps run to obtain moderate accuracy, and then run MALA initialized
with the ULA iterates (thereby providing a warm start to MALA). In practice,
we find that this hybrid procedure can generate highly accurate samples
in reasonably few number of iterations.

Another open question is to sharply delineate the fundamental gap
between the mixing times of first-order sampling methods and that of
zeroth-order sampling methods. Noting that MALA is a first-order
method while MRW is a zeroth-order method, from our work, we obtain
that two class of methods differ in a factor of the condition number
$\condition$ of the target distribution. It is an interesting question
to determine whether this represents a sharp gap between these two
classes of sampling methods.

The current state-of-the-art algorithm, namely Hamiltonian Monte
Carlo~\citep{neal2011mcmc} can be seen as a multi-step generalization
of MALA. Instead of centering the proposal after one gradient-step,
HMC simulates an ODE in an augmented space for a few time
steps. Indeed, MALA is equivalent to a particular one-step
discretization of the ODE associated with HMC. Nonetheless, the
practically used HMC makes use of multi-step discretization and is
more involved than MALA. Empirically HMC has proven to have superior
mixing times for a broad class of distributions (and not just
log-concave distributions).  A line of recent
work~\citep{bou2018coupling,mangoubi2017rapid,mangoubi2018dimensionally}
provides theoretical guarantees for HMC in different settings. Several
of these works analyze an idealized version of HMC or the discretized
version without the accept-reject step.  In a recent
pre-print~\citep{CheDwiWaiYu19}, we have provided some theoretical
guarantees on the convergence of Metropolized HMC, which is the most
practical version of HMC.

\subsection*{Acknowledgements}

This work was supported by Office of Naval Research grant DOD
ONR-N00014 to MJW, and by ARO W911NF1710005, NSF-DMS 1613002 and the
Center for Science of Information (CSoI), US NSF Science and
Technology Center, under grant agreement CCF-0939370 to BY.  In
addition, MJW was partially supported by National Science Foundation
grant NSF-DMS-1612948, and RD was partially supported by the Berkeley
Fellowship.


\appendix

\section{Some basic properties}
\label{sec:some_basic_properties}

In this appendix, we state a few basic properties of strongly-convex
and smooth functions that we use in our proofs.  See the
book~\citep{boyd2004convex} for more details.
\begin{lemma}[Equivalent characterizations of strong convexity]
\label{lemma:strong_convexity}
  For a twice differentiable convex function $\myfun:\realdim \mapsto
  \real$, the following statements are equivalent:
  \begin{enumerate}[label=(\alph*)]
    \item\label{item:sc_definition} The function $\myfun$ is
      $\scparam$-strongly-convex.
    \item\label{item:sc_convex_quadratic} The function $x \mapsto
      \myfun(x) - \displaystyle\frac{\scparam}{2}\enorm{x-\xstar}^2$
      is convex (for any fixed point $\xstar$).
    \item\label{item:sc_quadratic_bound} For any $x, y \in \realdim$, we have
    \begin{align*}
      \myfun(y) \geq \myfun(x) + \gradf(x)\tp(y-x) + \frac{\scparam}{2} \enorm{x-y}^2.
    \end{align*}
    \item\label{item:sc_lipschitz_bound} For any $x, y \in \realdim$,
      we have
    \begin{align*}
      \enorm{\gradf(x)-\gradf(y)} \geq \scparam \enorm{x-y}.
    \end{align*}
    \item\label{item:sc_gradient_x_bound} For any $x, y \in
      \realdim$, we have
    \begin{align*}
      \parenth{\gradf(x)-\gradf(y)}^\top(x-y) \geq \scparam \enorm{x-y}^2.
    \end{align*}
    \item\label{item:sc_hessian_bound} For any $x \in \realdim$, the
      Hessian is lower bounded as $\hessf(x) \succeq
      \scparam \Ind_\dims$.
  \end{enumerate}
\end{lemma}

\begin{lemma}[Equivalent characterizations of smoothness]
  \label{lemma:smoothness}
  For a twice differentiable convex function $\myfun:\realdim \mapsto
  \real$, the following statements are equivalent:
  \begin{enumerate}[label=(\alph*)]
    \item\label{item:smooth_definition} The function $\myfun$ is
      $\smoothness$-smooth.
    \item\label{item:smooth_convex_quadratic} The function $x \mapsto
      \displaystyle\frac{\smoothness}{2}\enorm{x-\xstar}^2-\myfun(x) $
      is convex (for any fixed point $\xstar$).
    \item\label{item:smooth_quadratic_bound} For any $x, y \in \realdim$, we have
    \begin{align*}
      \myfun(y) \leq \myfun(x) + \gradf(x)\tp(y-x) +
      \frac{\smoothness}{2} \enorm{x-y}^2.
    \end{align*}
    \item\label{item:smooth_lipschitz_bound} For any $x, y \in
      \realdim$, we have
    \begin{align*}
      \enorm{\gradf(x)-\gradf(y)} \leq \smoothness \enorm{x-y}.
    \end{align*}
      \item\label{item:smooth_gradient_x_bound} For any $x, y \in
      \realdim$, we have
    \begin{align*}
      \parenth{\gradf(x)-\gradf(y)}^\top(x-y) \leq \smoothness \enorm{x-y}^2.
    \end{align*}

    \item\label{item:smooth_hessian_bound} For any $x \in \realdim$,
      the Hessian is upper bounded as $\hessf(x) \preceq \smoothness\Ind_\dims$.
  \end{enumerate}
\end{lemma}

\bibliography{mala_bib}

\end{document}